\pdfoutput=1
\documentclass[10pt,twocolumn,letterpaper]{article}

\usepackage{iccv}
\usepackage{amsmath,graphicx}
\usepackage{color}
\usepackage{graphicx}
\usepackage{amsmath}
\usepackage{amssymb}
\usepackage{epsfig}
\usepackage{amssymb}
\usepackage{algorithm}
\usepackage{graphicx}
\usepackage{epstopdf}
\usepackage{array}
\usepackage{mathtools}
\usepackage[font=footnotesize, caption=false]{subfig}
\usepackage{url}
\usepackage[noend]{algpseudocode}

\iccvfinalcopy

\ificcvfinal\fi
\begin{document}

\title{Automatic segmentation of trees in dynamic outdoor environments}


\author{Amy Tabb\\
USDA-ARS-AFRS\\
Kearneysville, West Virginia, USA\\
{\tt\small amy.tabb@ars.usda.gov}
\and
Henry Medeiros \\
Marquette University, Electrical and Computer Engineering\\
Milwaukee, Wisconsin, USA \\
{\tt\small henry.medeiros@marquette.edu}
}

\maketitle
\ificcvfinal\thispagestyle{empty}\fi


\begin{abstract}
Segmentation in dynamic outdoor environments can be difficult when the illumination levels and other aspects of the scene cannot be controlled.  Specifically in orchard and vineyard automation contexts, a background material is often used to shield a camera's field of view from other rows of crops. In this paper, we describe a method that uses superpixels to determine low texture regions of the image that correspond to the background material, and then show how this information can be integrated with the color distribution of the image to compute optimal segmentation parameters to segment objects of interest.  Quantitative and qualitative experiments demonstrate the suitability of this approach for dynamic outdoor environments, specifically for tree reconstruction and apple flower detection applications.  \footnote{Mention of trade names or commercial products in this publication is solely for the purpose of providing specific information and does not imply recommendation or endorsement by the U.S. Department of Agriculture.  USDA is an equal opportunity provider and employer.  A. Tabb acknowledges the support of US National Science Foundation grant number IOS-1339211.}\footnote{The citation information for this paper is: A. Tabb and H. Medeiros. 2018. Automatic segmentation of trees in dynamic outdoor environments. Computers in Industry. 98,90-99.  DOI 10.1016/j.compind.2018.03.002} 
\end{abstract}

\section{Introduction}
\label{sec:intro}

Segmentation is a key step in many {object detection} contexts, and when the result is accurate, can reduce the amount of information presented to subsequent steps of an autonomous computer vision system.  This paper describes a method for segmentation of a mobile background unit from tree regions in an orchard setting as part of an automated pipeline to reconstruct {and measure} the shape of leafless trees for robotic pruning and phenotyping \cite{Tabb2013,tabb2014shape,tabb2017robotic}.  Since the images are acquired outdoors, illumination conditions are not stable and may change rapidly and widely.  Furthermore, the entire {tree reconstruction and measurement} process is automated, and hundreds of images must be acquired per tree. Hence, the segmentation method must be robust and not require {manual} parameter tuning. Since our goal is to use the segmentation step as part of a {real-time} automation application, the method must also be fast.

The ability to robustly extract the silhouettes of objects of interest is generally an important step in the generation of three-dimensional models of complex objects such as trees \cite{Tabb2013} and may form a preprocessing step for other tasks, such as flower detection {\cite{dias_flower}}. Existing silhouette extraction techniques based solely on thresholding and morphological characteristics of the object of interest, however, tend to generate unsatisfactory results, particularly with respect to segmentation. This problem, as with most computer vision tasks, is further aggravated in dynamic environments, which include situations such as drastically varying illumination conditions. Hence, we propose a novel method to segment an object (in this case a tree) from a low-texture background, which is robust to significant illumination changes. 

The segmentation method proposed in this paper assumes an item of interest in a image is positioned in front of a background material of homogeneous color. It locates the low texture regions of the image using superpixels and models them using a Gaussian Mixture Model (GMM). Pixels in the image are classified according to the GMM and a mask of the background material region is created. The method is fully autonomous and does not require user input or training, other than initial setting of thresholds.  The main contribution of this work is an unsupervised method to segment foreground objects in images that is sufficiently robust to operate with various models of cameras under natural outdoor illumination conditions and is fast enough to be used in automation contexts.  The method is verified through quantitative and qualitative experiments as well as comparisons to alternative approaches based on Otsu's method \cite{otsu1979threshold} and adaptive thresholding mechanisms. 

The remainder of this paper is organized as follows. Section \ref{sec:relatedwork} presents a brief overview of methods for the segmentation of foreground objects with a particular focus on agricultural applications. Section \ref{sec:methoddescription} describes our proposed approach. A comprehensive evaluation comparing the performance of our methods to alternative approaches to foreground object segmentation is given in Section \ref{sec:experiments}. Finally, Section \ref{sec:conclusions} concludes the paper and discusses possible future research directions.

\section{Related work}
\label{sec:relatedwork}

There is a significant amount of related work on segmentation in dynamic environments for the purposes of foreground detection as recently reviewed by Bouwmans \cite{BouwmansTraditional2014}.  Traditionally, in foreground detection, the assumption is that the background can be modeled because the cameras observing a scene are not moving. The background image hence remains relatively static and objects that move with respect to the camera are considered part of the foreground.  A popular approach for foreground, or motion, detection is that of Stauffer and Grimson \cite{stauffer1999adaptive}, in which each image pixel is modeled as a mixture of Gaussians. Various extensions of \cite{stauffer1999adaptive}, from a hierarchical approach for real-time execution \cite{ParkHierarchical2006} to models that consider non-Gaussian distributions \cite{ChanGeneralized2011}, have been explored as well for the context of relatively static backgrounds.  In the scenarios under consideration in this work, however, the background may change {considerably}, so motion detection approaches are not applicable.

In the agricultural context, many applications require the segmentation of plants from soil with a moving camera; this problem has been recently surveyed by Hamuda \textit{et al.} \cite{HamudaSurvey2016}.  Concerning applications of tree segmentation, Byrne and Singh \cite{Byrne1998precise} use co-occurrence statistics to oversegment images into tree versus non-target tree regions for use in autonomous diameter measurement for a automated forestry application.  In a similar application, Ali \cite{ali2006tree} uses a combination of color and texture features fed into an artificial neural network and k-nearest neighbor classifiers to perform classification of pixels into tree and non-tree classes. Similar to our work, Botterill \textit{et al.} in \cite{botterill2016robot} also use a blue background for their design of a robotic grapevine pruner.  However, their unit is an over-the-row unit and does not have to navigate the illumination challenges {as we do with a one- or two-sided unit}.  In \cite{botterill2016robot}, training data is hand-labeled into three classes: background, wire, foreground, and color features and used to train a support vector machine (SVM), which is later used for classification. Zheng \textit{et al.} \cite{zheng2011detailed} address segmenting root material from gels by using a harmonic background subtraction method with hysteresis thresholding. 

Mobile background units, such as the ones used in our work, have been used for various purposes, from apple harvest \cite{gongal2016apple, davidson2016proof, silwal2014apple}, to grape pruning \cite{botterill2016robot}, and tree shape estimation \cite{Tabb2013,tabb2014shape,tabb2017robotic}. The advantage of using such a background, especially with trees planted in rows, is that the influence of neighboring rows of trees is eliminated.  Some units, such as those shown in \cite{botterill2016robot, gongal2016apple, davidson2016proof, silwal2014apple}, are over-the-row units to shield the imaging area from variations in illumination from the environment.  Another option has been to acquire images at night with artificial illumination to create a static background and mitigate illumination variation, such as in \cite{amatya2016detection} to detect cherry, in \cite{liu2012image} to detect tree branches using RGB-D cameras, and in \cite{wang2012automated,linker2017procedure} to detect apple fruit. Nighttime-only operation, however, significantly restricts the practical applicability of any agricultural robotic system.


\section{Method description}
\label{sec:methoddescription}

We assume a low-texture background object is present in each image, and we model the hue component of this background object according to a Gaussian mixture model with $k$ components: $p(h) = \sum_{j = 1}^k w_j \mathcal{N}(\mu_j, \sigma_j^2)$, where $\mu_j$ and $\sigma_j$ are the mean and variance of the $j$-th mixture component and $w_j$ is its corresponding weight. In the following steps, we show how we estimate this distribution and then use it to assign probabilities for each pixel in the image.  We also assume that the object of interest is positioned between the background object and the camera (see Figure \ref{fig:image1}a). Regions that extend beyond the background object are truncated. Algorithm \ref{alg:overall} shows an overview of the proposed approach. Each step of the algorithm is explained in detail in the following subsections. While the first two steps of the algorithm are independent and can be performed in parallel, the remaining steps depend upon one another and hence need to be performed in order. The sequence of steps is illustrated in Figure \ref{fig:image1}.

\begin{algorithm}
\caption{Proposed segmentation approach} \label{alg:overall}
\begin{algorithmic}[1]
\Require{Image in hue-saturation-value (HSV) color space}
\Ensure{Segmented image}
\State Compute the set of superpixels $\mathbb{S}$ using the SEEDS method \cite{VandenBergh2015} and find the subset $\mathbb{R} \subset \mathbb{S}$ of low-texture superpixels.
\State Generate a binary image $T$, by thresholding the hue channel using Otsu's algorithm.
\State Determine GMM $p(h) = \sum_{j = 1}^k w_j \mathcal{N}(\mu_j, \sigma_j^2)$, which represents the background, based on $\mathbb{R}$ and $T$.
\State Generate label image $L$ by assigning labels to individual output pixels according to the GMM.
\State Create a mask to eliminate regions outside of the background object.
\end{algorithmic}
\end{algorithm}

\subsection{Step 1: Computation of low-texture superpixels}

The first step of our approach consists of converting the image to the HSV color space and partitioning it into superpixels. We have chosen to compute the superpixels using the superpixels extracted via energy-driven sampling (SEEDS) method proposed in \cite{VandenBergh2015} and implemented in OpenCV \cite{opencv_library}. In this superpixel approach, the image is divided into a grid pattern, which serves as initial superpixel assignment.  The superpixel assignments are refined by iteratively modifying their boundaries.  

\begin{figure}[!ht]
\subfloat[]  
	{  
	\includegraphics[width=0.48\linewidth]{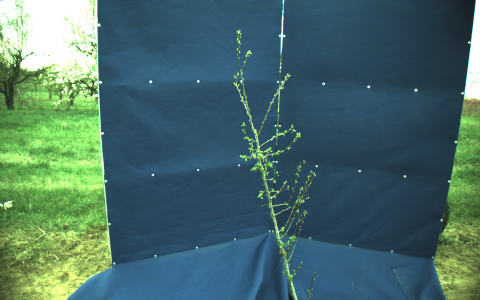}  
	\label{sf:original} }
\subfloat[]  
	{  
	\includegraphics[width=0.48\linewidth]{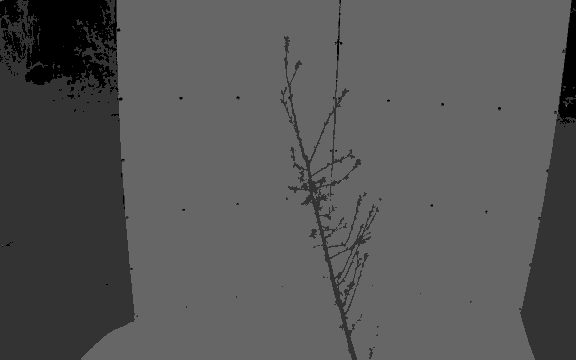}  
	\label{sf:hue} }

\subfloat[]  
	{  
	\includegraphics[width=0.48\linewidth]{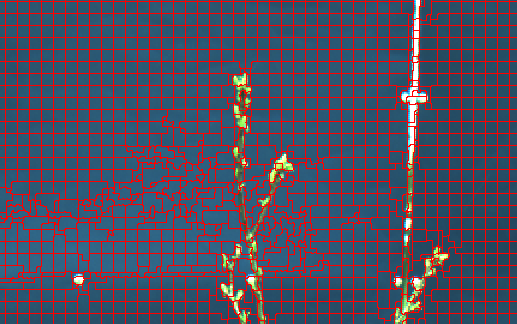} 
\label{sf:superpixels}  }
\subfloat[]  
	{  
	\includegraphics[width=0.48\linewidth]{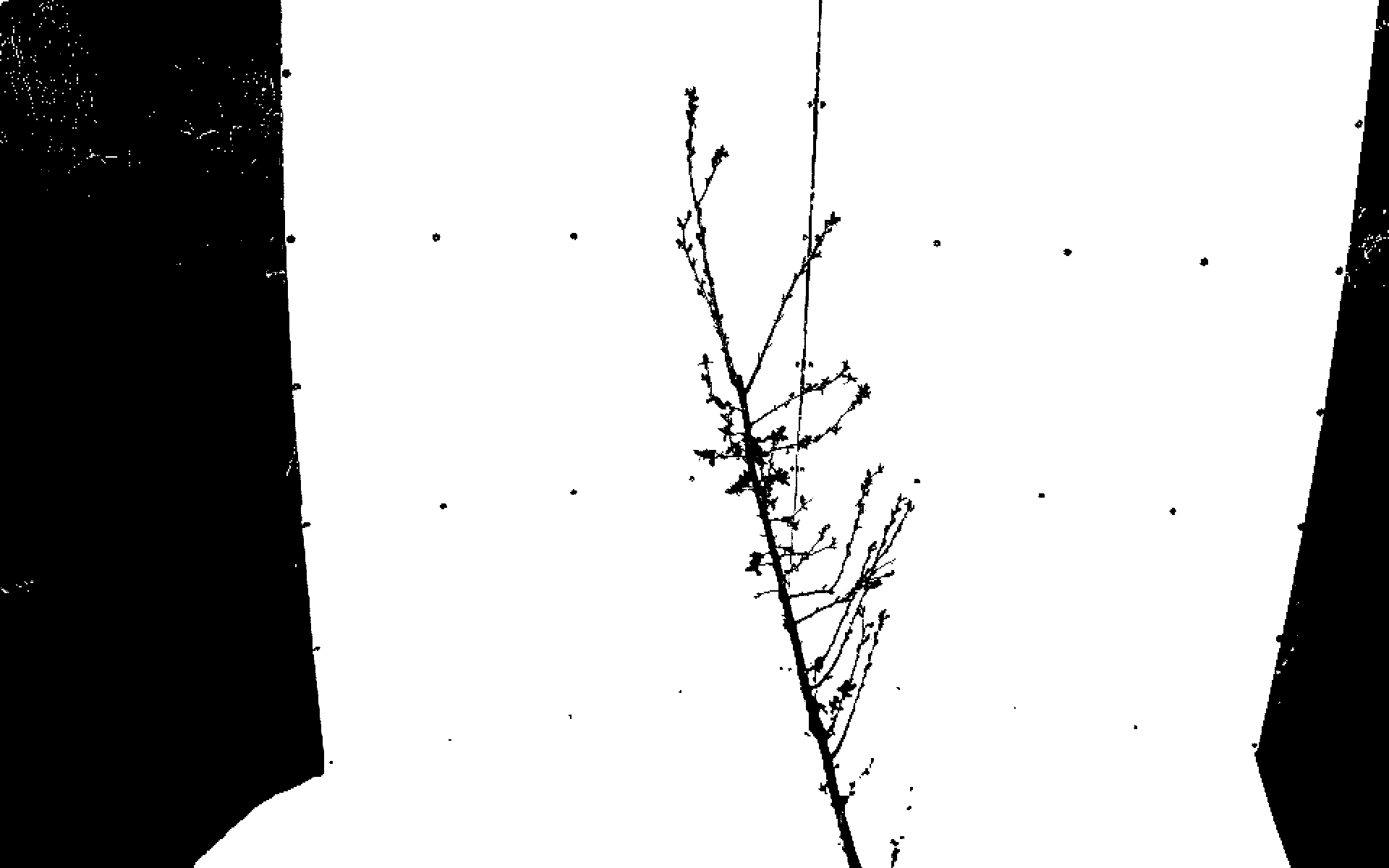}  
	\label{sf:T} }

\subfloat[]  
	{  
	\includegraphics[width=0.48\linewidth]{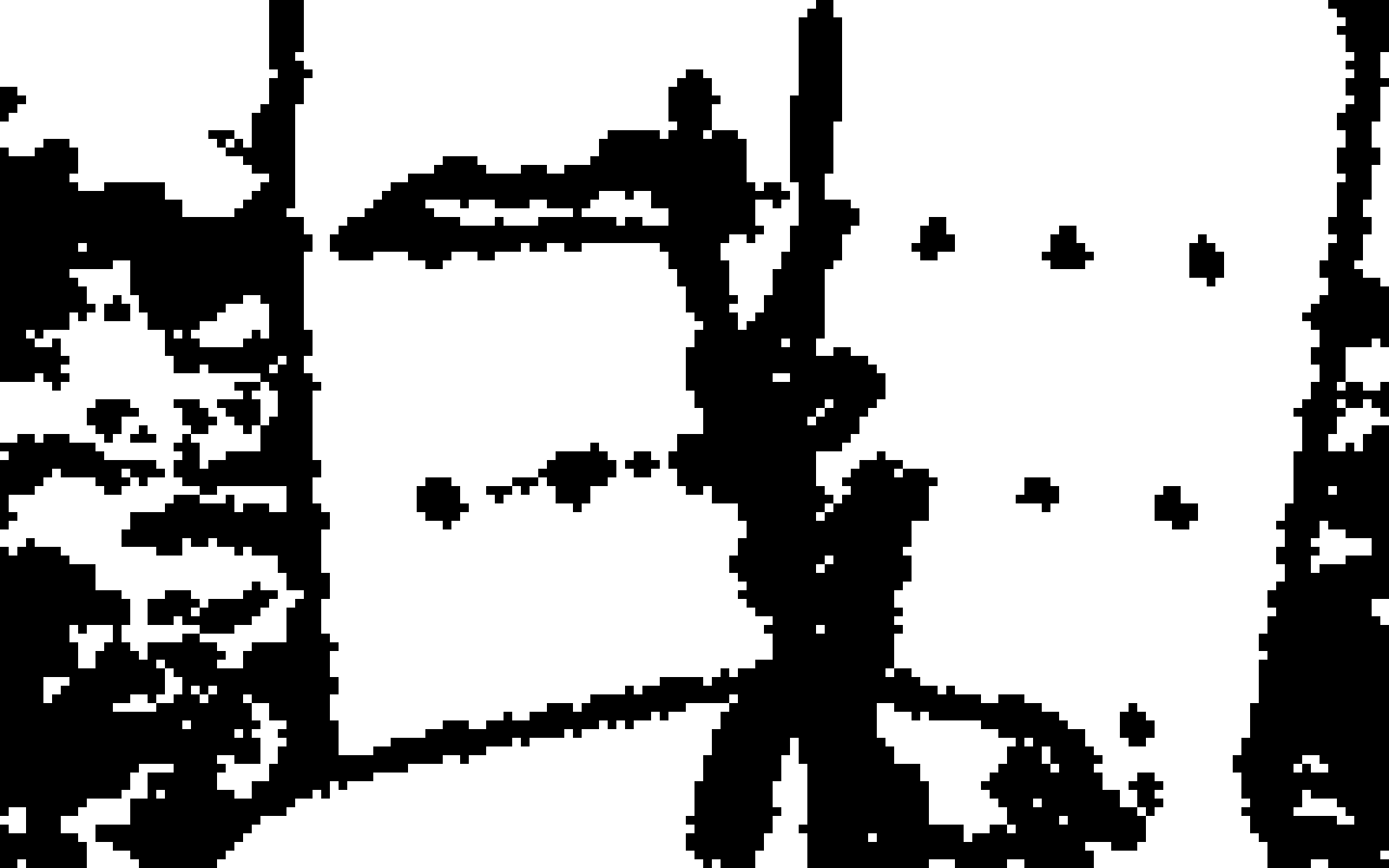}   
	\label{sf:Rect}}
\subfloat[]  
	{  
	\includegraphics[width=0.48\linewidth]{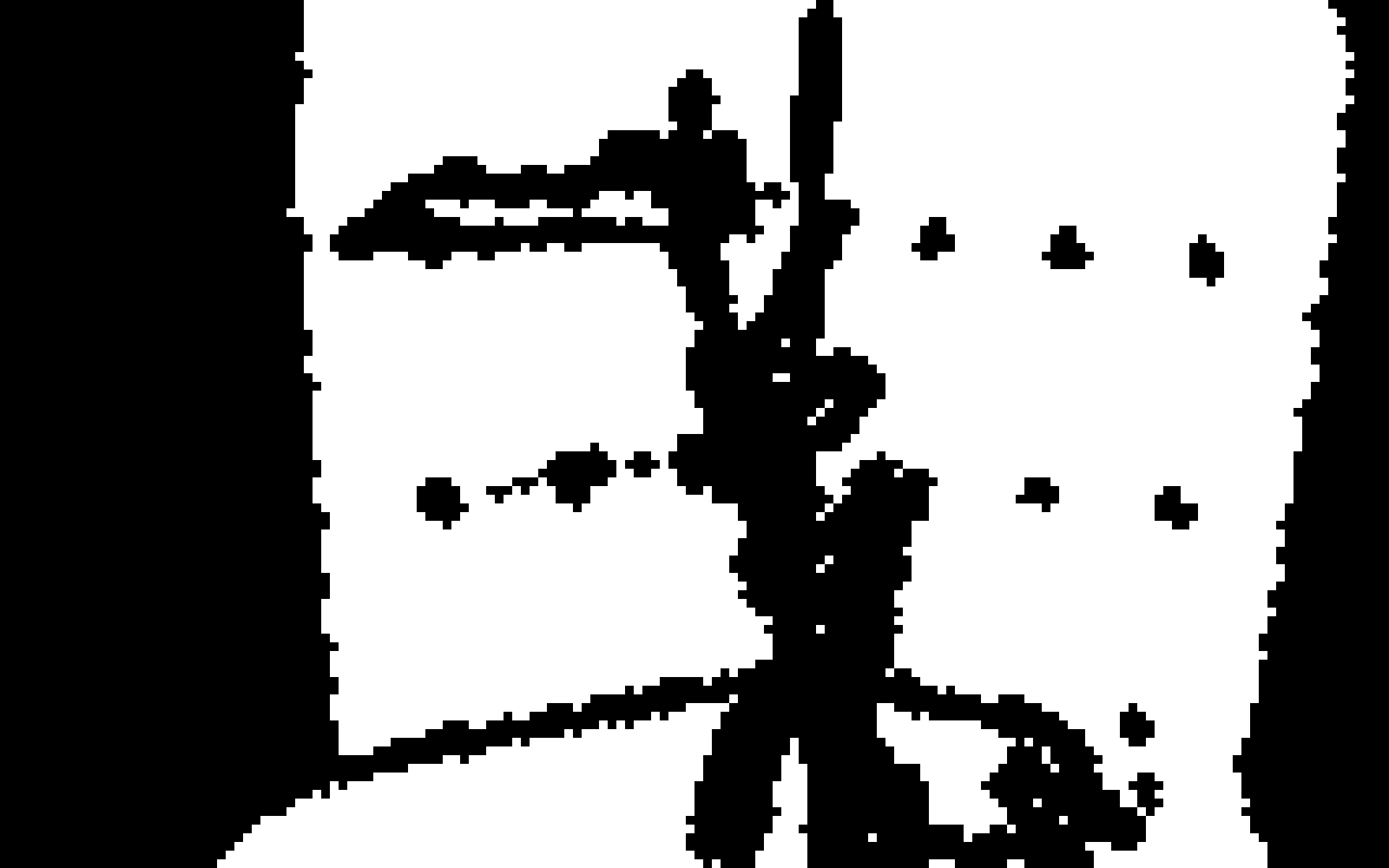}   
	\label{sf:B}}

\subfloat[]  
	{  
	\includegraphics[width=0.48\linewidth]{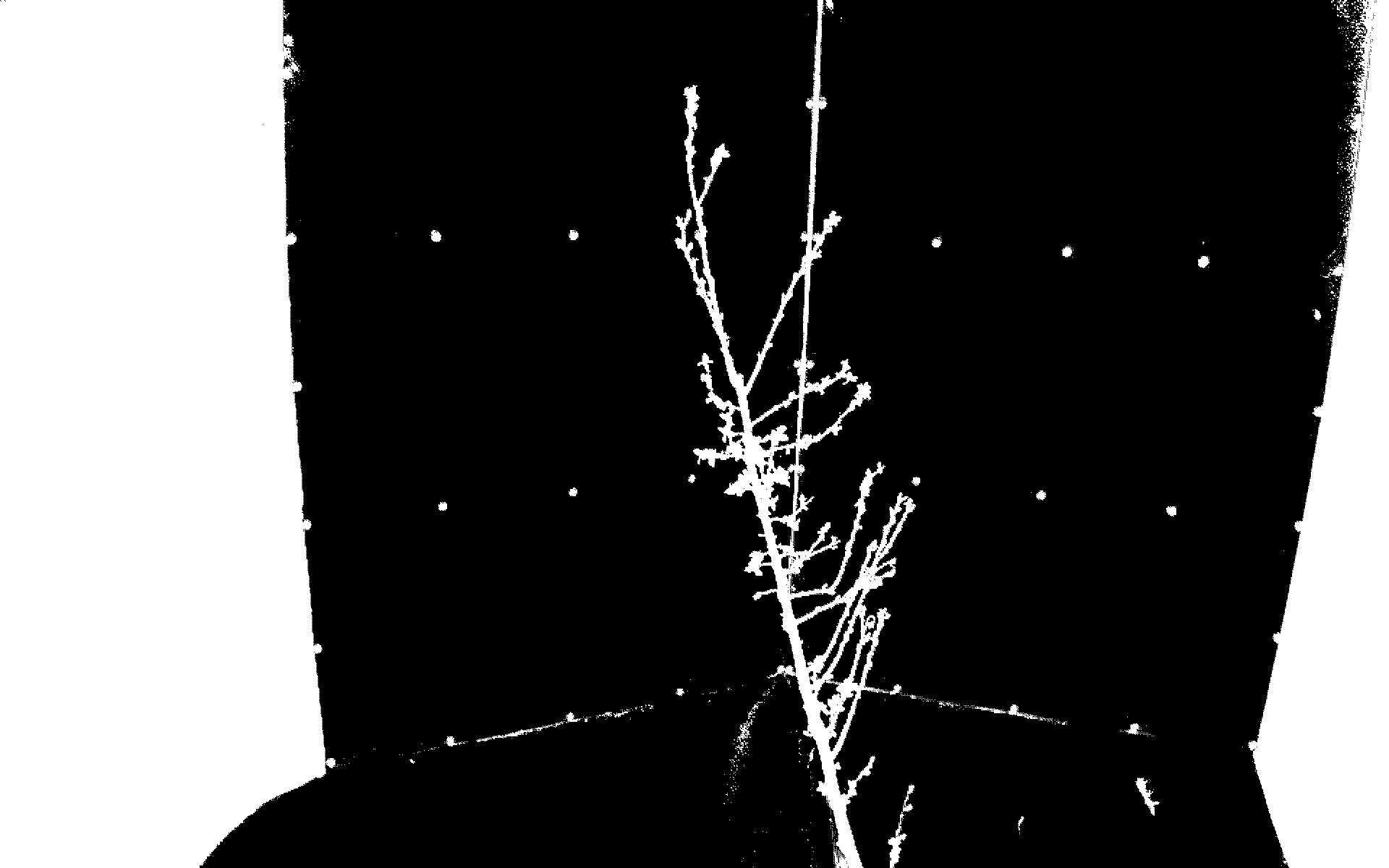}   
	\label{sf:Ptree}}
\subfloat[]  
	{
	\includegraphics[width=0.48\linewidth]{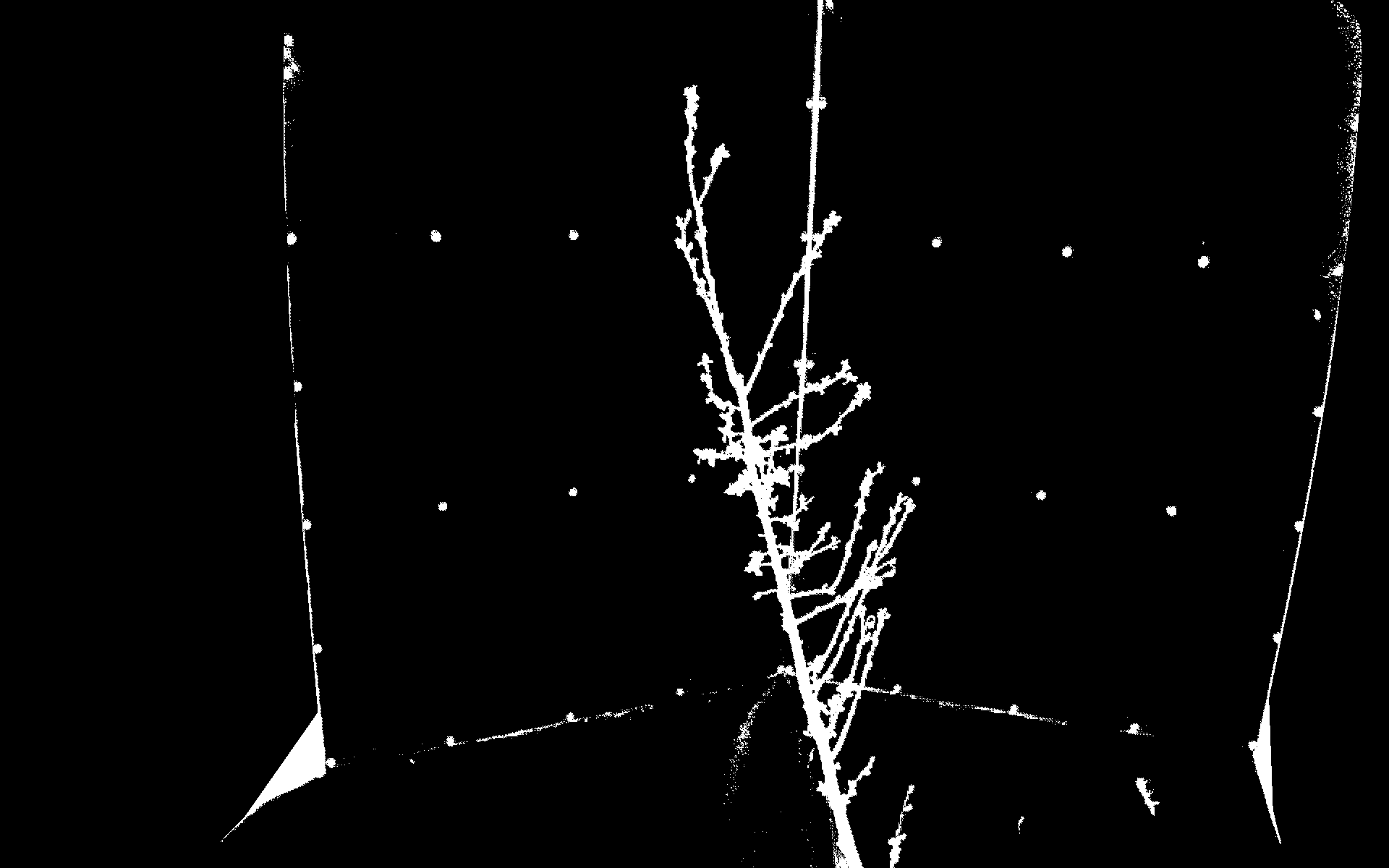}	
\label{sf:final}}
\caption{\textbf{[Best viewed in color]} \protect\subref{sf:original} Original RGB image showing the object of interest (tree) in front of the background object. \protect\subref{sf:hue} Hue channel. \protect\subref{sf:superpixels} Close-up of portion of the original image (top portion of branch) with superpixels overlaid in red.  \protect\subref{sf:T} Threshold result from step 2 ($T$).  \protect\subref{sf:Rect} The set of superpixels $\mathbb{R}$ is shown in white, indicating low texture regions. \protect\subref{sf:B} Set of superpixels $\mathbb{B}$, where white pixels indicate locations where the GMM is estimated in the hue image.  \protect\subref{sf:Ptree} Label image $L$ after label assignment step. \protect\subref{sf:final} $L$ after application of the mask in step 5.}
\label{fig:image1}
\end{figure}

Starting from an initial superpixel division of a grid, in SEEDS, pixels change label as a result of a maximizing a cost function based on a color likelihood term and optional shape prior.    We set the parameters of the SEEDS algorithm such that low texture regions have superpixels whose shape is unchanged from the initial grid assignment. That is, let $\mathbb{S}$ be the set of superpixels generated by SEEDS. Then there is a set of superpixels $\mathbb{R} \subset \mathbb{S}$, which are rectangular in shape. These are the superpixels which are unchanged from the initial assignment and correspond to low-texture regions. Figures \ref{sf:original} and \ref{sf:hue} show the original RGB image and the hue channel of its corresponding HSV representation.  Figure \ref{sf:superpixels} then shows the superpixels generated according to our proposed procedure. As the image shows, most superpixels on the background object are rectangular in shape.

\subsection{Step 2: Generation of thresholded hue image using Otsu's algorithm}
\label{ss:step2}
We then generate the binary image $T$ by thresholding the hue channel of the HSV image using Otsu's algorithm \cite{OtsuThreshold1979}. In the binary image, pixels with value below the threshold value are black and the remaining pixels are white. As explained in the next section, the image $T$ is used to generate hypotheses of low-texture regions in the image, since the color of the low-texture background object is relatively constant. In our application, the blue background object has a higher hue value than other common colors in the images such as brown, gray, or green, which facilitates the application of the proposed approach. We are not limited to a single background color though. As long as the hue value of the majority of the pixels in the background differ from those of the foreground object, $T$ can be generated using multi-level thresholding algorithms \cite{HuangOptimal2009}.

\subsection{Step 3: Estimation of the distribution of the background}

The first two steps consisted of coarse detectors for the blue background.  The superpixel approach in step one finds low texture regions, while $T$ found in step two indicates regions likely to be the background judging by relative hue as compared to the rest of the image. We now combine the information from these two steps to generate a more robust background detector. 

We determine regions where $T$ overlaps superpixels in $\mathbb{R}$ using the procedure summarized in Algorithm \ref{alg:stepone_two_overlap}. Briefly, the algorithm iterates through the superpixels in $\mathbb{R}$.  If the percentage of white pixels in the corresponding area of the thresholded image $T$ exceeds a value $\zeta \in [0, 1]$, then the superpixel is added to a set $\mathbb{B}$.  The set $\mathbb{B}$ hence consists of all the superpixels which belong to a low-texture region as determined by the SEEDS algorithm \emph{and} by its constant color.

All of the pixel locations in the set of superpixels $\mathbb{B}$ are then used to estimate the probability distribution of the background pixels. Let $h_i$ be the value of the $i$th pixel in the hue image. We assume the pixel intensities of the background material regions over the whole image can be represented by a mixture of $k$ normal distributions.   That is, the probability density is 
\begin{equation}
p(h_i) = \sum_{j = 1}^k w_j \mathcal{N}(h_i; \mu_j, \sigma^2_j)
\label{eq:GMM}
\end{equation}
where $\mathcal{N}$ is the Gaussian probability density function 
\begin{equation}
\mathcal{N}(h_i; \mu_j, \sigma^2_j)=\frac{1}{\sigma_j \sqrt{2 \pi}}e^{-\frac{(h_i-\mu_j)^2}{2\sigma_j^2}}
\label{eq:normal}
\end{equation}
The parameters of the distribution, $k$, $w_j$, $\mu_j$ and $\sigma_j$, may be obtained using the expectation-maximization (E-M) algorithm from the pixels in $\mathbb{B}$.  Initial values of the distribution are estimated using k-means clustering.  However, E-M can take a long time to converge with a large number of samples, as is the case when the number of pixels in $\mathbb{B}$ is large. We use a faster approach than E-M to create a GMM with satisfactory results as follows. Essentially, the information from the superpixels is grouped into clusters, and then the components of the GMM are computed from the clusters using all of the pixels in the superpixels that compose those clusters.  First, the sample means of the individual superpixels in $\mathbb{B}$ are clustered into $k$ groups using the kmeans++ algorithm \cite{arthur2007k}.  Then, from these clusters, the means and variances of the $k$ clusters are determined from all of the pixels contained in all the superpixels within that cluster.  Finally, the weights are computed, such that if the number of superpixels in cluster $j$ is $s_j$, then $w_j = s_j/|\mathbb{B}|$.  

\begin{algorithm}
\caption{Determination of background pixels} \label{alg:stepone_two_overlap}
\begin{algorithmic}[1]
\Require{Set $\mathbb{R}$ of low-texture superpixels and thresholded hue image $T$}
\Ensure{Set $\mathbb{B}$ of background superpixels}
\State $\mathbb{B} = \lbrace \rbrace$
\For {each superpixel $r_i \in \mathbb{R}$} 
\State {$t_i = $ number of white pixels in $T$ for the region of $r_i$.}
\If {$t_i/area(r_i) > \zeta$}
\State $\mathbb{B} = \mathbb{B} \cup {r_i}$
\EndIf
\EndFor
\end{algorithmic}
\end{algorithm} 

\subsection{Step 4: Pixel label assignment based on the background distribution}

The hue image is compared to the GMM to generate an image of pixel labels $L$ as follows. Let $p(h_i)$ be the likelihood of $h_i$ as predicted by the Gaussian Mixture Model, and let $p_t$ be a corresponding threshold value.  Then, the pixel labels $L_i$ are created according to
\begin{equation}
L_i=
\begin{cases}
1,  \quad &p(h_i) \leq p_t\\
0     \quad &p(h_i) > p_t
\end{cases}
\label{eq:foreground_label}
\end{equation}
Since the pixels in the set $\mathbb{B}$ were used to generate the background model, their labels can be automatically assigned to $L_i = 0$ during model computation (Step 3) and do not need to be revisited in this step. This has the dual benefit of reducing computation time as well as avoiding spurious noisy pixels in $\mathbb{B}$ that would make the overall method more sensitive to the value of the threshold $p_t$.  

\subsection{Step 5: Create mask to eliminate regions outside the background object}

The final step creates a mask image $M$, in which regions including the object of interest, in this case the tree, and the background object, are labeled white and all other regions are labeled black.  This mask is generated over the course of three steps.  First, $M$ is set to the inverse of $L$ and small connected components with area smaller than a threshold of $\epsilon_a$ are removed from $M$. These components typically represent false background detections in the sky or the surrounding natural environment.  It remains to connect the white sections of $M$.  The contours for each connected component are found, and then for each point on the contour, the closest point on a different contour is found and a line is drawn between the contours.  To conclude, all of the holes are filled in $M$. 

The mask is applied to the pixel label image $L$ so that if the $i$th pixel of $M$ is $0$, then this pixel is outside the region of interest and the corresponding pixel in $L$ is labeled accordingly.  In this application, we label those pixels $0$ and the pixels inside the region of interest are labeled $1$. Alternatively, our method allows pixels that do not belong to the region of interest to be labeled using a special marker value so that downstream processing steps could recognize them and treat them accordingly.

\section{Experiments}
\label{sec:experiments}

We evaluated our proposed approach on six datasets containing a total of 1001 images. As shown in Table \ref{tab:DatasetInfo}, five of these datasets were acquired outdoors with different camera models and resolutions, and one of them was acquired indoors. Most datasets consisted of images collected by multiple cameras, { and for Dataset 3 each camera has a different aperture setting}. These datasets reflect a range of illumination conditions. {All of the datasets consist of images of leafless fruit trees, although Dataset 5 is composed of images of flowering apple trees.}  

We found that the following parameters showed satisfactory performance for all of the datasets: threshold parameter $\zeta=0.8$, mask generation parameter $\epsilon_a=2000$ and threshold $p_t = 0.003$. The number of distributions in the GMM was set to $k = 10$. The SEEDS superpixel was configured for number iterations = 10, number histogram bins = 2, number of superpixels = 16,000, number of levels = 1, prior = 0, and double step = false. Also, basic morphological operations could have been applied to our results to eliminate noisy detections, but in order to present the method in a general sense, we did not apply any further processing steps past step 5.
{Parameter selection should take into account the size of the background unit in the images, as well as the thickness of the object to be detected.}

\begin{table*}[ht]
\centering{}\caption{\label{tab:DatasetInfo} Description of the six datasets used in this paper.  The Pt. Grey camera model is BFLY-PGE-23S6C-C.}
{
\begin{tabular}{l|ccccc}
\hline 
Dataset & Camera model & No. cameras & Image size & Environment & Date and time \\ \hline
\hline 
1 & Pt. Grey & 2 & $1900 \times 1200$ & indoor & 11/9/2016 10:20 am  \\ \hline
2 & Pt. Grey & 2 & $1900 \times 1200$ & outdoor & 3/23/2017 10:19 am \\ \hline
3 & Pt. Grey & 3 & $1900 \times 1200$ & outdoor & 4/4/2016 1:36pm \\ \hline
4 & Pt. Grey & 2  & $1900 \times 1200$ & outdoor &  4/3/2017 5:13 pm \\ \hline
5 & GoPro HERO Black & 3 & $2705 \times 1520$ & outdoor & 4/13/2017 1:16pm \\ \hline
6 & JAI BB-500 GE & 1 & $1600 \times 1200$ & outdoor & 2/4/2015 11:20 am \\ \hline
\end{tabular}
}
\end{table*}

\subsection{Background composition and construction}
\label{sec:background}

{While the choice of whether to have a one-sided, or over-the-row background strongly depends on the application, we chose to use a range of one-sided background designs, as opposed to over-the-row units used in \cite{botterill2016robot, gongal2016apple, davidson2016proof, silwal2014apple} for a few reasons, most of which revolve around flexibility with data acquisition. The first reason is that we consider acquiring data in a variety of systems, some in orchards we manage, and some in orchards we have never visited.  Over-the-row units need to be built for a specific block spacing and tree height, whereas our mobile backgrounds have a small footprint, and in the worst case will be too short and not capture the tops of trees in images, versus damaging the tops of trees with an over-the-row unit. In addition, some orchard blocks have supports that meet at the top.  Over-the-row units would have trouble navigating these physical obstacles.  Finally, another advantage of our one-sided background is that we do not require supplementary illumination to acquire data.}

\begin{figure}[ht!]
\centering
\subfloat[Background from Dataset 1: laboratory setup]  
	{  \label{sf:lab_robot}
\includegraphics[trim=0cm 3cm 0cm 0cm, clip=true, width=0.40\linewidth]{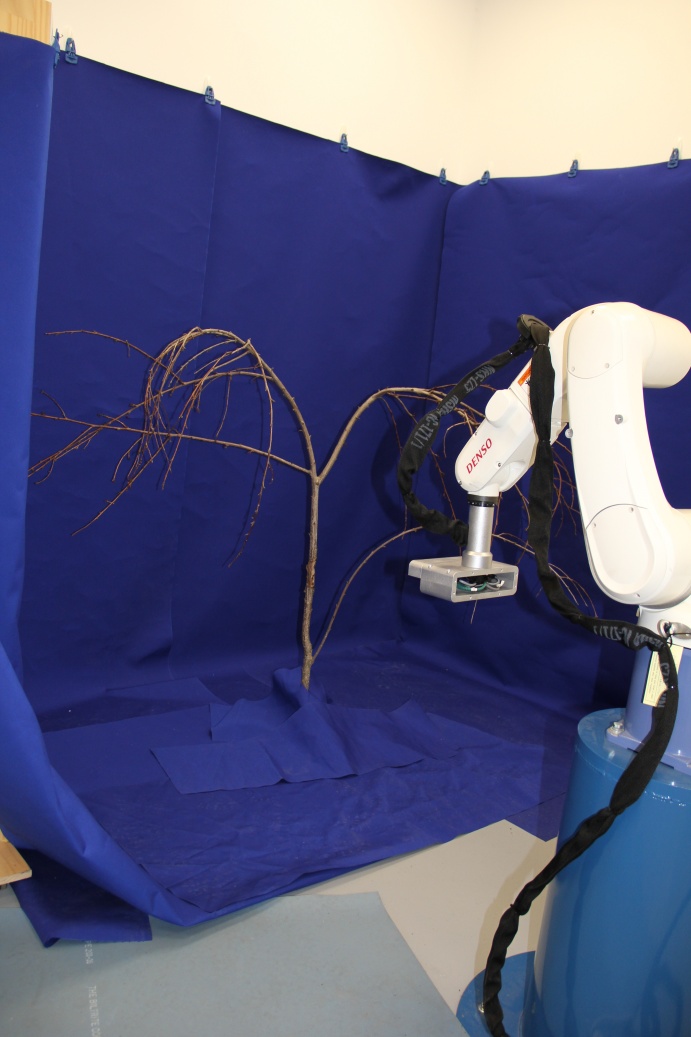}    
}
\subfloat[Background from Dataset 4]  
	{  \label{sf:geneva_phenotyping}
\includegraphics[trim=0cm 3cm 0cm 2cm, clip=true, width=0.40\linewidth]{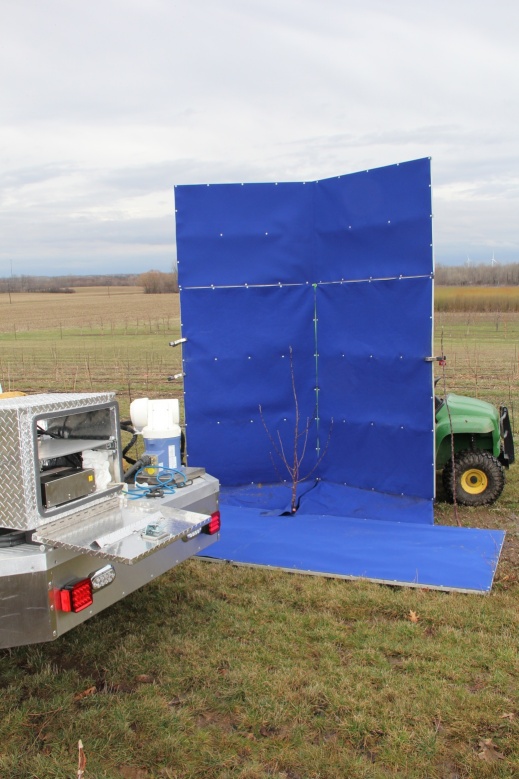}    
}
\\

\subfloat[Background from Dataset 6]  
{	\label{sf:kville_pruning}
\includegraphics[trim=0cm 0cm 4cm 0cm, clip=true, width=0.40\linewidth]{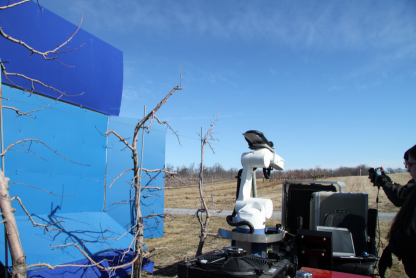}    
}
\subfloat[Background from Dataset 2]  
	{   \label{sf:kville_phenotyping}
\includegraphics[trim=0cm 0cm 5cm 0cm, clip=true, width=0.40\linewidth]{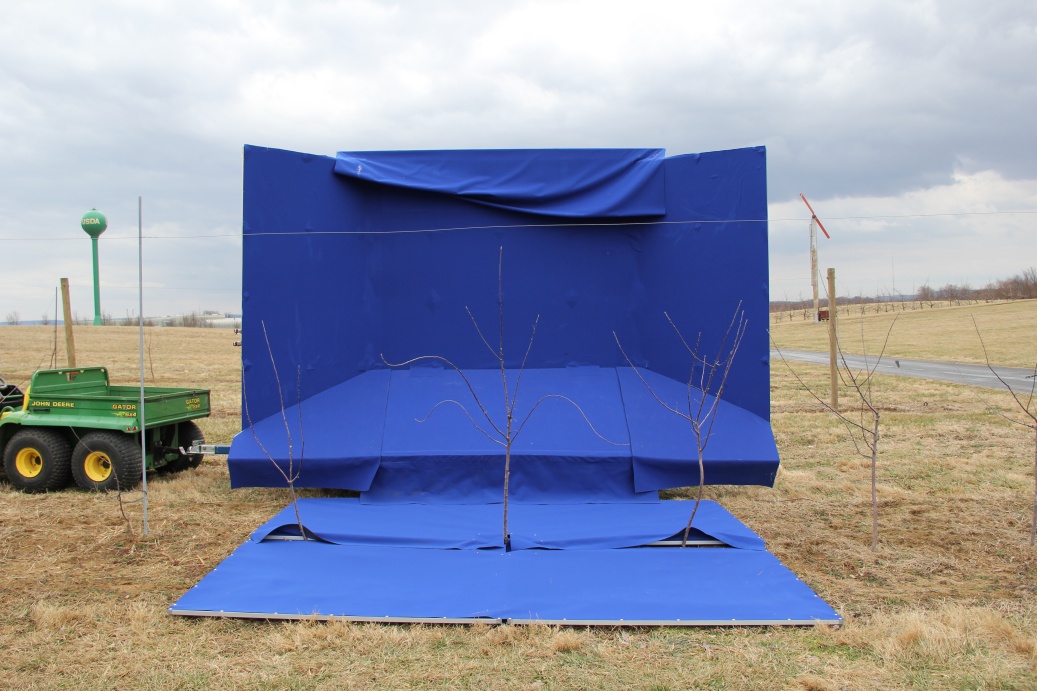}    
}
\caption{\textbf{[Best viewed in color]}{ Different background unit designs used in this paper.  They ranged from hanging blue fabric on laboratory walls and floors \ref{sf:lab_robot}, mounting a frame on a all-terrain vehicle \ref{sf:geneva_phenotyping}, to a frame mounted on a small trailer \ref{sf:kville_pruning} and \ref{sf:kville_phenotyping} .} }
\label{fig:backgrounds}
\end{figure}

{In our experiments, we used a range of designs for the background as illustrated in Figure \ref{fig:backgrounds}.  In general, the only constraint was that the tree regions all be covered by the background, when viewed by the camera.  We have used a range of materials as well, but over time have most frequently used marine heavyweight upholstery material (Sunbrella, Ocean Blue), because these materials are mold and fade resistant. In Figure \ref{sf:lab_robot}, a very simple background is created with hooks on the walls and blue material.  Figure \ref{sf:geneva_phenotyping} shows another one of our background designs, which we constructed for data acquisition at a site at a significant distance from the laboratory for a phenotyping application. The characteristics of that site include windy conditions, narrow trees, and a set of trees that droop on the ground. For these reasons, a v-shaped frame is bolted to an all-terrain vehicle, the material is riveted to the frame, and a ground section was constructed of the same fabric and frame material.}  

{Figure \ref{sf:kville_pruning} shows an early version of our background trailer constructed for the pruning application; a frame was built on the trailer and painted blue, and two wings, a lower piece, and top piece were constructed of similar materials as \ref{sf:geneva_phenotyping}.  The background trailer from \ref{sf:kville_pruning} was used in Figure \ref{sf:kville_phenotyping}, this time for phenotyping, by the addition of a ground section to deal with trees that droop on the ground.  In addition, blue material was glued to the frame so that the color was uniform.} {The background units that have ground panels are only used for phenotyping; while placing the ground panels does require some time it is small.  Since the information returned from the robot vision system is much richer than that of manual measurement, the setup time is not seen as a barrier to using the system within the phenotyping context. }

\subsection{Quantitative analysis}

\begin{table*}[ht]
\centering{}\caption{\label{tab:Quantitative-results}Quantitative results comparing Adaptive Gaussian Threshold (AGT), Adaptive Mean Threshold (AMT), Otsu's method on the hue image, and our proposed approach in precision, recall, and the F-score. Best results are shown in boldface.}
\resizebox{\linewidth}{!}
{
\begin{tabular}{l|cccc|cccc|cccc}
\hline 
 & \multicolumn{4}{c}{Precision} & \multicolumn{4}{c}{Recall} & \multicolumn{4}{c}{F-score}\tabularnewline
\hline 
Image & AGT & AMT & Otsu & Ours & AGT & AMT & Otsu & Ours & AGT & AMT & Otsu & Ours \tabularnewline
\hline
\hline
1 &0.022 &0.025 &0.204 &0.361 &0.505 &0.591 &0.799 &0.739 &0.042 &0.048 &0.325 &0.485 \\ \hline
2 &0.586 &0.513 &0.760 &0.543 &0.614 &0.705 &0.776 &0.959 &0.600 &0.594 &0.768 &0.693 \\ \hline 
3 &0.085 &0.086 &0.687 &0.593 &0.519 &0.596 &0.678 &0.846 &0.146 &0.151 &0.683 &0.697 \\ \hline 
4 &0.615 &0.582 &0.683 &0.665 &0.623 &0.753 &0.886 &0.932 &0.619 &0.656 &0.771 &0.776 \\ \hline 
5 &0.468 &0.461 &0.693 &0.385 &0.419 &0.451 &0.362 &0.903 &0.442 &0.456 &0.476 &0.540 \\ \hline 
6 &0.674 &0.661 &0.728 &0.481 &0.671 &0.739 &0.803 &0.985 &0.673 &0.698 &0.764 &0.646 \\ \hline 
7 &0.894 &0.856 &0.977 &0.960 &0.258 &0.333 &0.684 &0.861 &0.400 &0.480 &0.805 &0.907 \\ \hline 
8 &0.931 &0.889 &1.000 &0.969 &0.262 &0.342 &0.750 &0.902 &0.409 &0.495 &0.857 &0.934 \\ \hline 
9 &0.805 &0.761 &0.934 &0.876 &0.320 &0.405 &0.681 &0.896 &0.458 &0.529 &0.787 &0.886 \\ \hline 
10 &0.913 &0.860 &0.996 &0.971 &0.261 &0.339 &0.695 &0.842 &0.406 &0.487 &0.819 &0.902 \\ \hline 
11 &0.818 &0.775 &0.981 &0.884 &0.305 &0.398 &0.745 &0.906 &0.445 &0.526 &0.847 &0.895 \\ \hline 
12 &0.838 &0.793 &0.963 &0.904 &0.303 &0.392 &0.706 &0.877 &0.445 &0.524 &0.815 &0.890 \\ \hline 
Mean & 0.638	&	0.605 &	\textbf{0.800} &	0.716 &	0.422 &	0.504 &	0.714 &	\textbf{0.887} &	0.424 &	0.470 &	0.726 &	\textbf{0.771}\\ \hline
Median & 0.739 &	0.711 &	\textbf{0.847} &	0.771 &	0.370 &	0.428 &	0.725 &	\textbf{0.899} &	0.443 &	0.510 &	0.779	&	\textbf{0.831}\\ \hline
\end{tabular}
}
\end{table*}

In order to quantify the performance of our approach, we compare it with three existing methods: Adaptive Gaussian Thresholding (AGT), Adaptive Mean Thresholding (AMT) \cite{gonzales1992digital}, and Otsu's method \cite{otsu1979threshold} evaluated on the hue image. For the AGT and AMT methods, the block size was set to 11 and the constant subtraction parameter C was set to 2. Since these comparison methods do not take into account the region outside the background material, which would heavily influence the number of false positives, we also apply our masking operation, Step 5 from Algorithm \ref{alg:overall} to the resulting images.  It is important to emphasize that while Step 5 substantially simplifies the thresholding task for the alternative approaches, it is the greatest contributor to incorrectly segmented pixels in our proposed approach. Finally, it should be noted that no knowledge of the target bounding box is used in the evaluation.

\begin{figure*}[ht]
\centering

\includegraphics[width=0.24\linewidth]{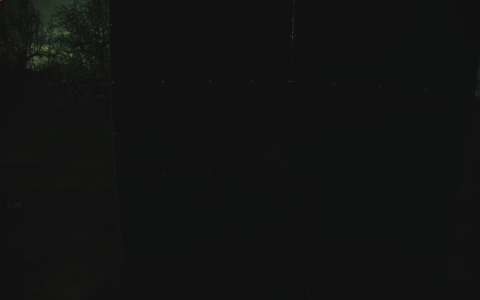} 
\includegraphics[width=0.24\linewidth]{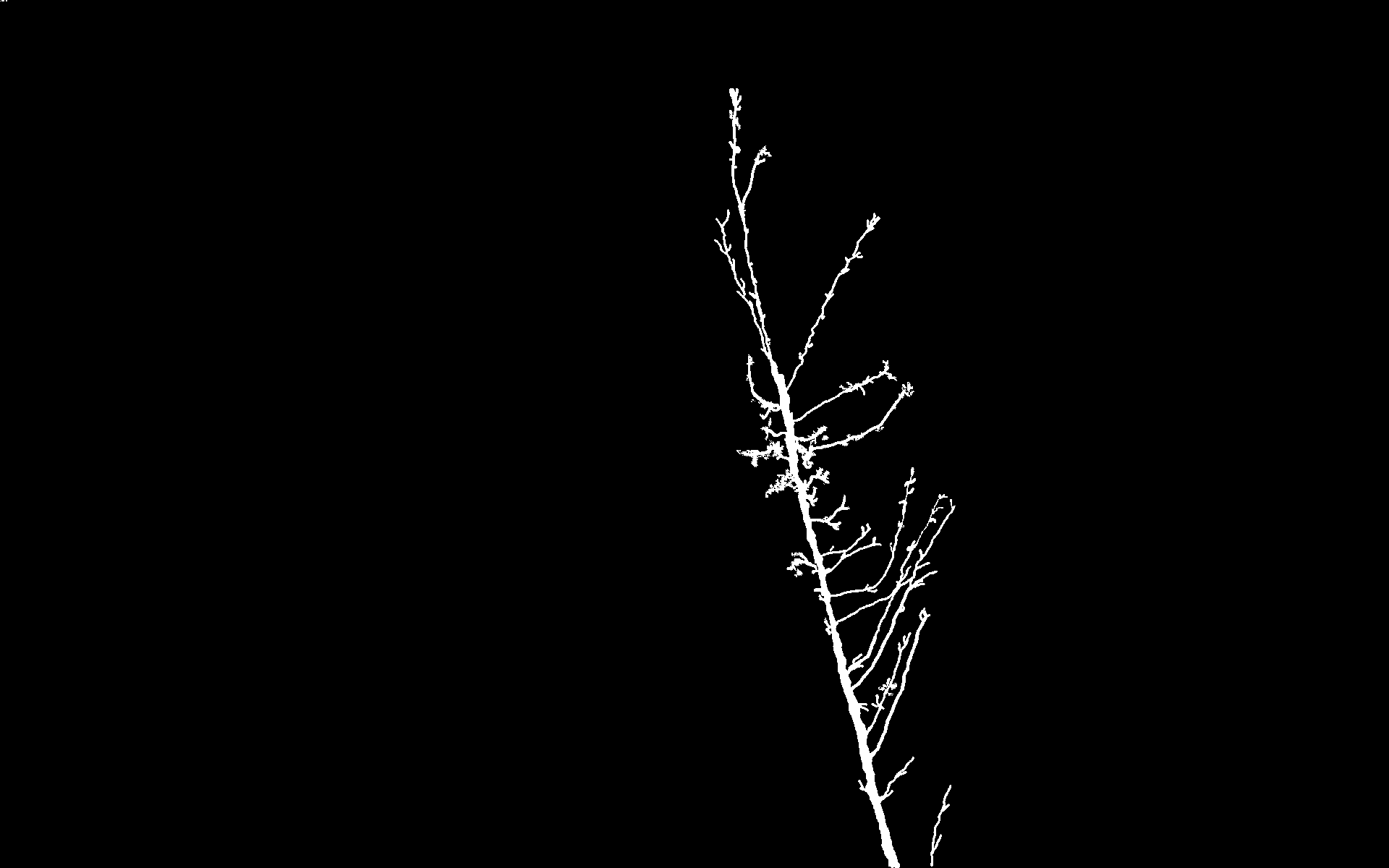}  
\includegraphics[width=0.24\linewidth]{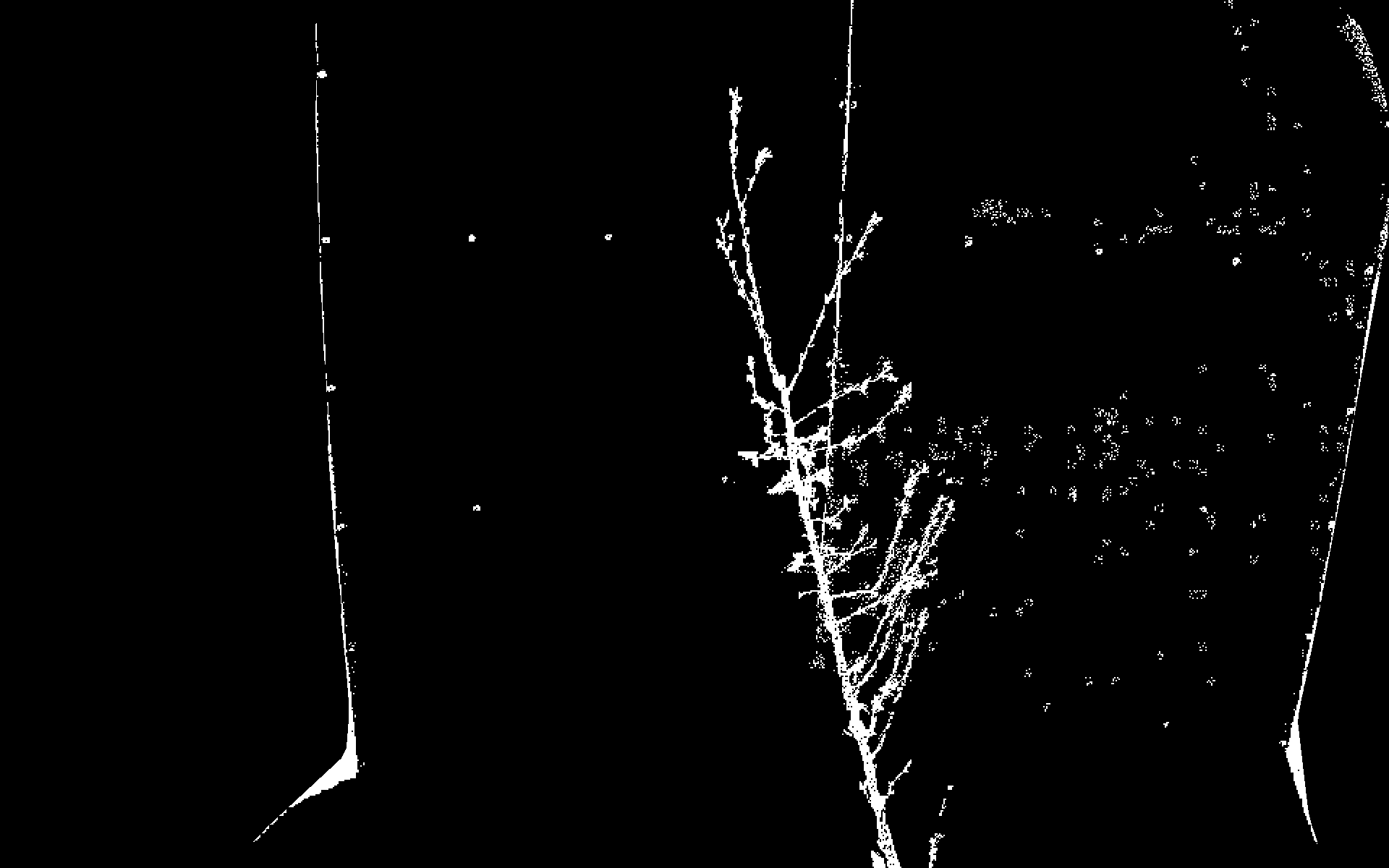} 
\includegraphics[width=0.24\linewidth]{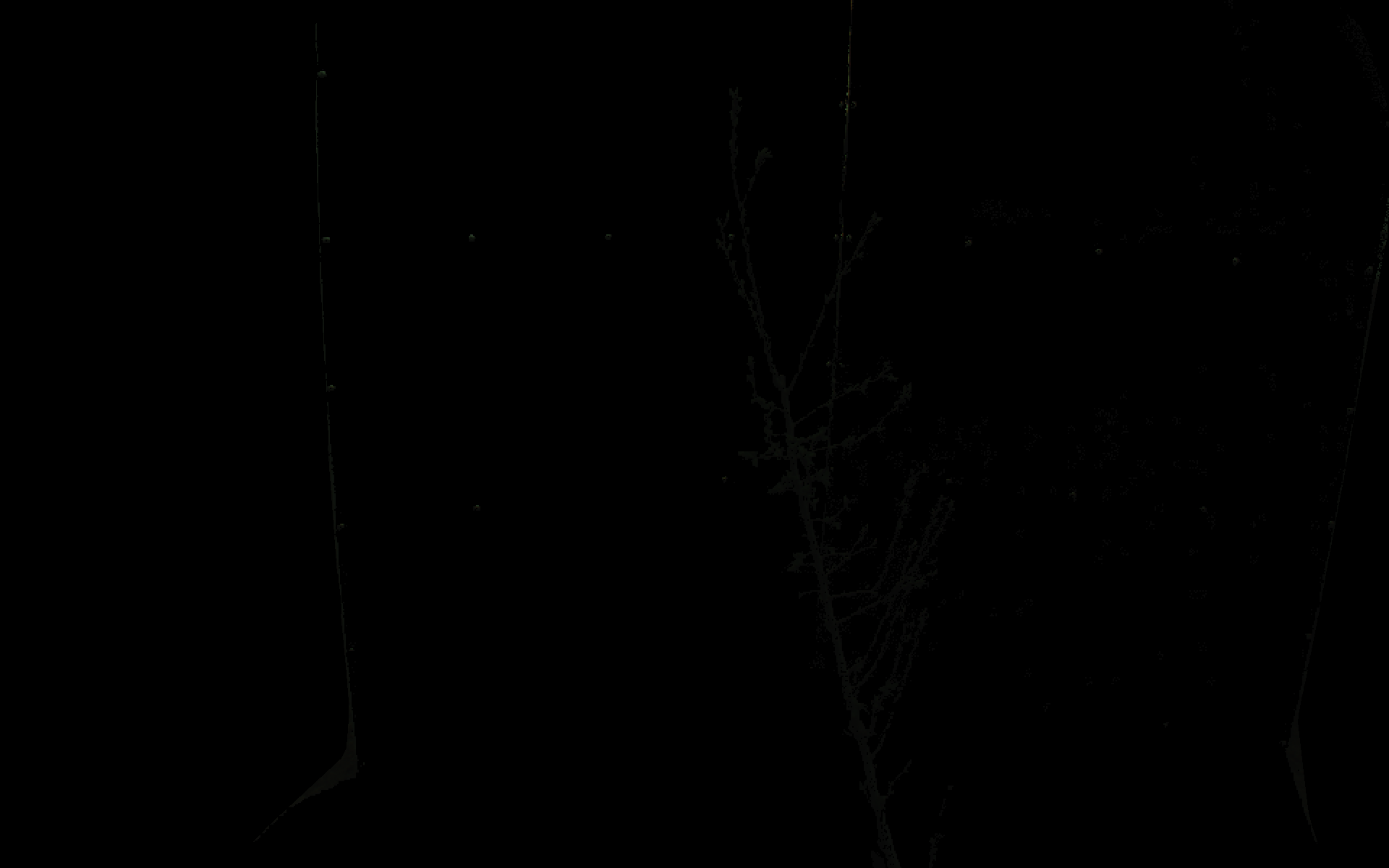}

	\includegraphics[width=0.24\linewidth]{images/Results/raw1_small}    
	\includegraphics[width=0.24\linewidth]{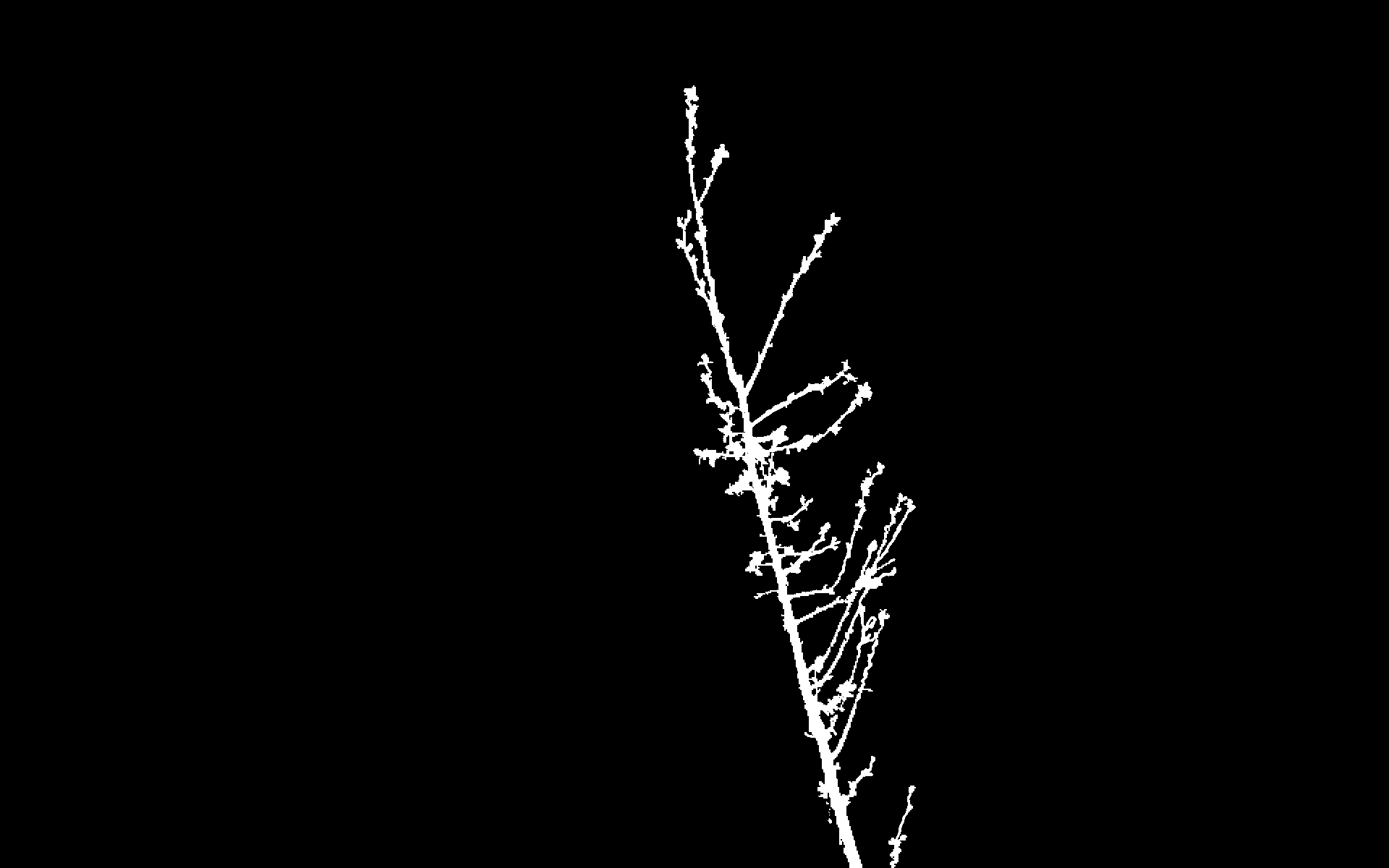}    
	\includegraphics[width=0.24\linewidth]{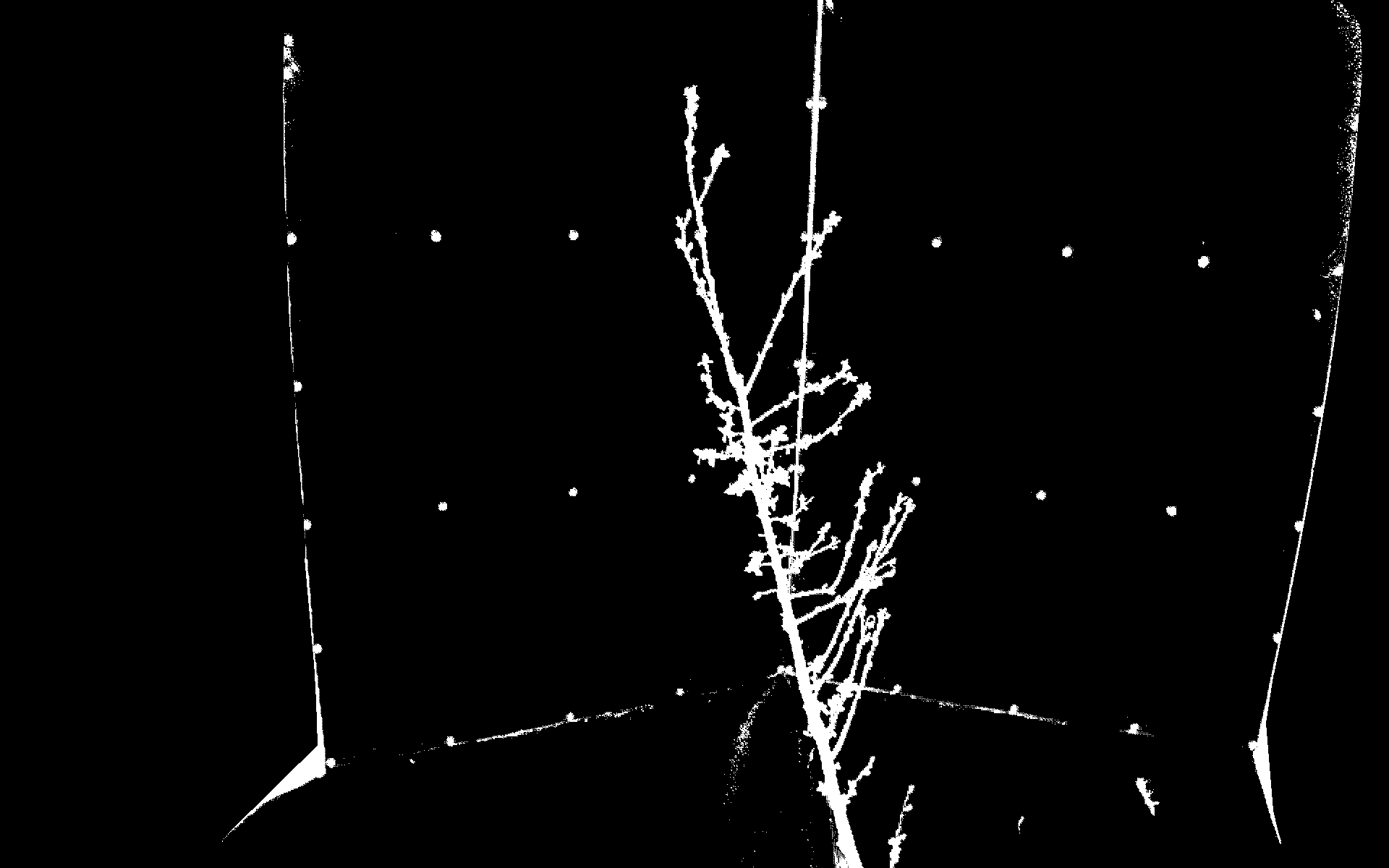}    
	\includegraphics[width=0.24\linewidth]{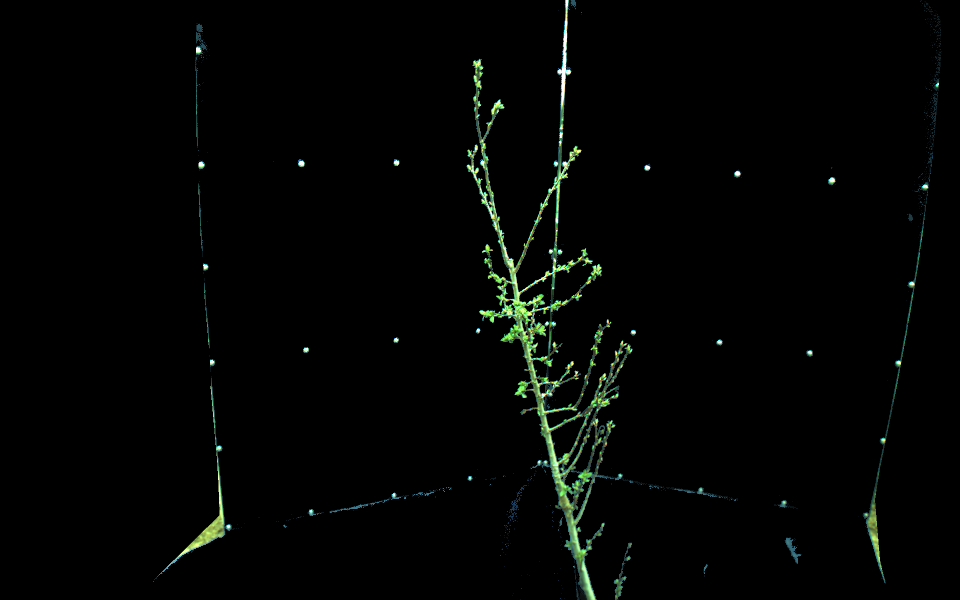}

	\includegraphics[width=0.24\linewidth]{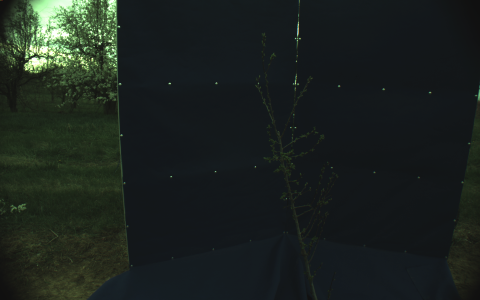}    
	\includegraphics[width=0.24\linewidth]{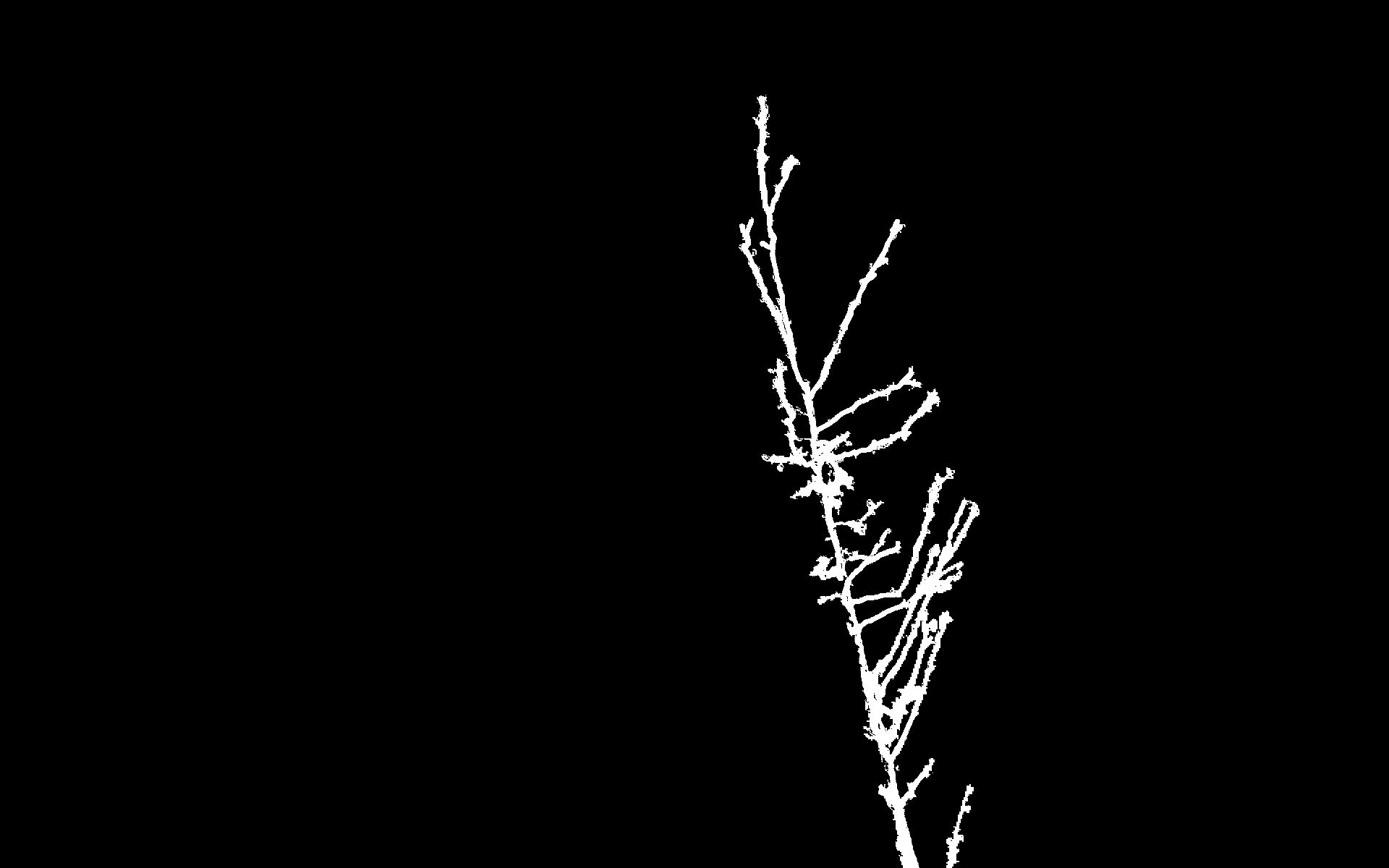}    
	\includegraphics[width=0.24\linewidth]{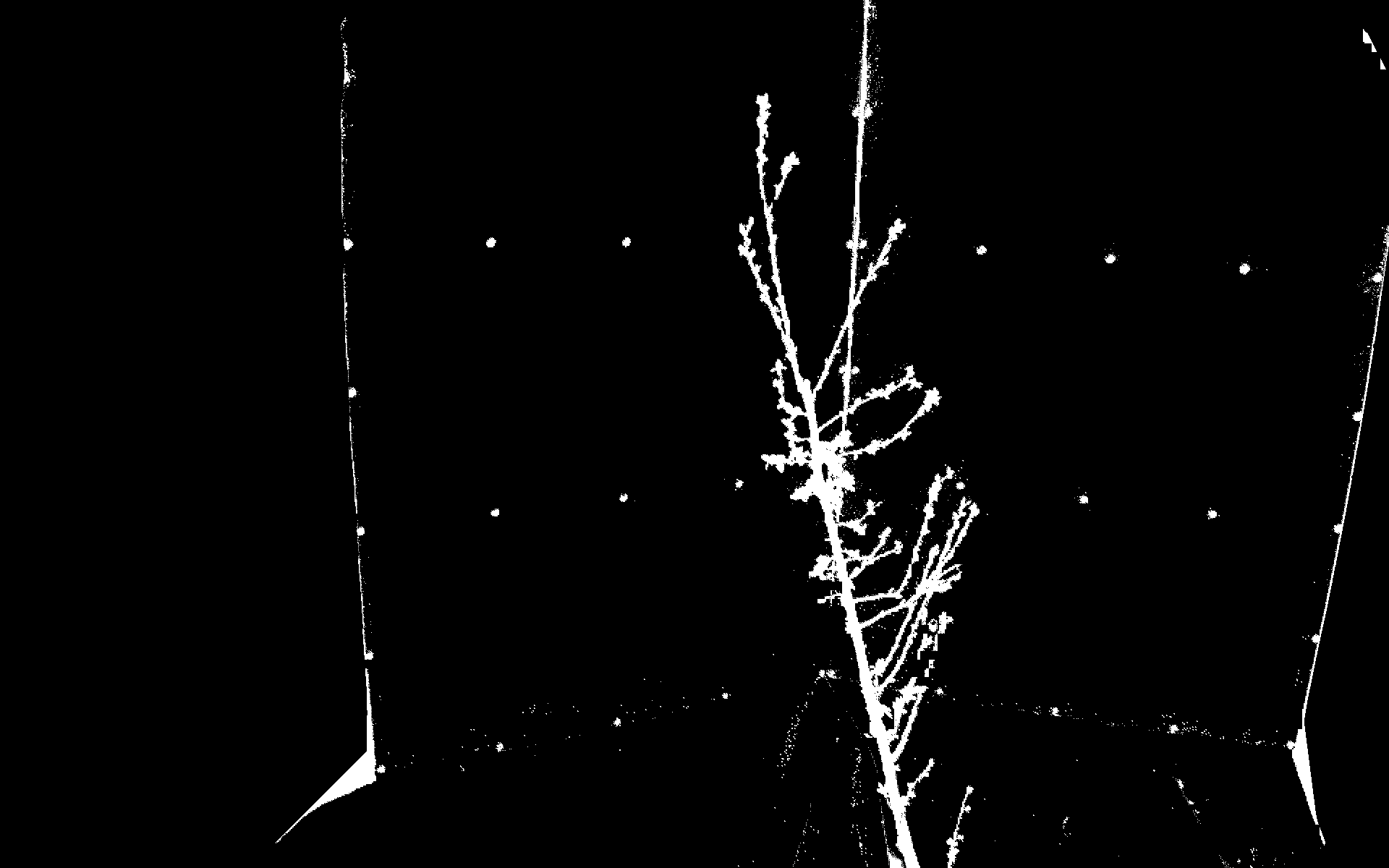}   
	\includegraphics[width=0.24\linewidth]{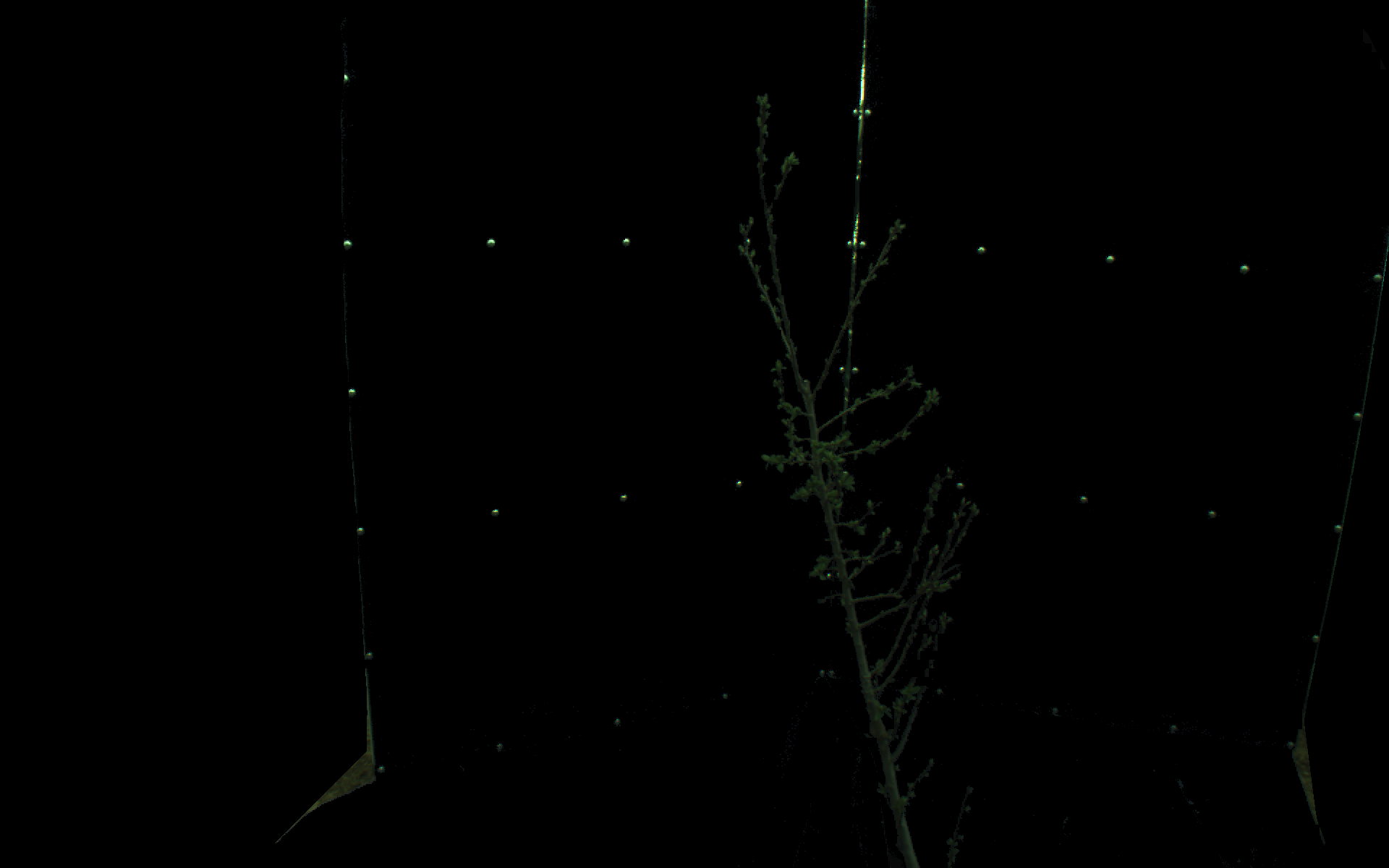}   

	\includegraphics[width=0.24\linewidth]{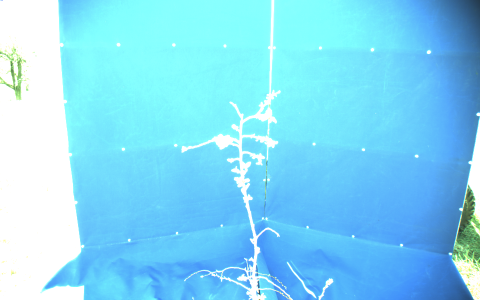}    
	\includegraphics[width=0.24\linewidth]{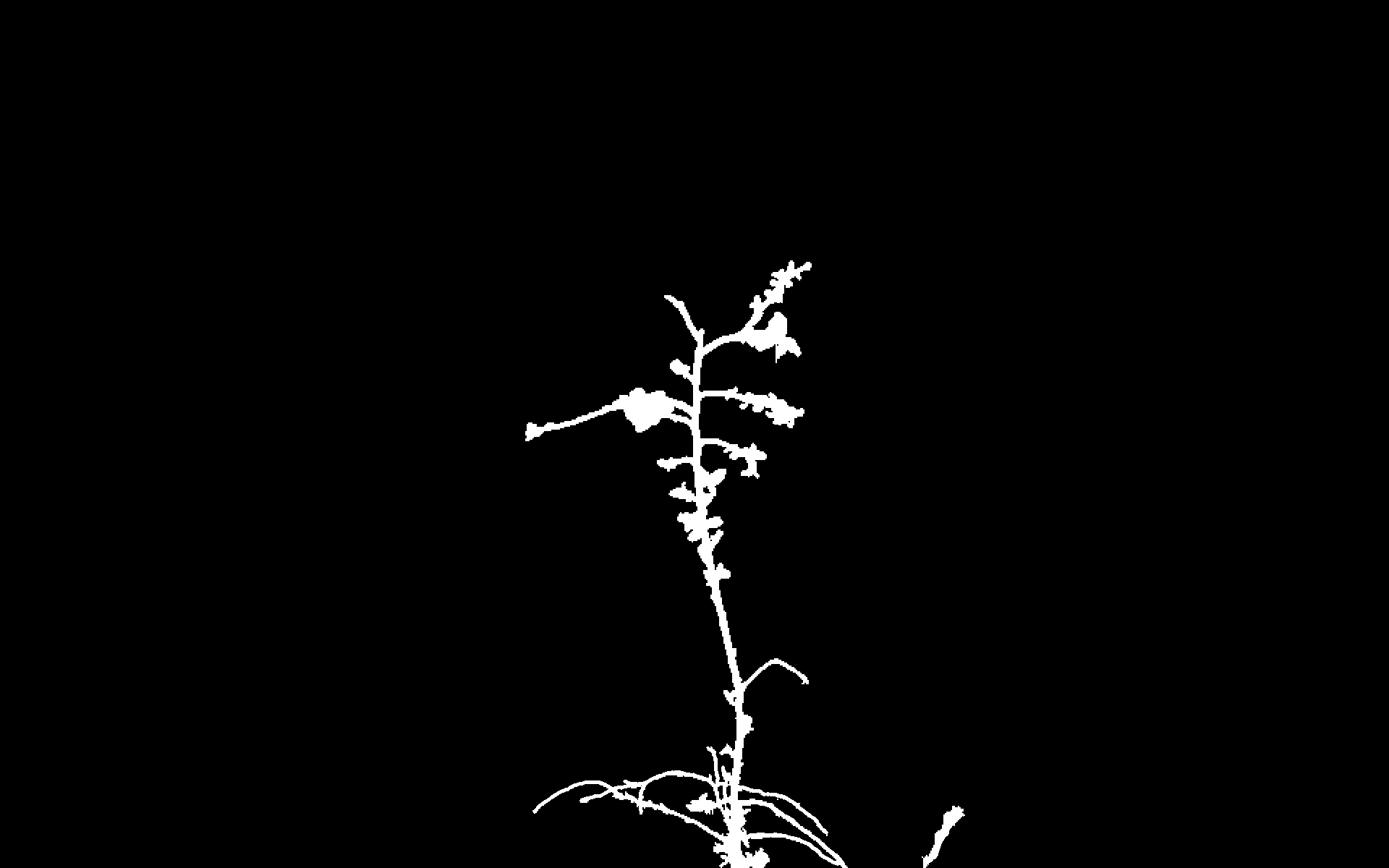}    
	\includegraphics[width=0.24\linewidth]{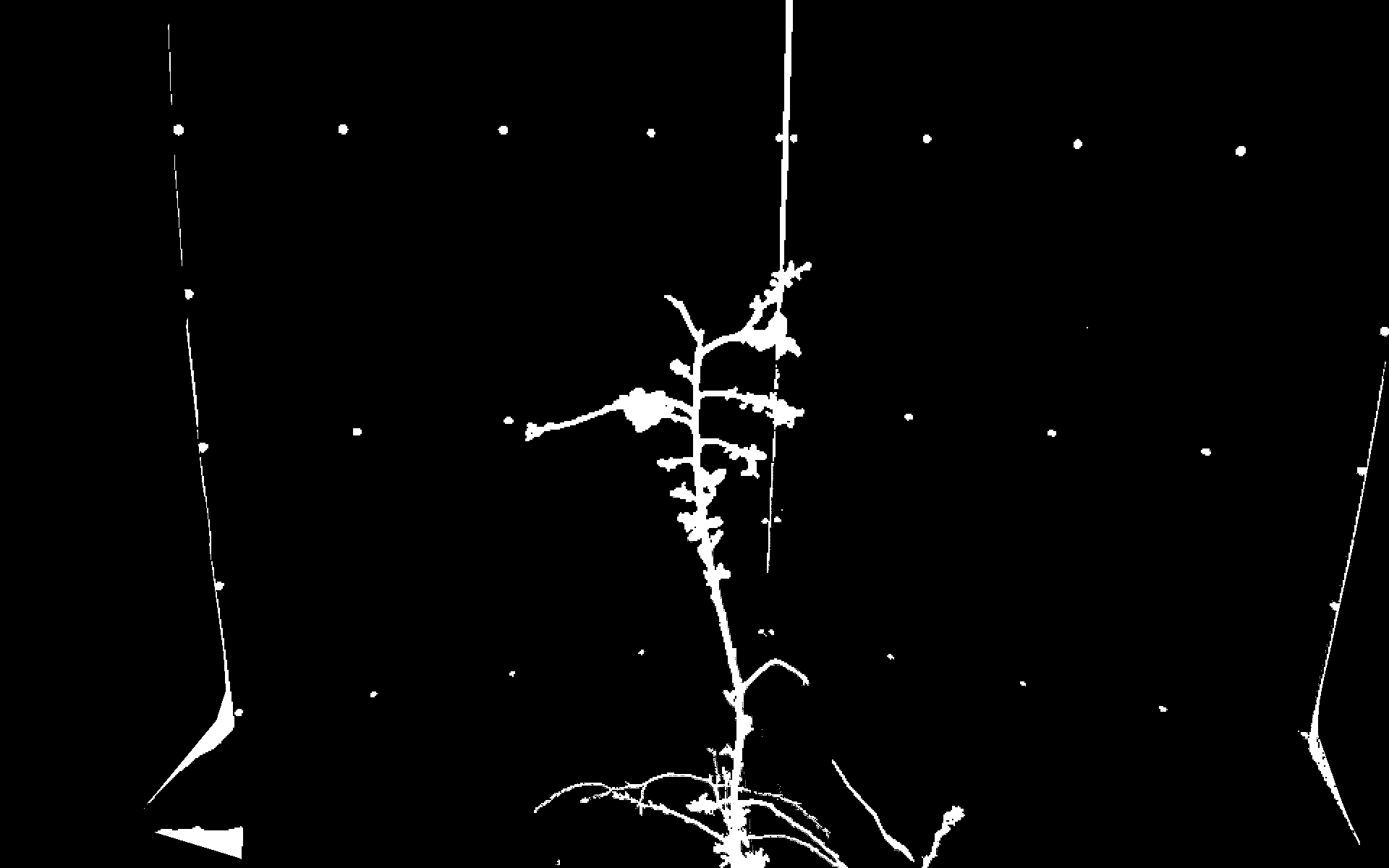}    
	\includegraphics[width=0.24\linewidth]{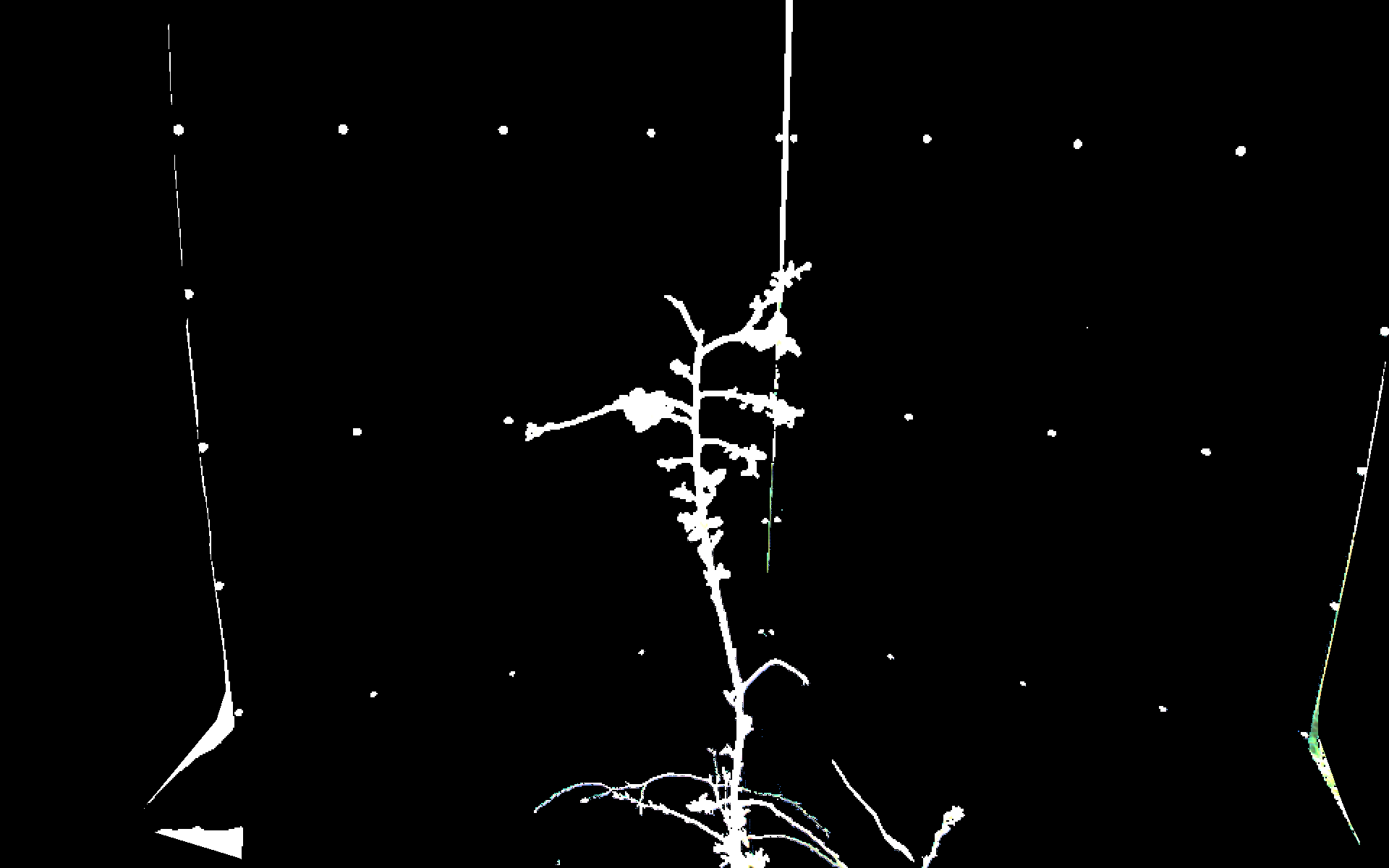}    

	\includegraphics[width=0.24\linewidth]{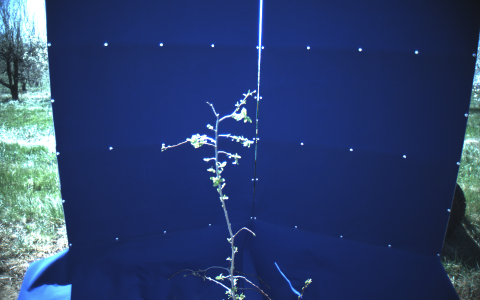}    
	\includegraphics[width=0.24\linewidth]{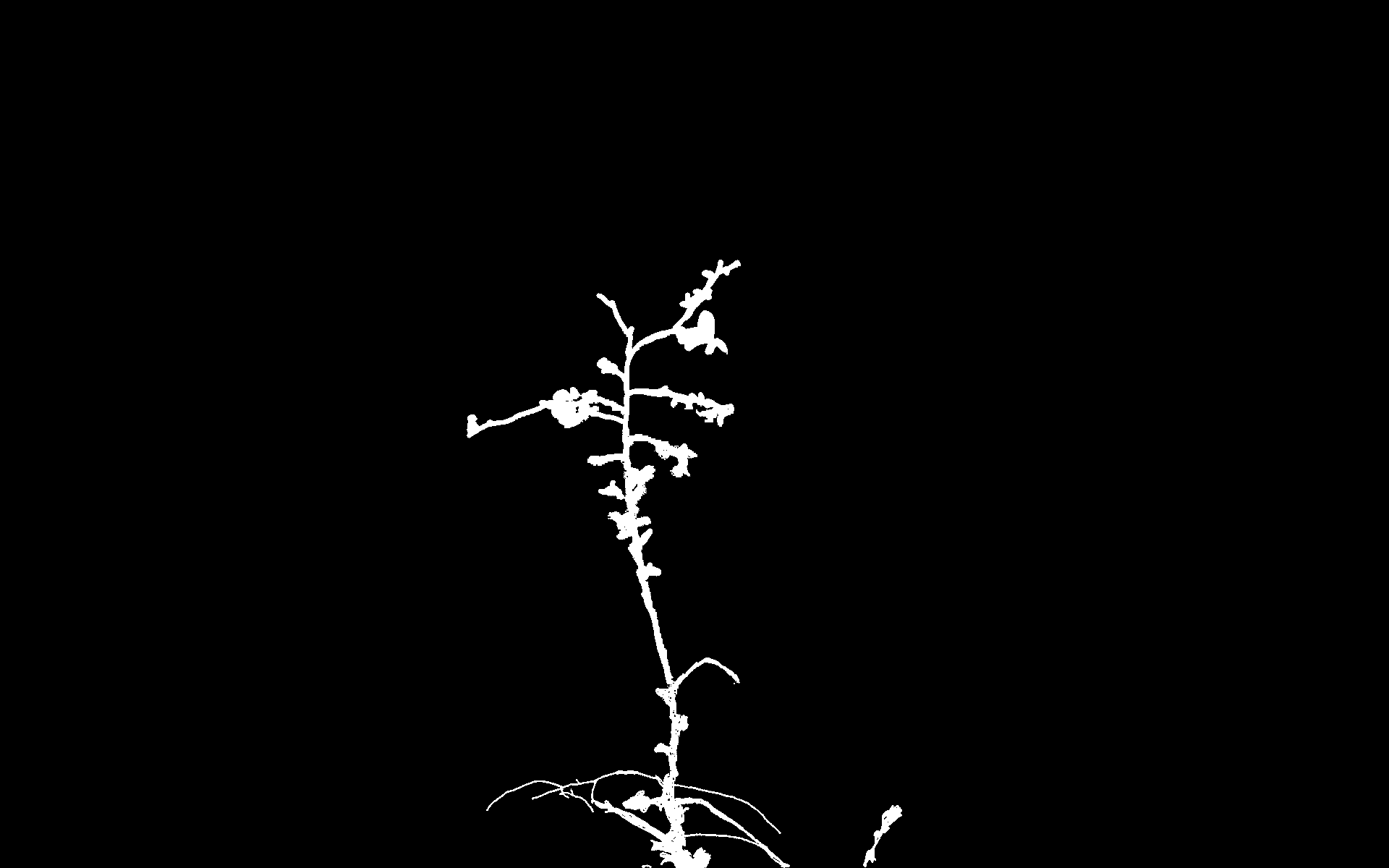}   
	\includegraphics[width=0.24\linewidth]{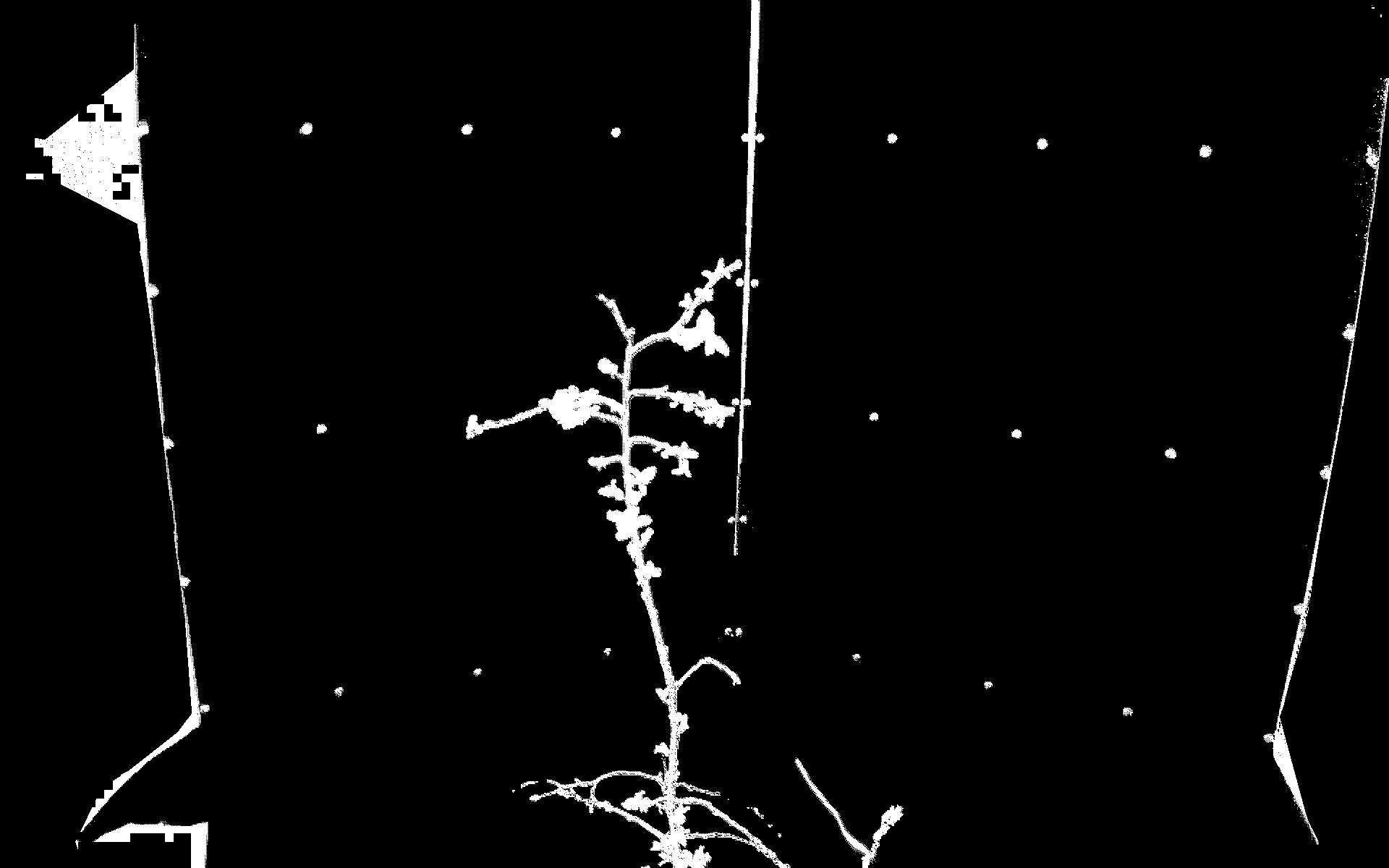}   
	\includegraphics[width=0.24\linewidth]{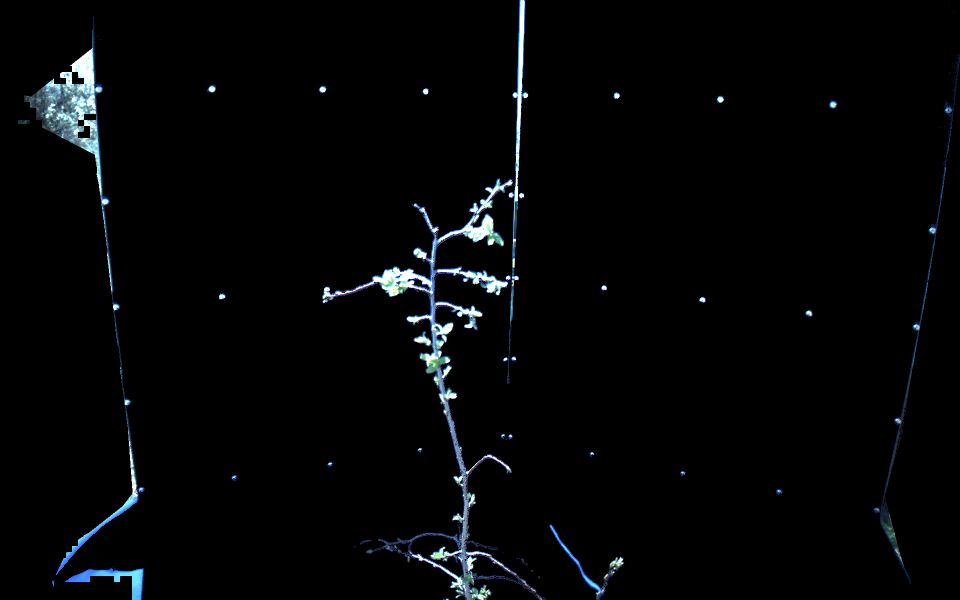}    

	\includegraphics[width=0.24\linewidth]{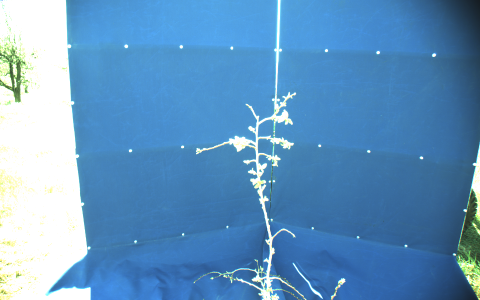}   
	\includegraphics[width=0.24\linewidth]{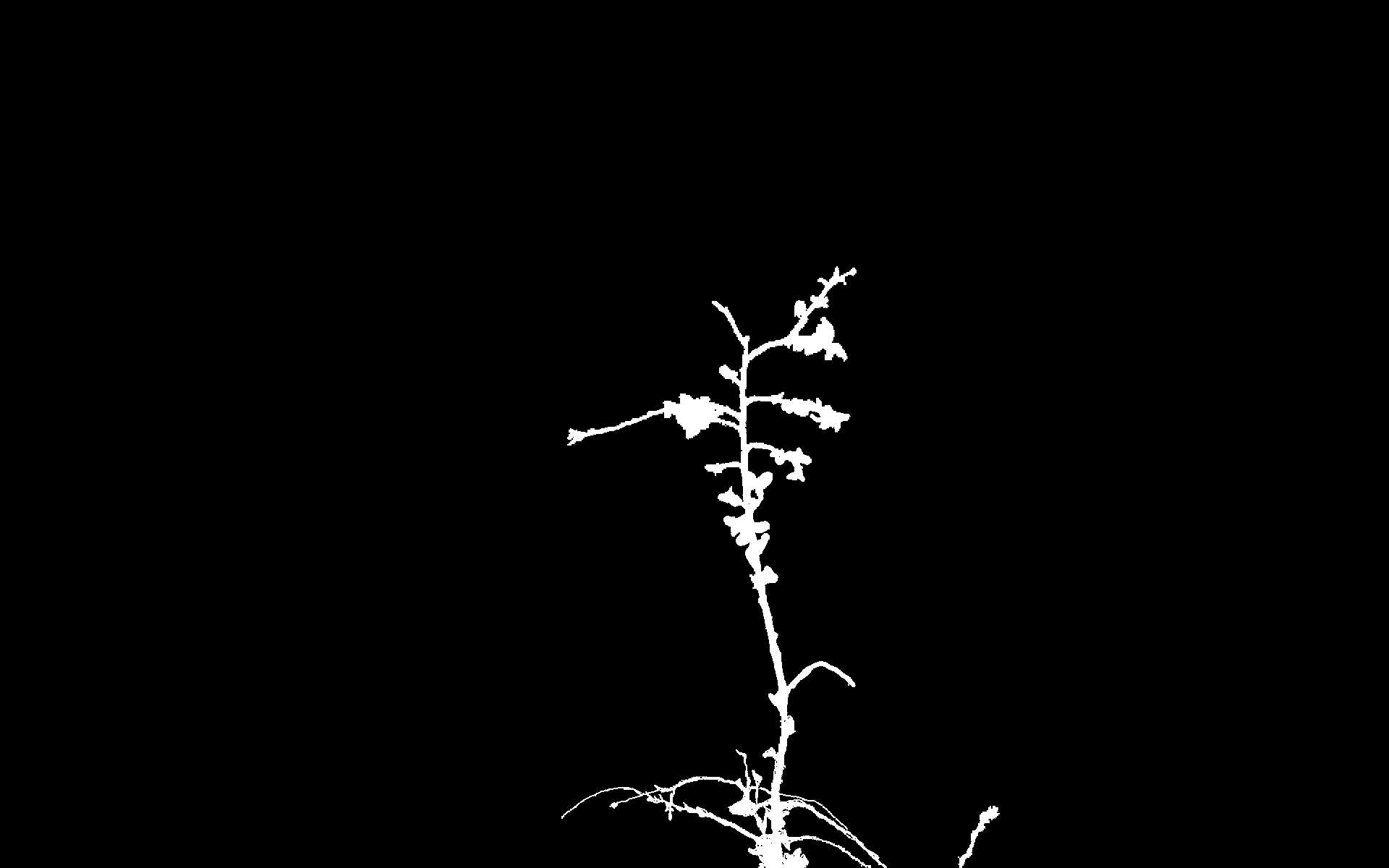}   
	\includegraphics[width=0.24\linewidth]{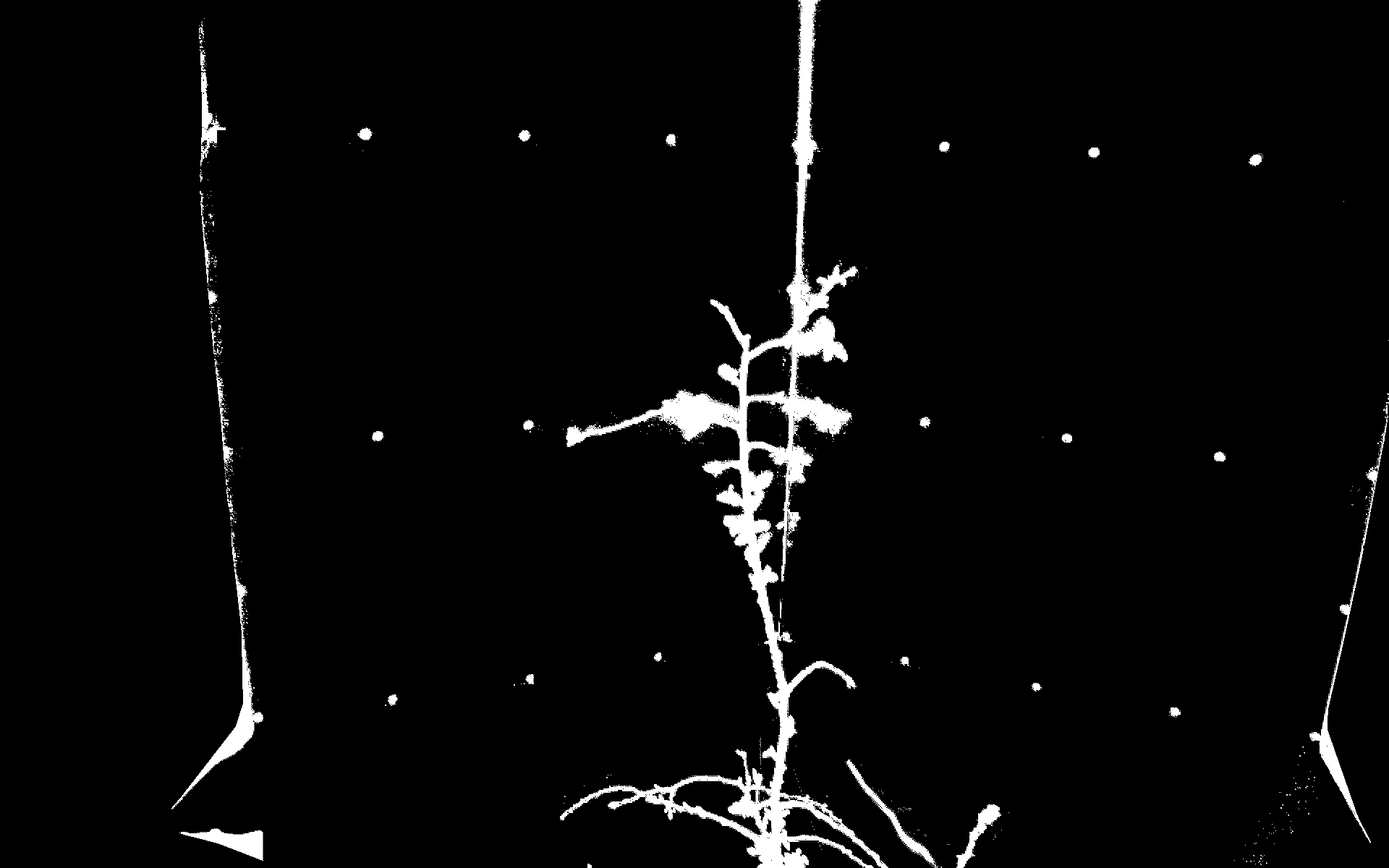}    
	\includegraphics[width=0.24\linewidth]{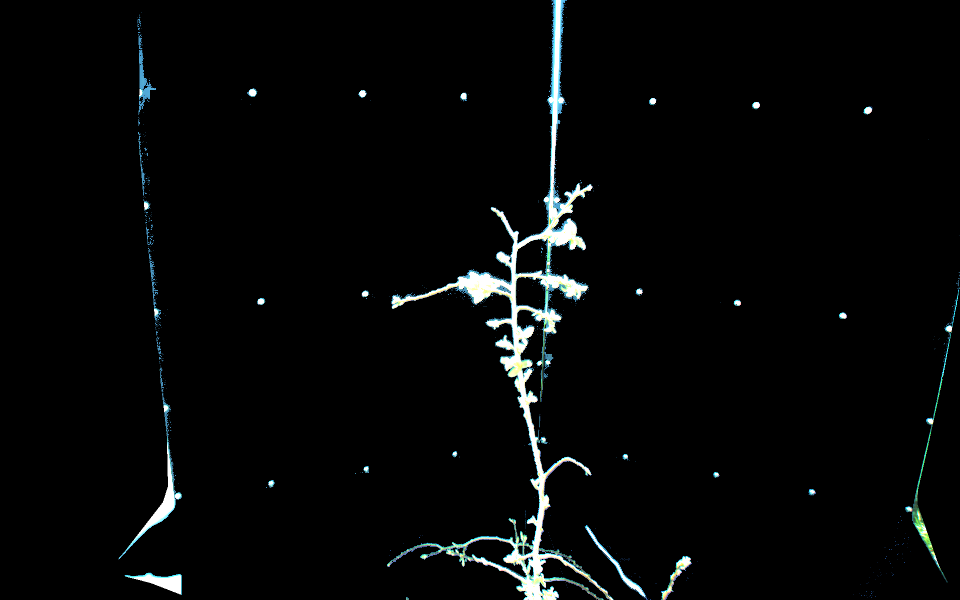}    

\caption{\textbf{[Best viewed in color]} Images of trees in front of a low-texture blue background with varying illumination levels (first column), hand-labeled ground truth (second column), the corresponding segmentations obtained using the proposed approach (third column), and the segmentation result used to mask the original image (fourth column), for the first half of the quantitative dataset (images 1-6).}
\label{fig:examples1}
\end{figure*}

\begin{figure*}[ht]
\centering
	\includegraphics[width=0.24\linewidth]{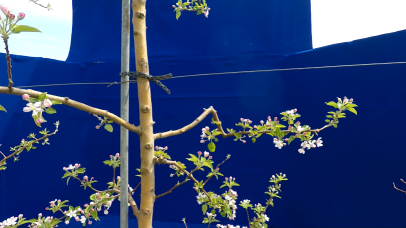}   
	\includegraphics[width=0.24\linewidth]{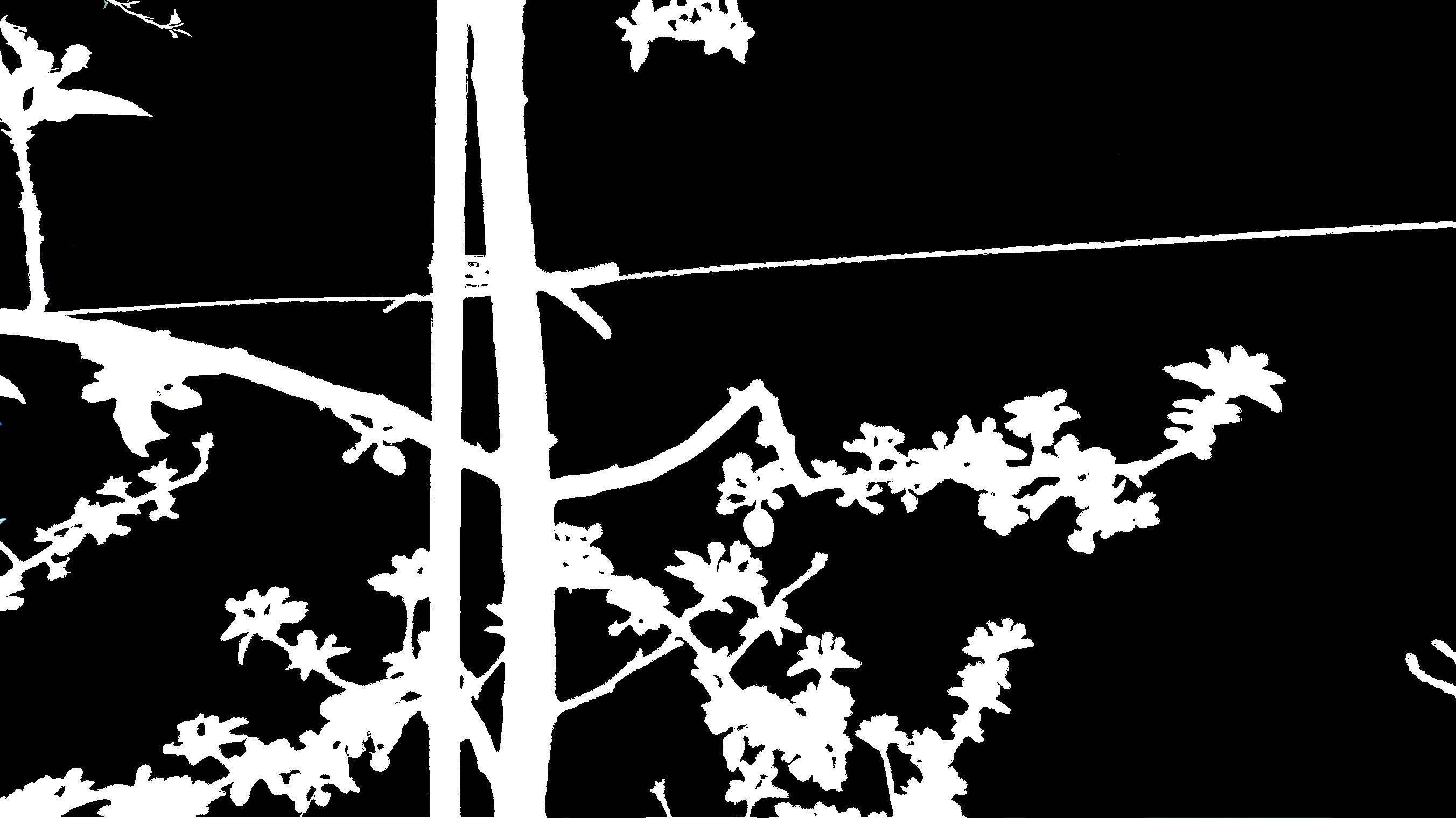}   
	\includegraphics[width=0.24\linewidth]{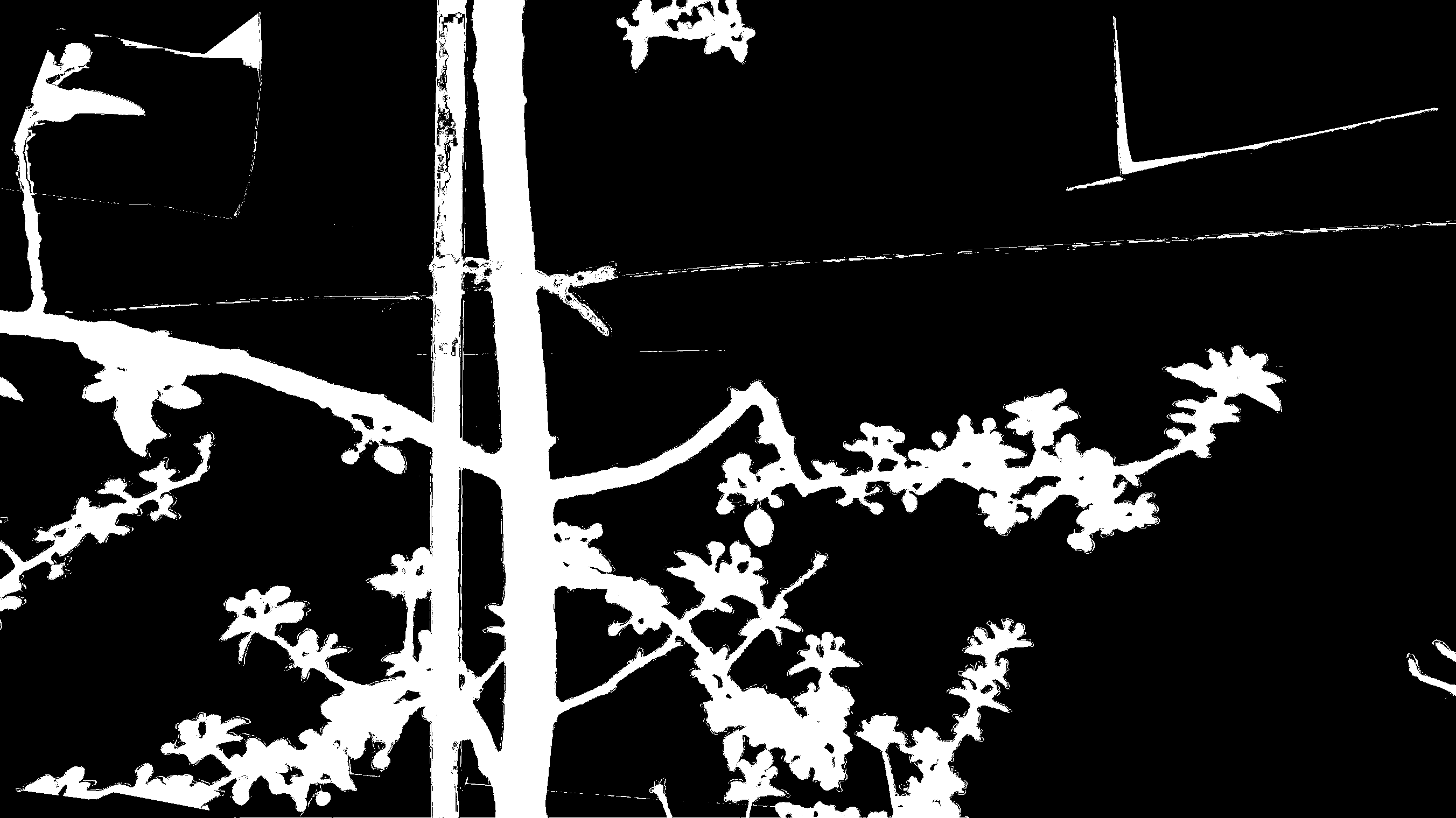}   
	\includegraphics[width=0.24\linewidth]{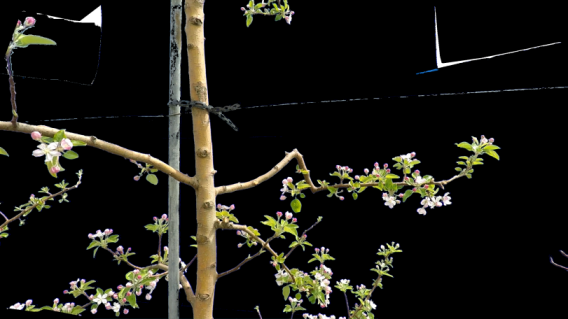}    

	\includegraphics[width=0.24\linewidth]{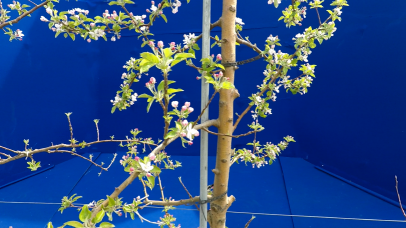}   
	\includegraphics[width=0.24\linewidth]{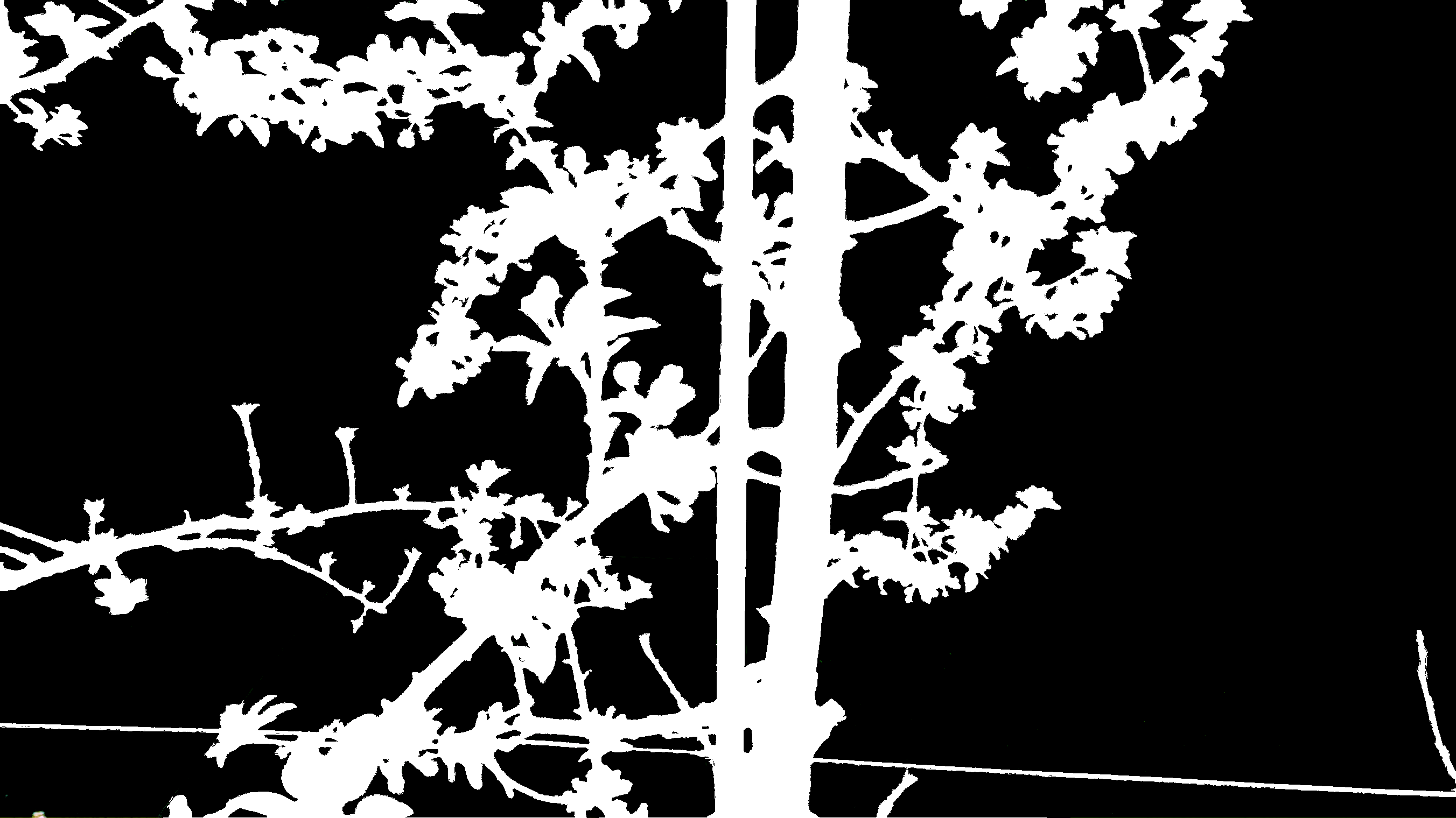}   
	\includegraphics[width=0.24\linewidth]{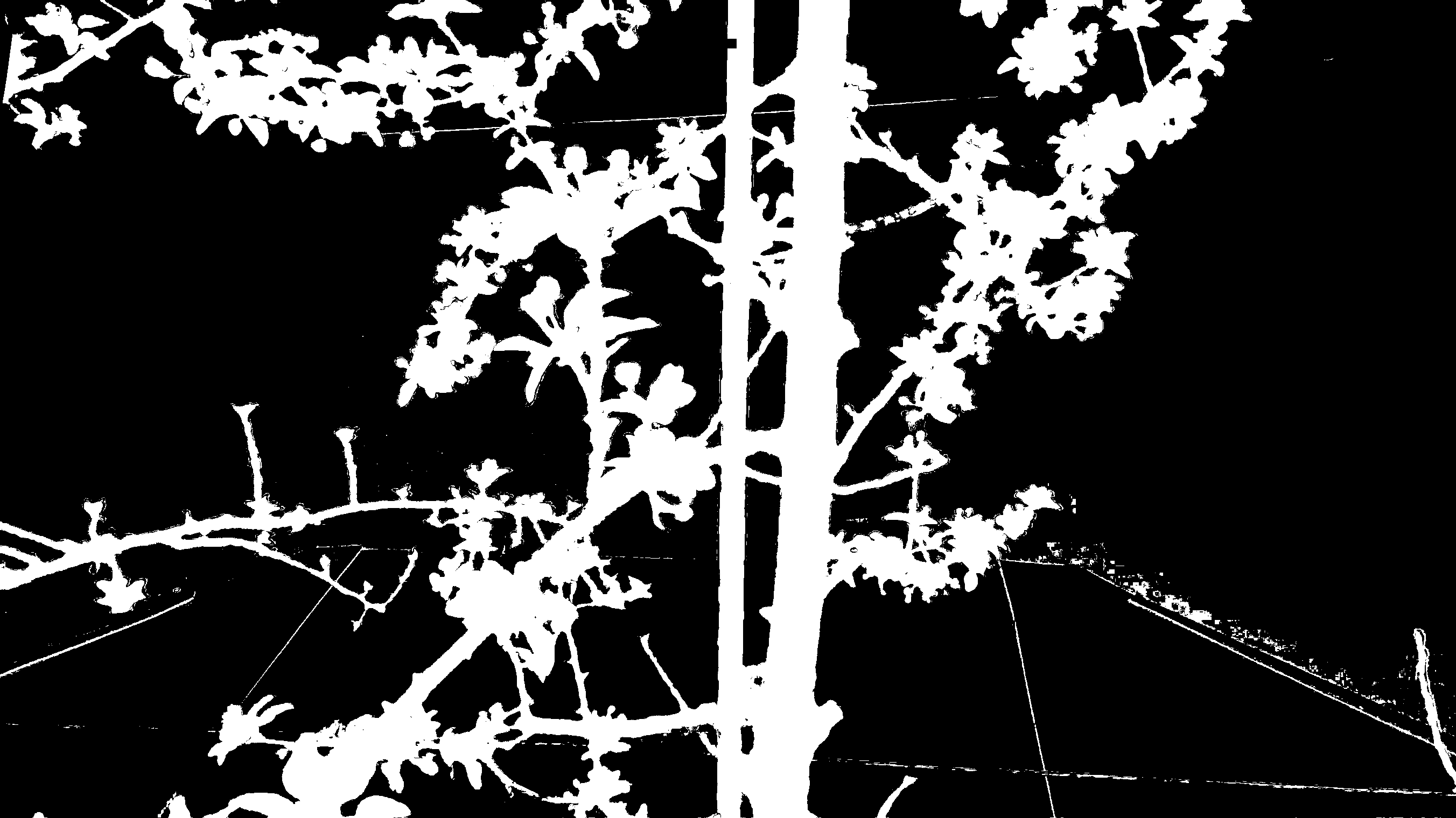}   
	\includegraphics[width=0.24\linewidth]{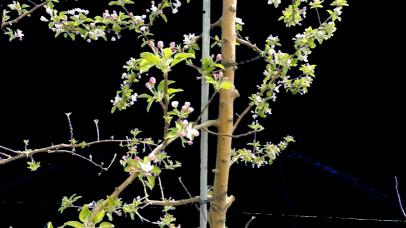}   %

	\includegraphics[width=0.24\linewidth]{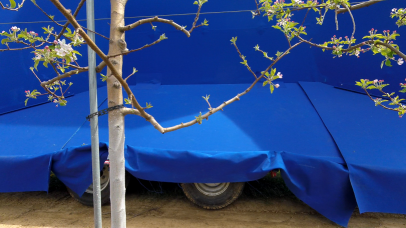}  
	\includegraphics[width=0.24\linewidth]{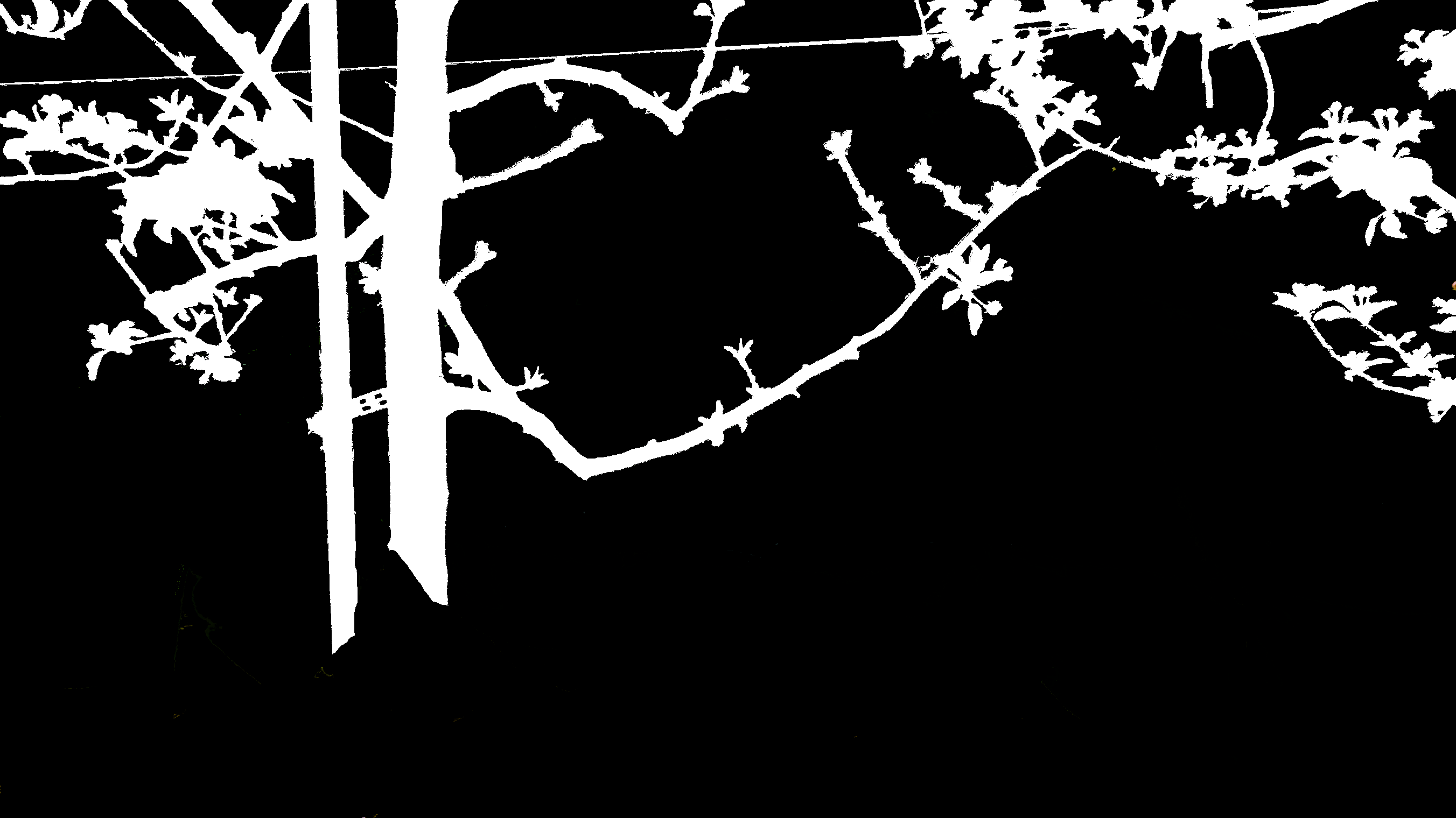}   
	\includegraphics[width=0.24\linewidth]{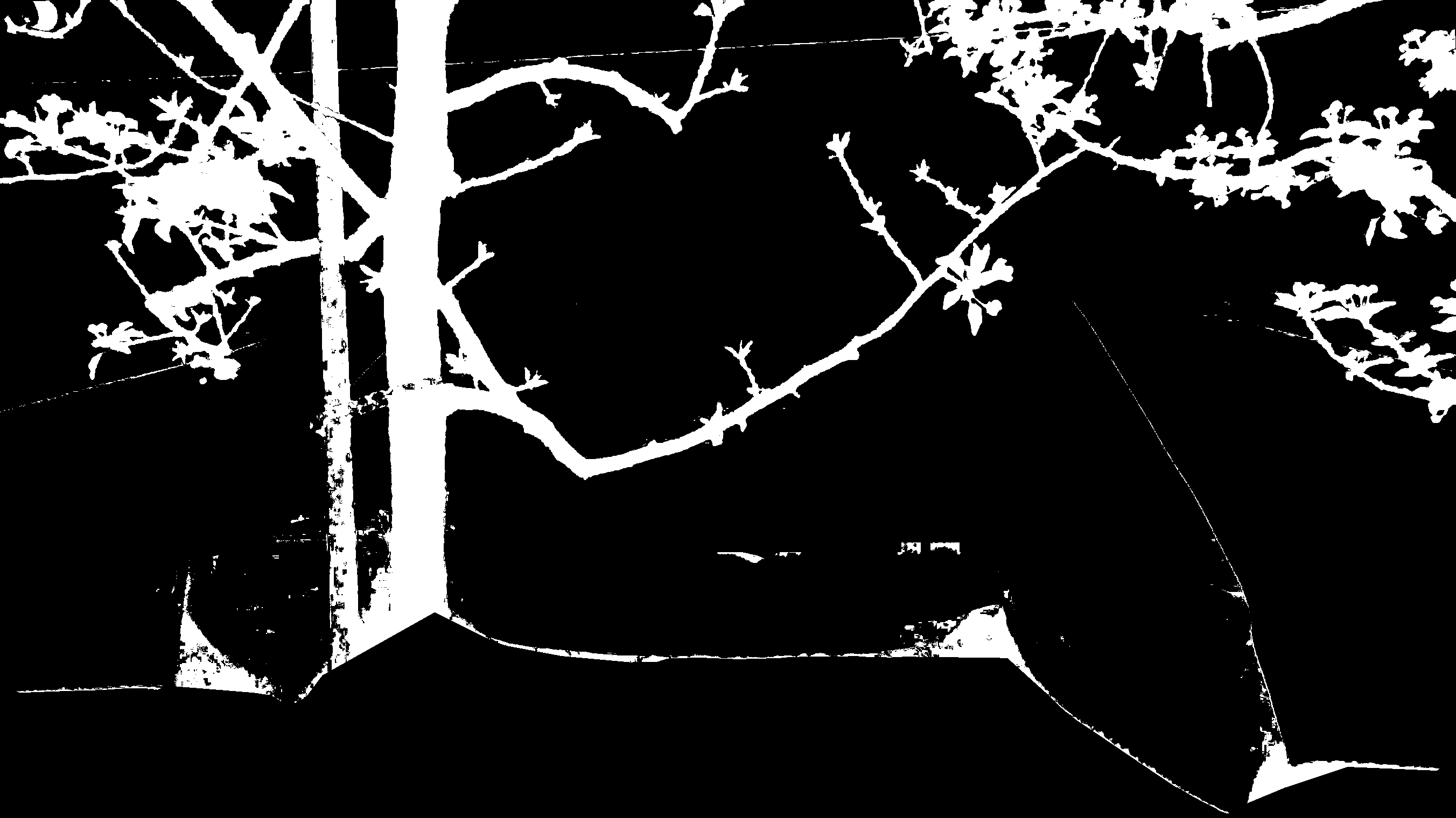}  
	\includegraphics[width=0.24\linewidth]{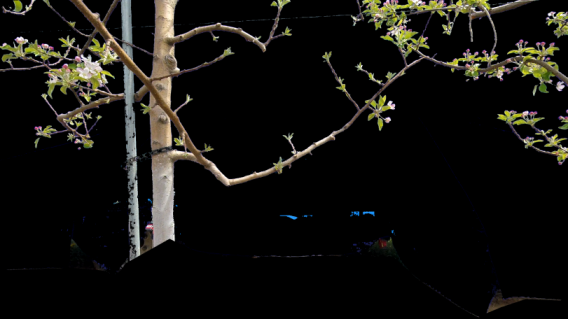}   

	\includegraphics[width=0.24\linewidth]{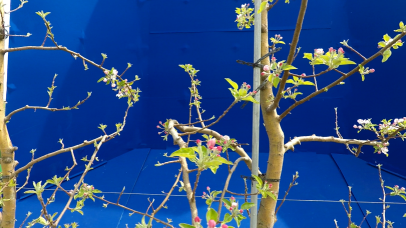}   
	\includegraphics[width=0.24\linewidth]{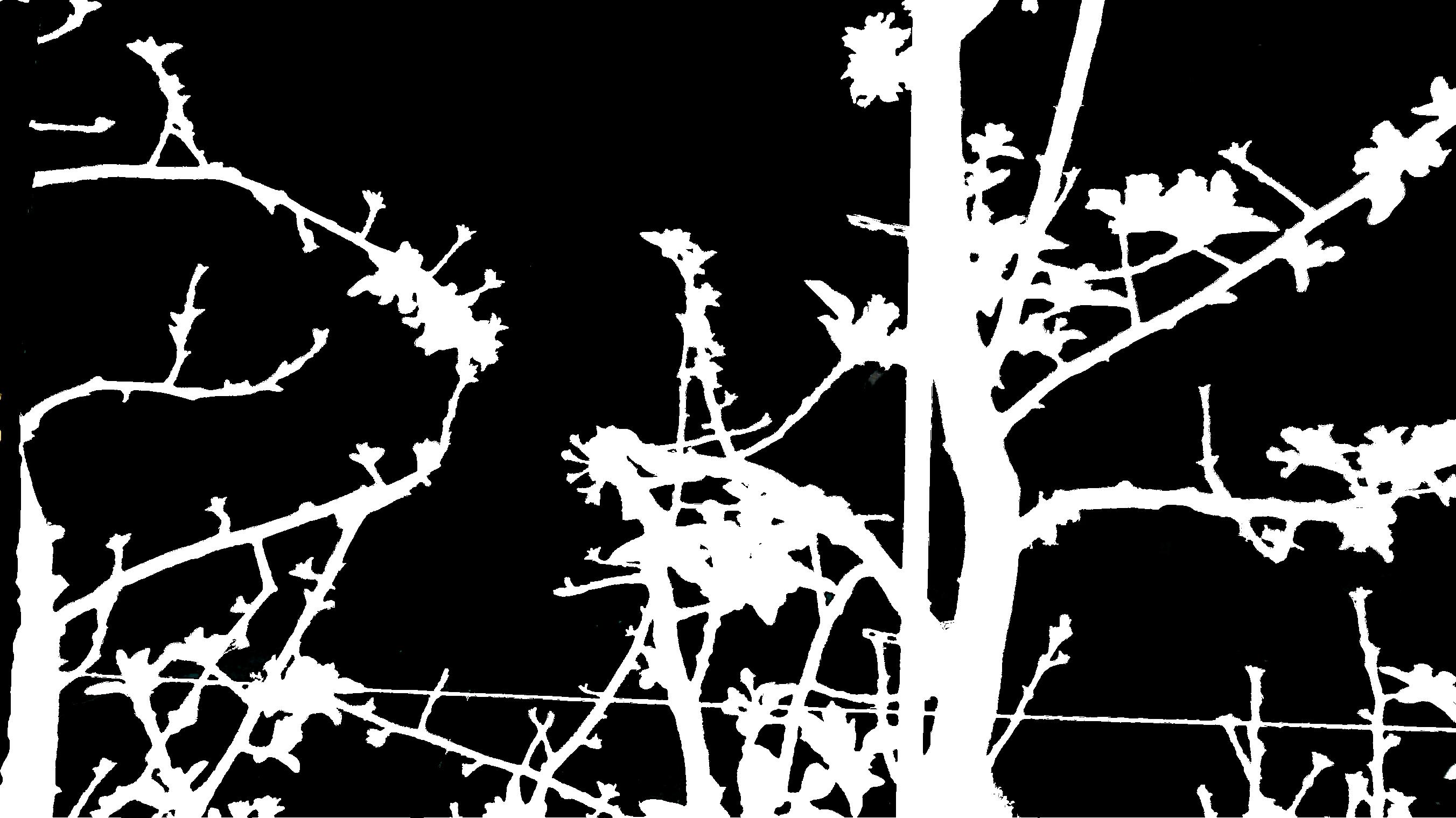}   
	\includegraphics[width=0.24\linewidth]{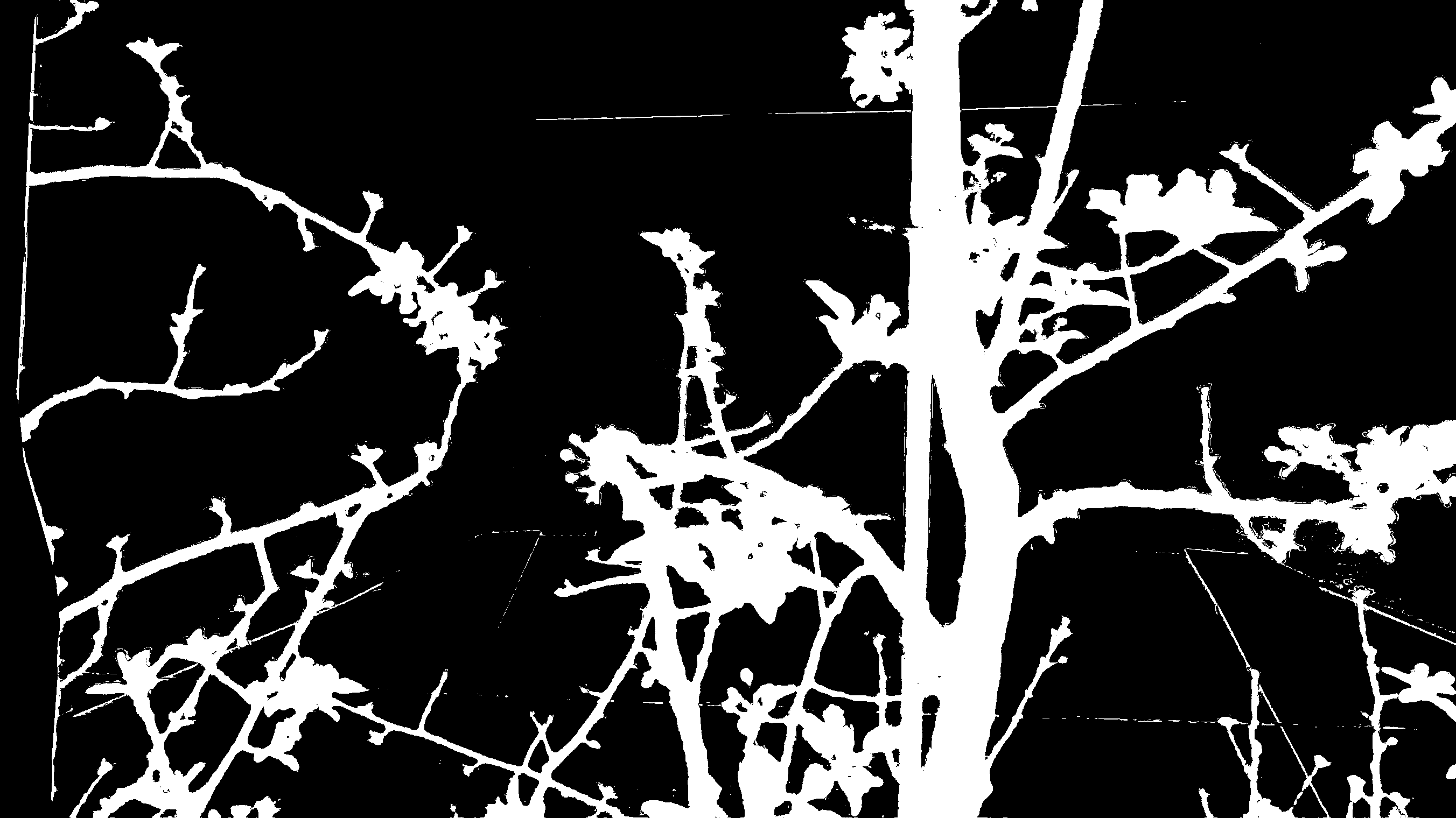}   
	\includegraphics[width=0.24\linewidth]{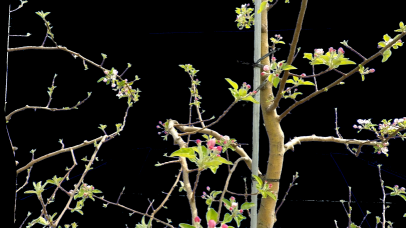}   

	\includegraphics[width=0.24\linewidth]{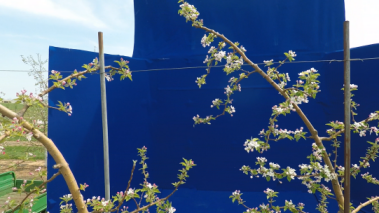}   
	\includegraphics[width=0.24\linewidth]{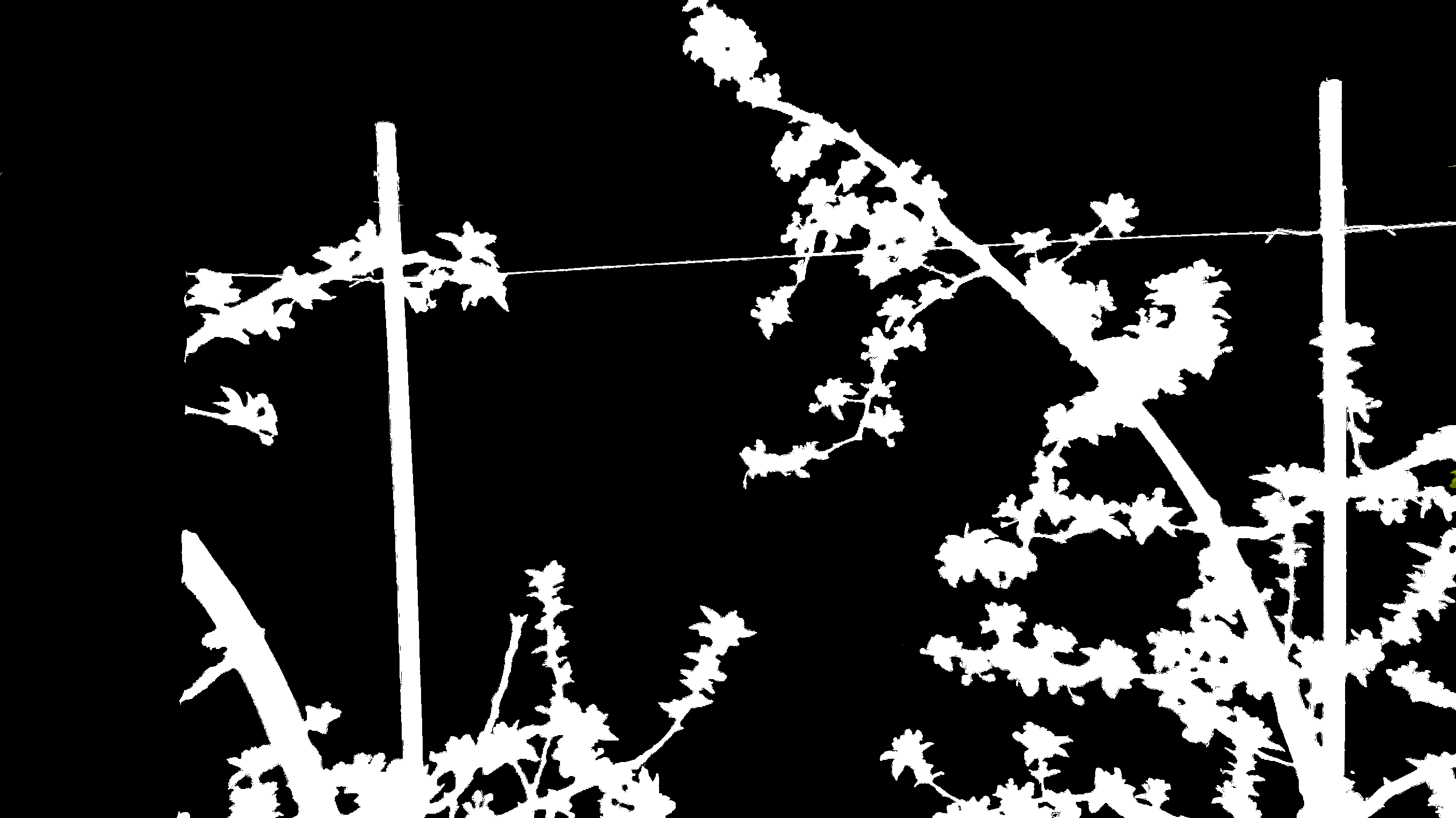}   
	\includegraphics[width=0.24\linewidth]{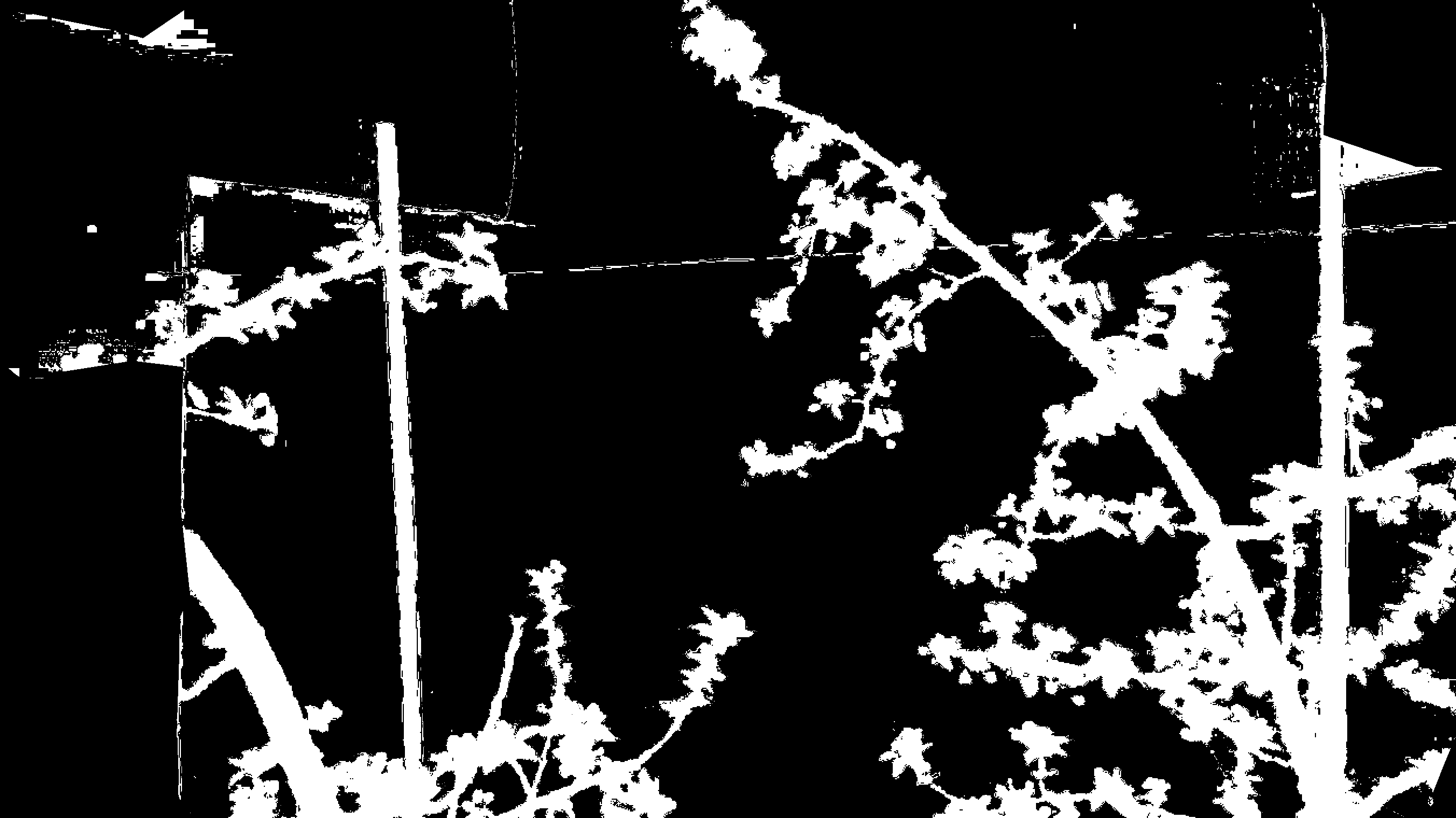}   
	\includegraphics[width=0.24\linewidth]{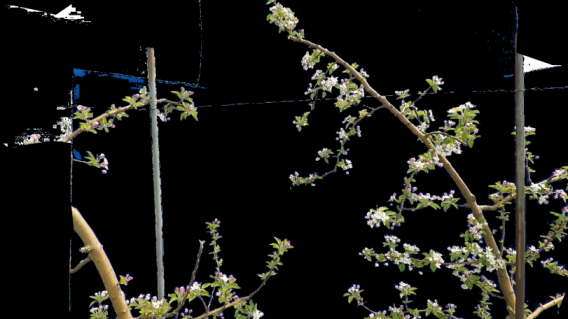}   

	\includegraphics[width=0.24\linewidth]{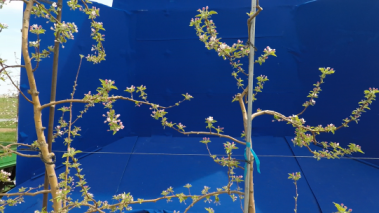}   
	\includegraphics[width=0.24\linewidth]{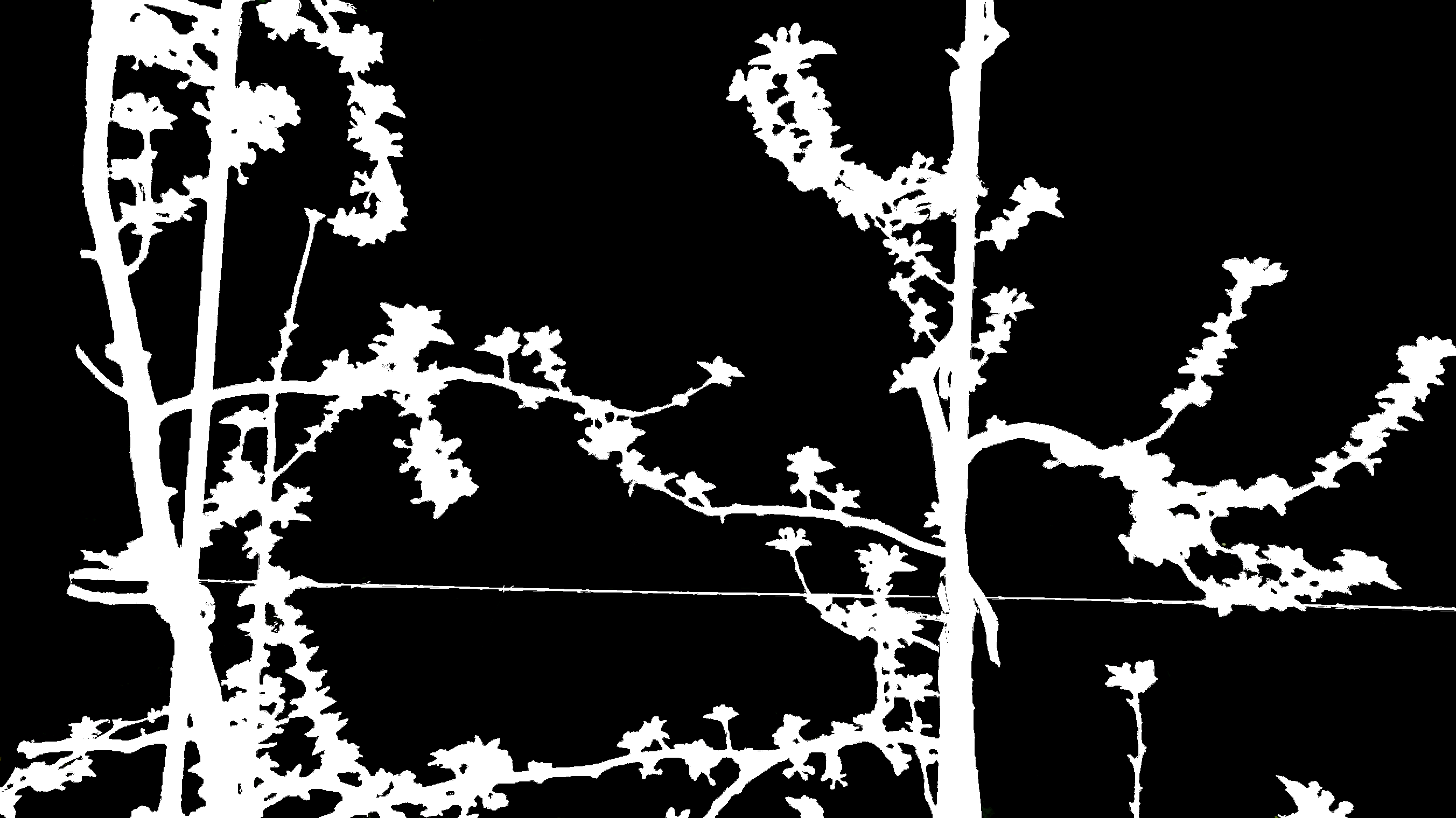}   
	\includegraphics[width=0.24\linewidth]{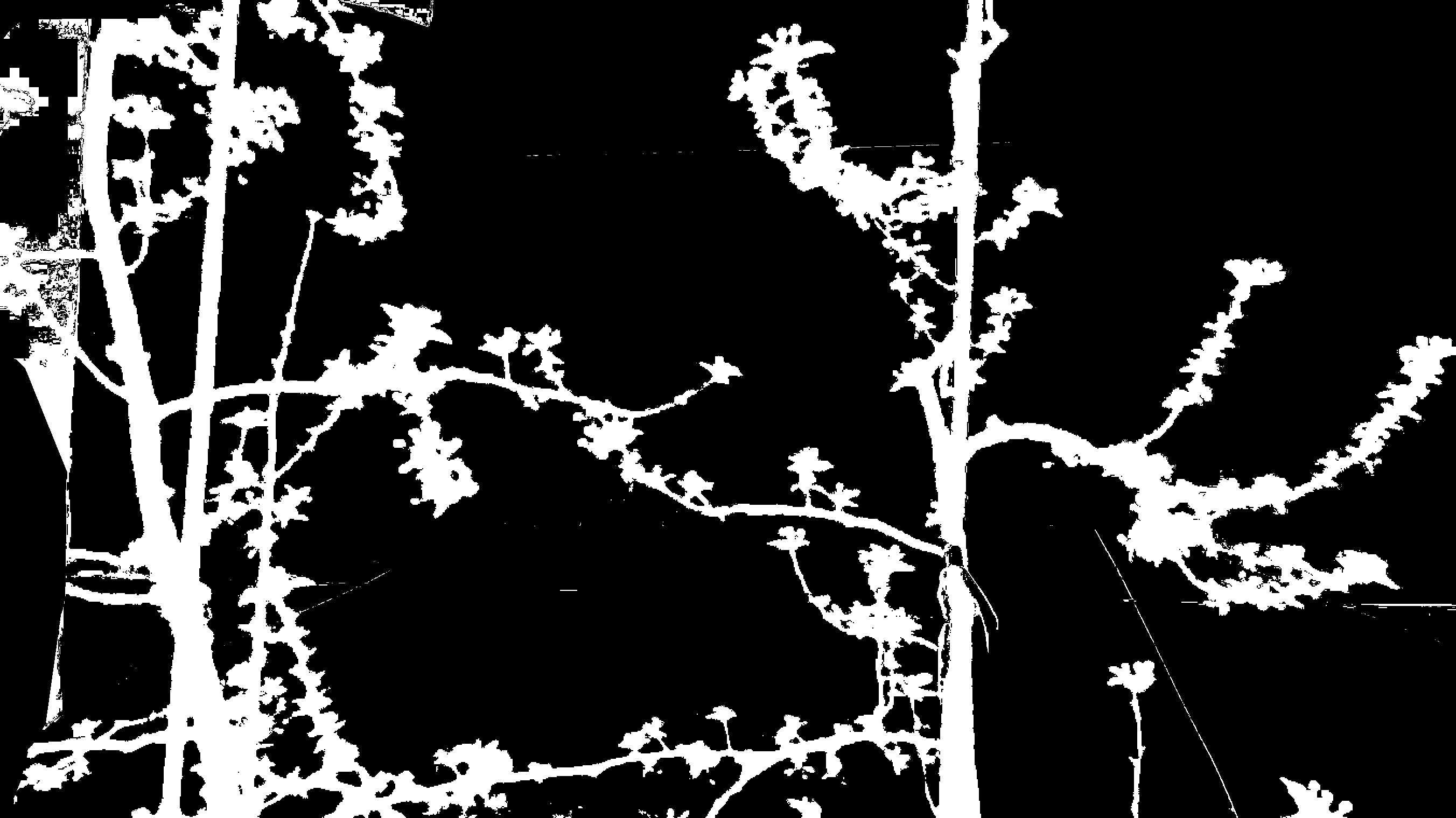}    
	\includegraphics[width=0.24\linewidth]{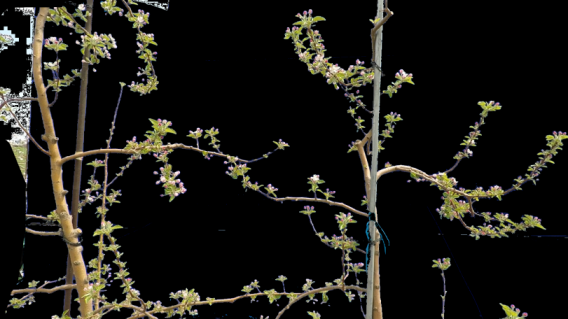}   
\caption{\textbf{[Best viewed in color]} Images of trees in front of a low-texture blue background with varying illumination levels (first column), hand-labeled ground truth (second column), the corresponding segmentations obtained using the proposed approach (third column), and the segmentation result used to mask the original image (fourth column), for the second half of the quantitative dataset (images 7-12).}
\label{fig:examples2}
\end{figure*}

Ground truth data was generated by hand-labeling tree versus non-tree pixels in 12 images with an image editor. The images of the quantitative results are from Datasets 3 and 5, which reflect two different imaging environments.  In Dataset 3, the trees are leafless and small and the background material is taller than the trees while also covering a portion of the ground.  In addition, there is a significant range of gray-level variation between the images (displayed in Figure \ref{fig:examples1}).  Dataset 5 shows a much bigger background unit, and the trees have leaves as well as flowers.  There is no portion of the background that is on the ground (see Figure \ref{fig:examples2} for examples).  Since the method was intended to segment regions that are between the camera and the blue background material, regions outside of the background unit are marked as non-tree, even if a tree is present.  The exception is the top of the background, as the sky is often marked as part of the background, and recovering the entire height of the tree is desirable.

Table \ref{tab:Quantitative-results} summarizes the performance of the comparison approaches as well as our approach in terms of the precision, recall, and the $F_1$ score. As the table shows, performance of the adaptive thresholding methods was not satisfactory.  The performance of Otsu's method in conjunction with our Step 5 mask, is generally higher in terms of precision but lower in terms of recall. The recall values for Otsu's methods are lower than that of our method for 11 out of the 12 images. Although its precision is higher for 11 out of the 12 images, for Images 7-12, which correspond to Dataset 5, the precision is on average only 0.048 higher than that of our proposed approach. Nonetheless, the $F_1$ score is higher on average using our method, particularly for Dataset 5, which included leaves and flowers. In addition, if the mask from Step 5 is not used in the Otsu method, its mean precision is reduced to $0.413$, with a corresponding $F_1$ scores $0.421$, since its recall remains essentially unchanged at  $0.720$.

\begin{figure}[ht!]
\centering 
\subfloat[Sensitivity of $p_t$]  
	{  
	\label{fig:sensitivity_pt}
	\includegraphics[width=0.90\linewidth]{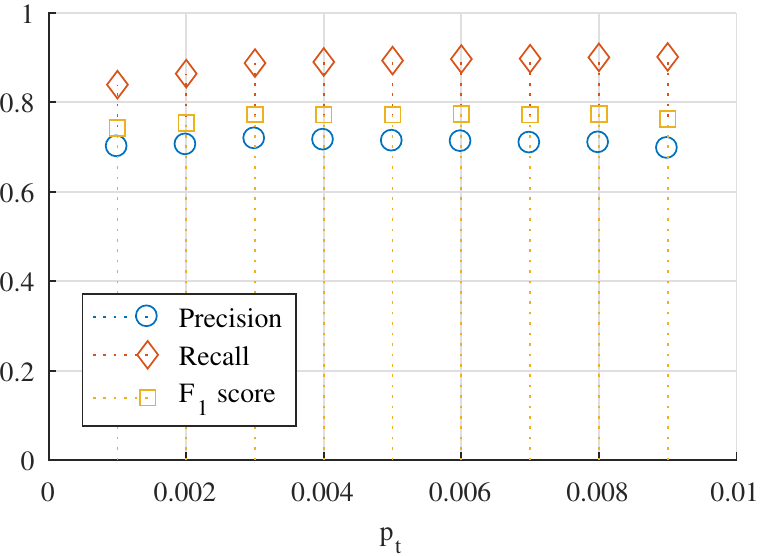}  
}

\subfloat[Sensitivity of $k$]  
{
\label{fig:sensitivity_k}
	\includegraphics[width=0.90\linewidth]{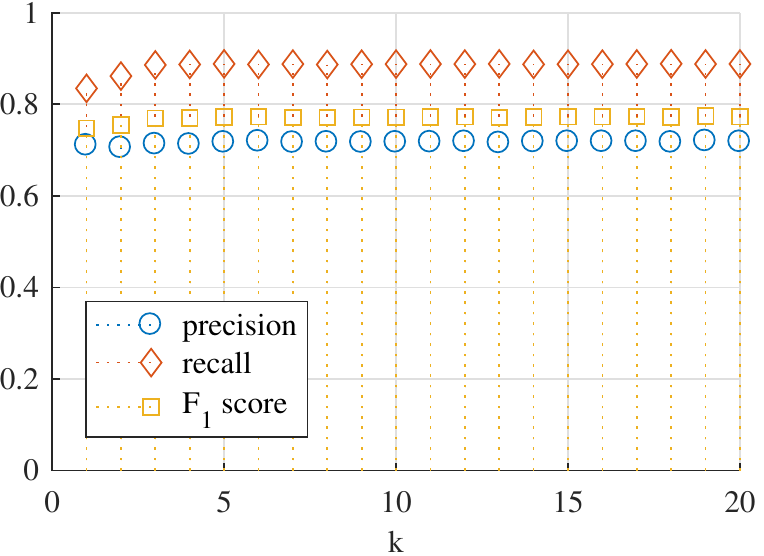} 
}
\caption{{\textbf{[Best viewed in color]} Sensitivity analysis of the proposed method to different values of threshold $p_t$ and $k$, the number of distributions in the Gaussian Mixture Model.  Values for average precision, recall, and $F_1$ score are displayed for the set of images used for the quantitative experiments.}}
\label{fig:sensitivity}
\end{figure}

In order to demonstrate that the performance of the method is relatively stable with respect to the threshold $p_t$, Figure \ref{fig:sensitivity_pt} shows a plot of the mean precision, recall, and $F_1$ score for a range of $p_t$ values. As the figure indicates, the $F_1$ score of the method varies at most $0.016$ for this range of $p_t$. {The choice of the number of distributions of the Gaussian mixture model, $k$, was also analyzed using the quantitative dataset and the results are shown in Figure \ref{fig:sensitivity_k}.  The plot shows that for the quantitative comparison dataset, if $k \in [3, 20]$ comparable results are obtained.}

\clearpage

\subsection{Qualitative analysis}

\begin{figure*}
\centering
\includegraphics[width=0.25\linewidth]{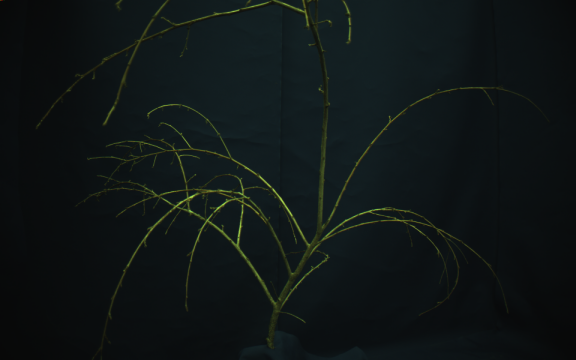}    
\includegraphics[width=0.25\linewidth]{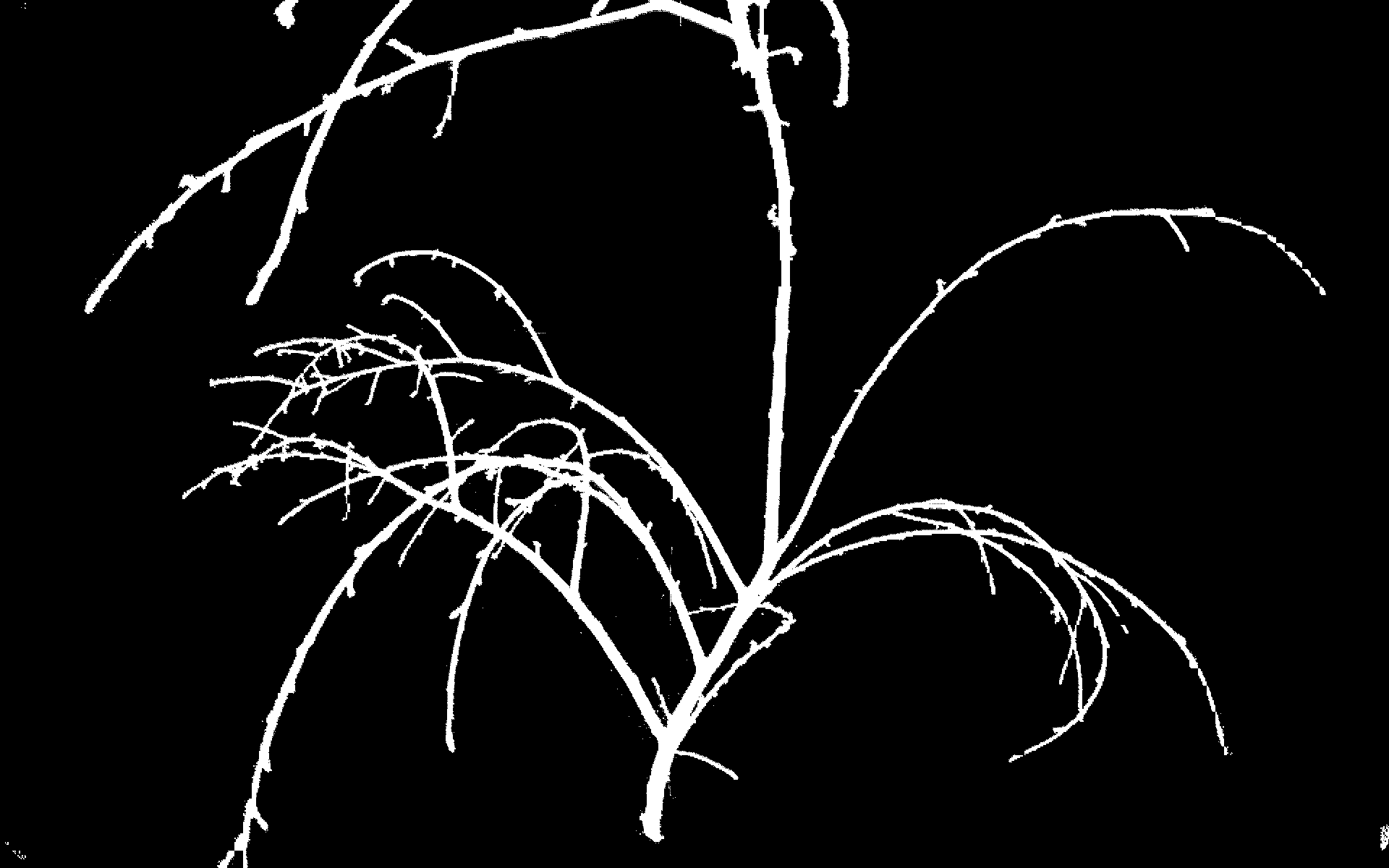}    
\includegraphics[width=0.25\linewidth]{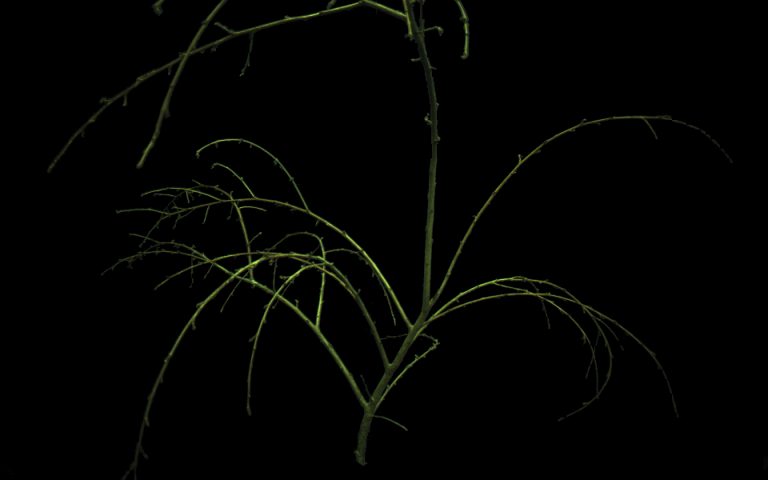}    

\includegraphics[width=0.25\linewidth]{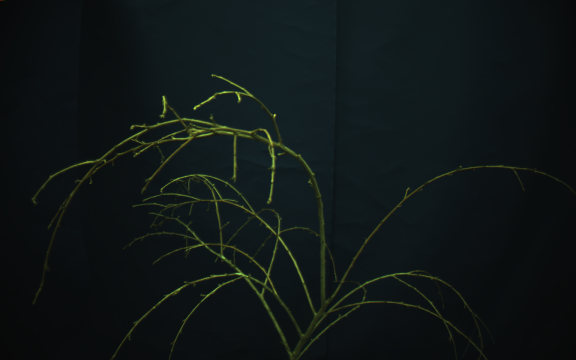}    
\includegraphics[width=0.25\linewidth]{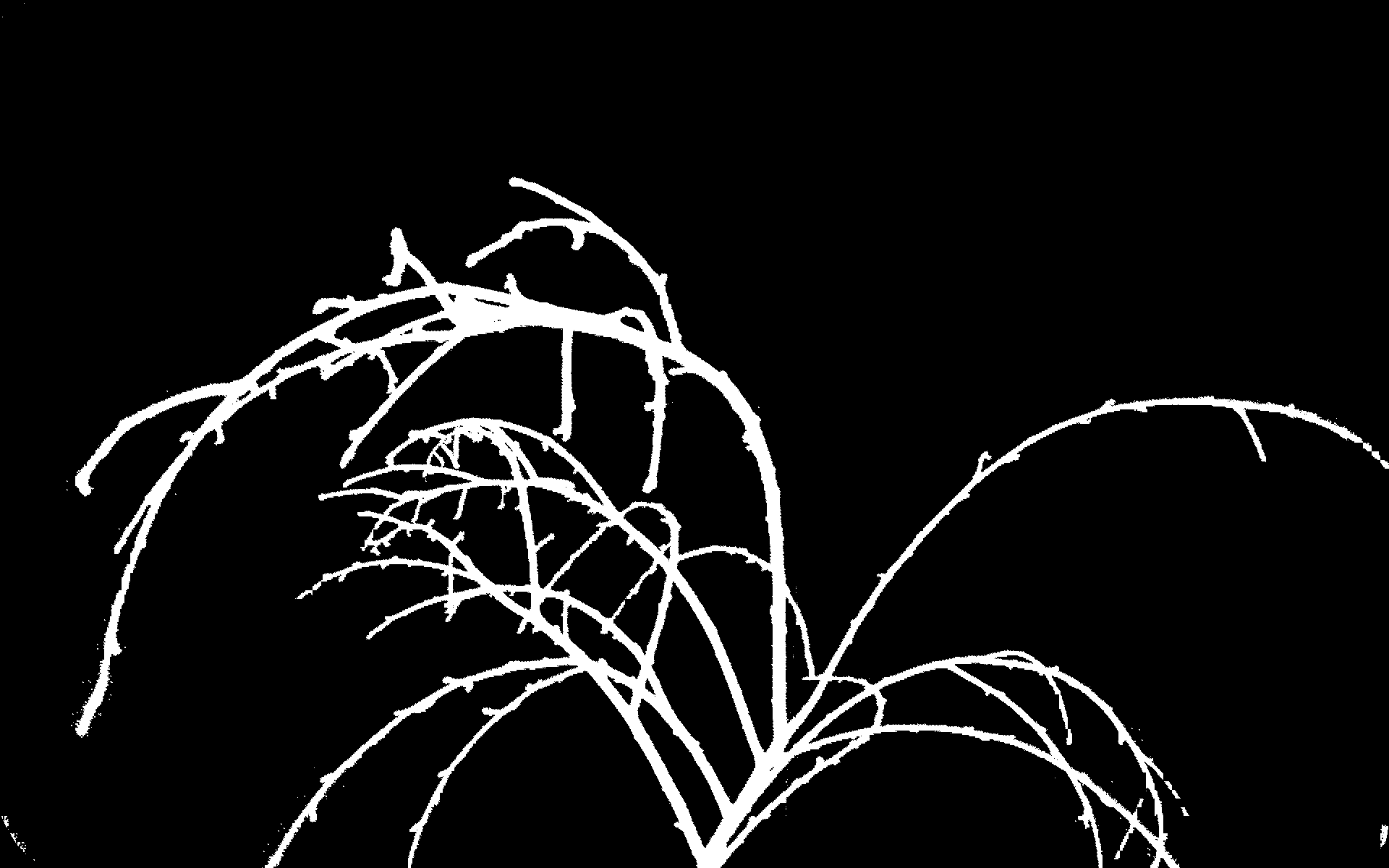}    
\includegraphics[width=0.25\linewidth]{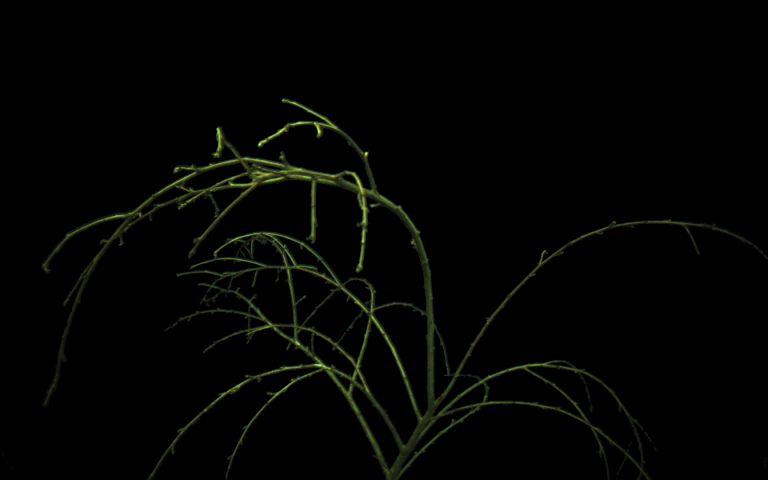}   
 
\caption{\textbf{[Best viewed in color]} Results from Dataset 1, acquired indoors.  Original images (first column), the corresponding segmentations obtained using the proposed approach (second column), and original image masked with the segmentation result (third column).  }
\label{fig:dataset1}
\end{figure*}

\begin{figure*}
\centering
\includegraphics[width=0.25\linewidth]{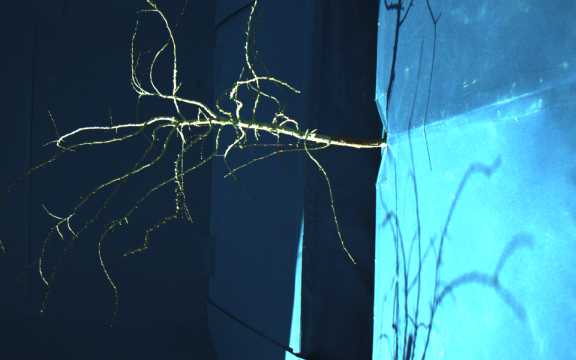}    
\includegraphics[width=0.25\linewidth]{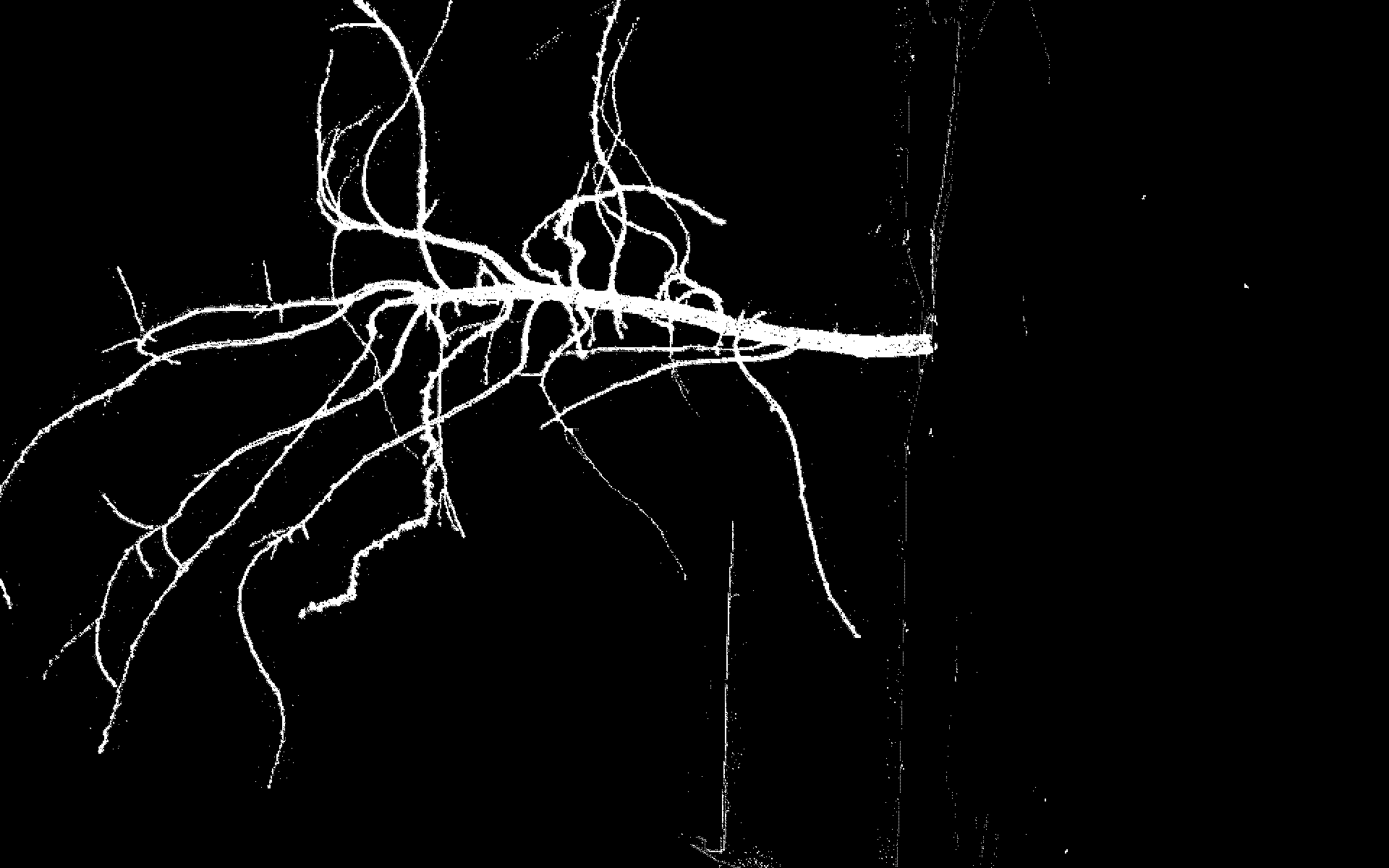}    
\includegraphics[width=0.25\linewidth]{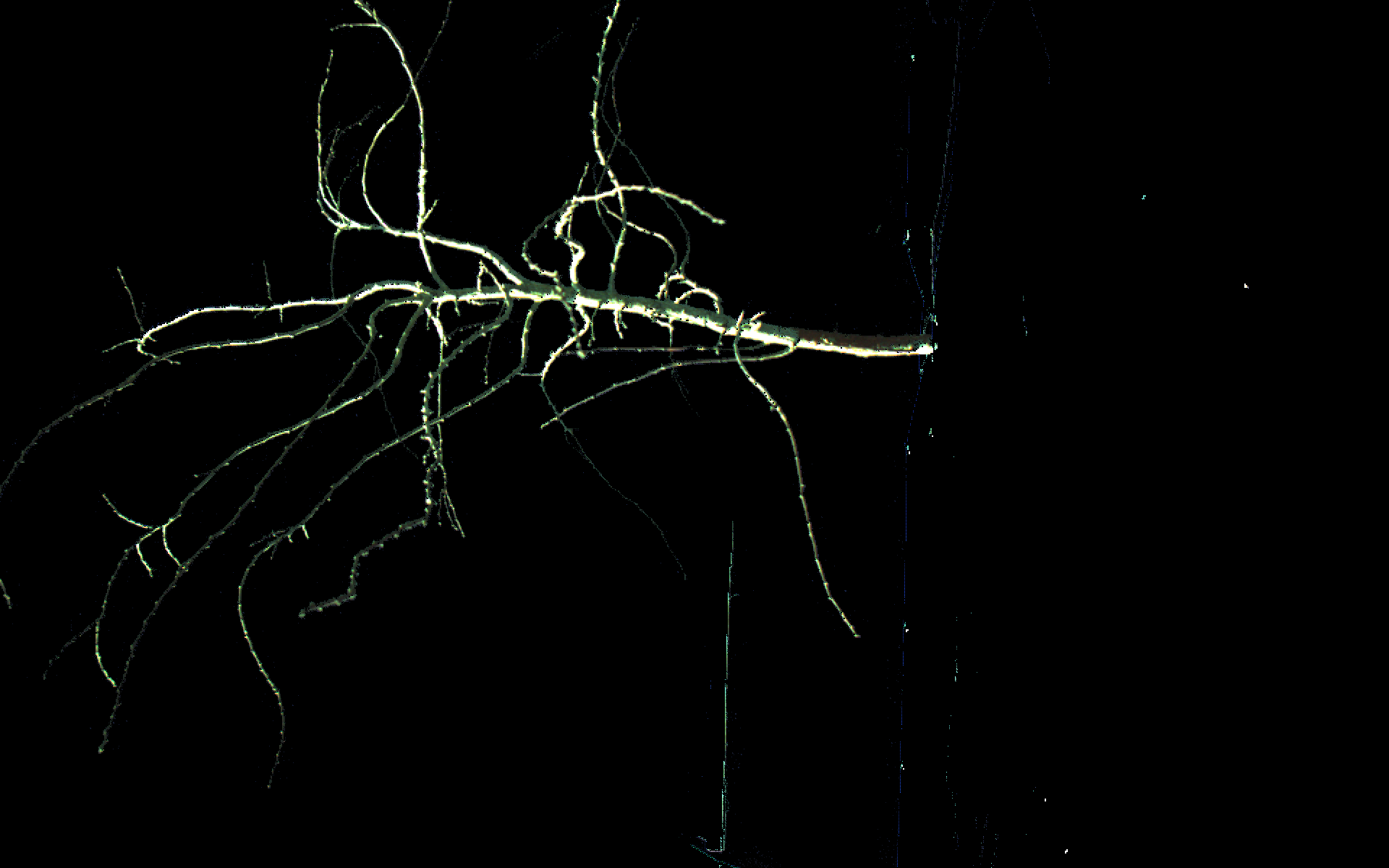}    

\includegraphics[width=0.25\linewidth]{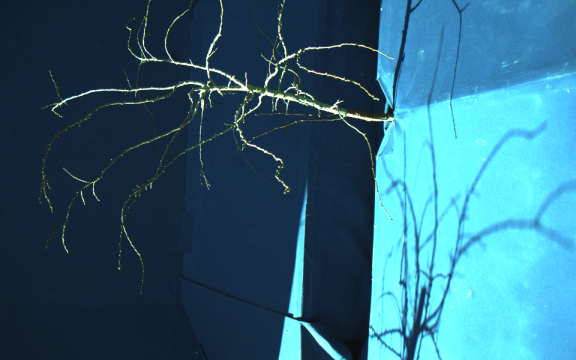}    
\includegraphics[width=0.25\linewidth]{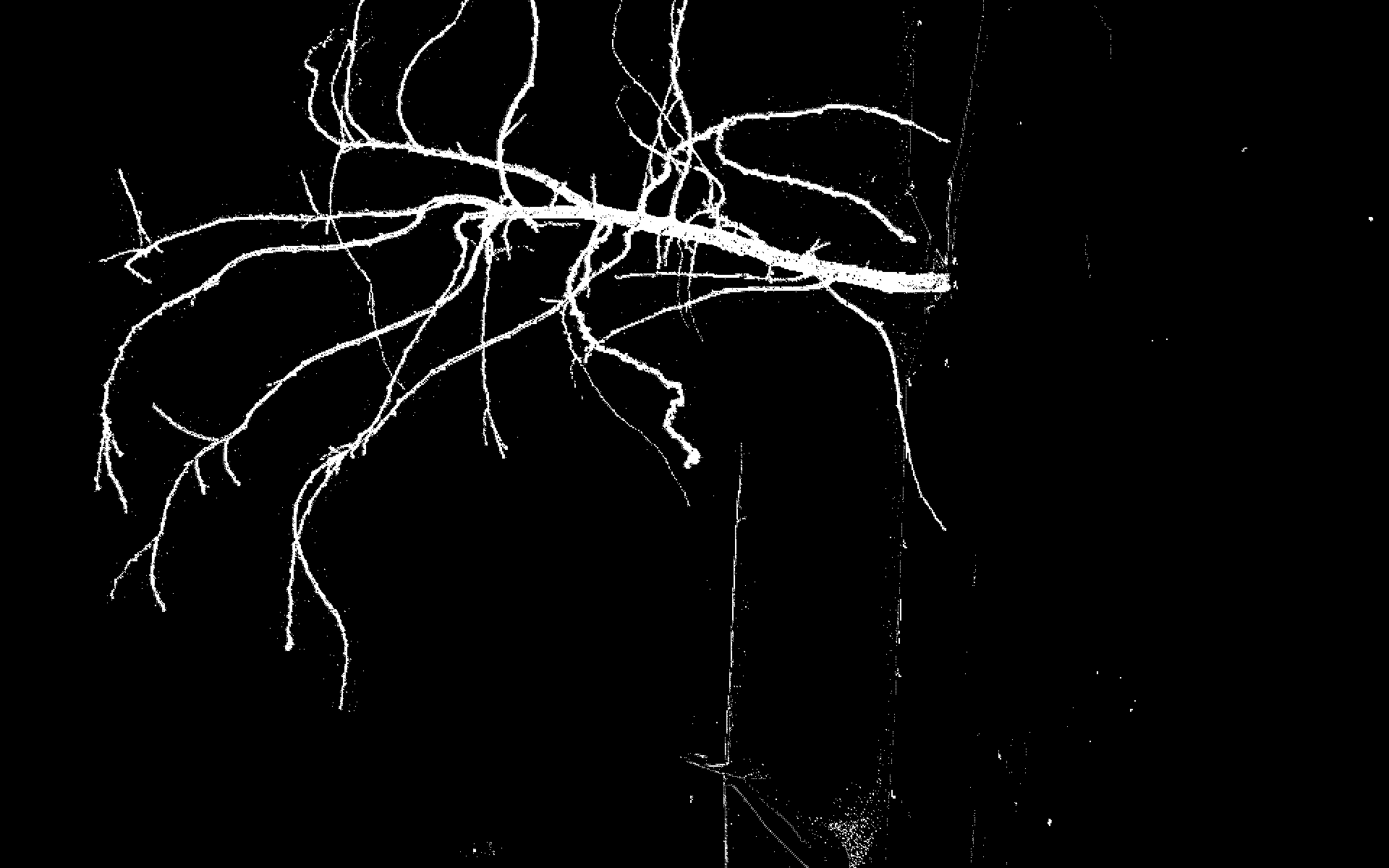}    
\includegraphics[width=0.25\linewidth]{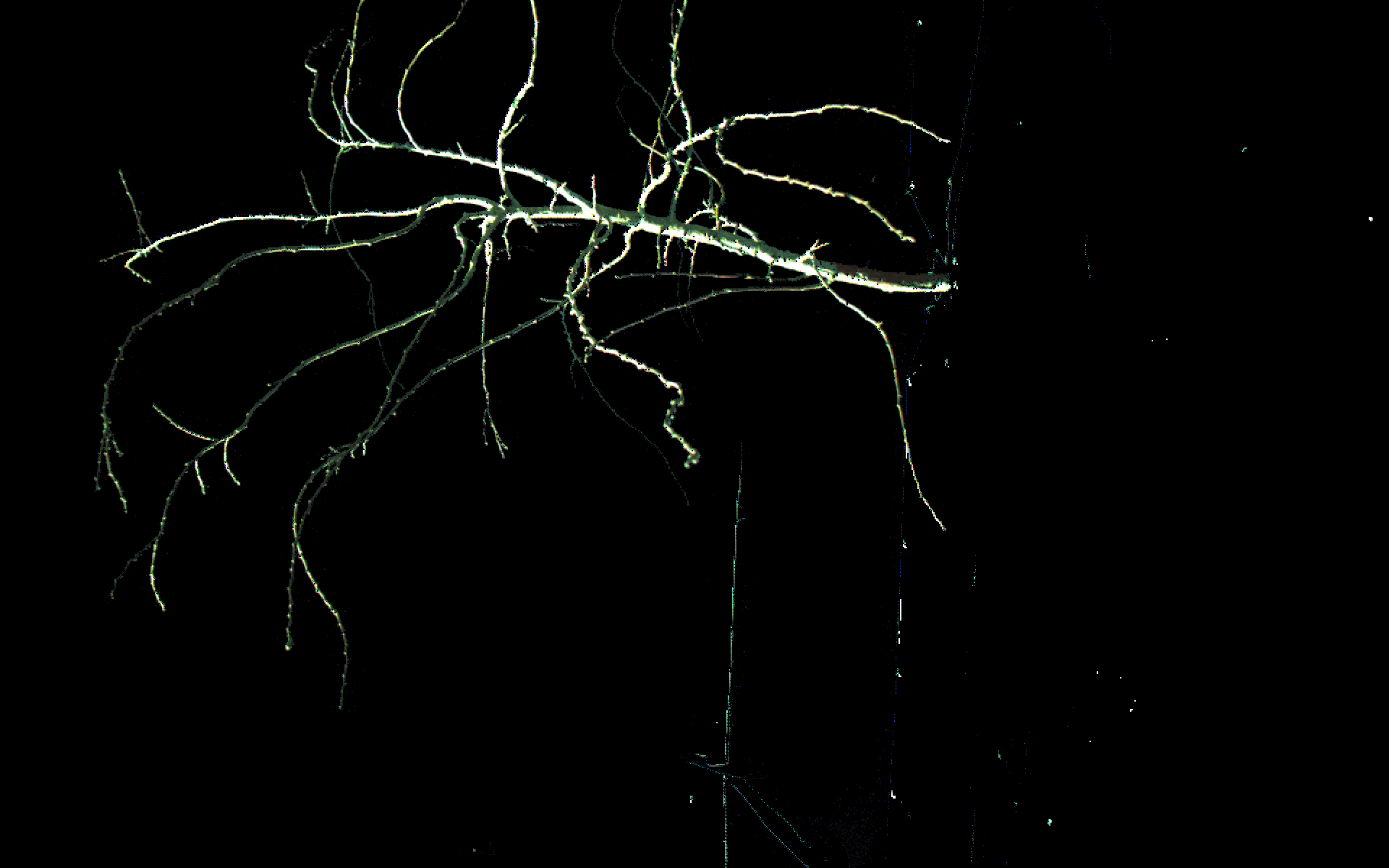}    

\caption{\textbf{[Best viewed in color]} Results from Dataset 2, acquired outdoors with significant shadow, sun, and sideways camera orientation.  Original images (first column), the corresponding segmentations obtained using the proposed approach (second column), and original image masked with the segmentation result (third column). }
\label{fig:dataset2}
\end{figure*}

\begin{figure*}
\centering
\includegraphics[width=0.25\linewidth]{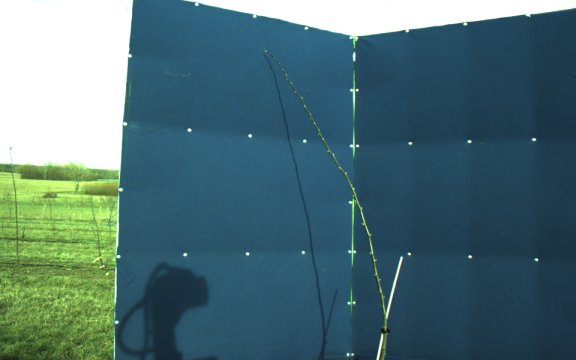}    
\includegraphics[width=0.25\linewidth]{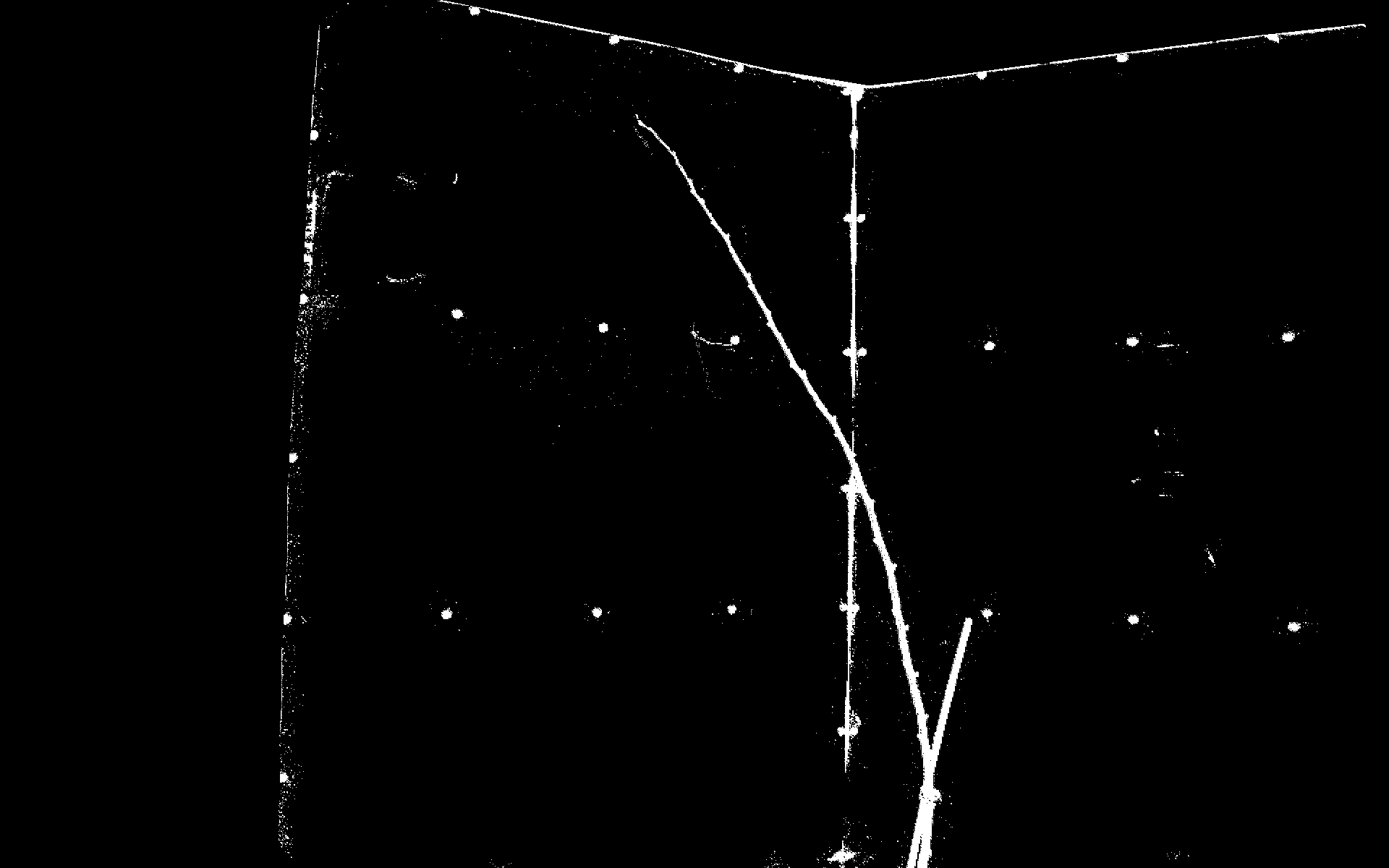}   
\includegraphics[width=0.25\linewidth]{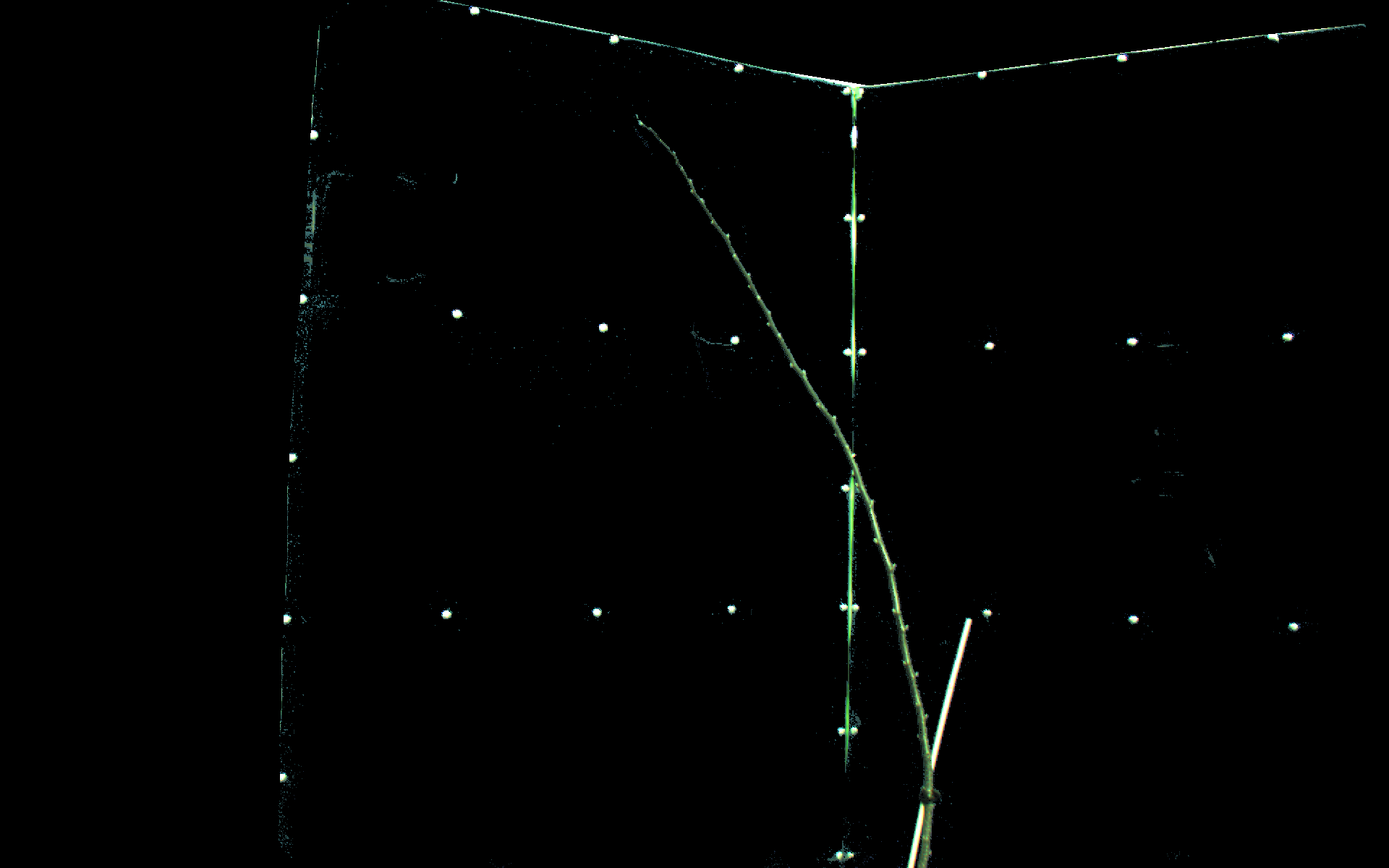}    

\includegraphics[width=0.25\linewidth]{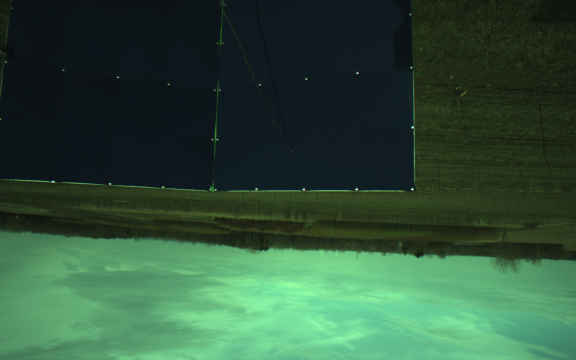}    
\includegraphics[width=0.25\linewidth]{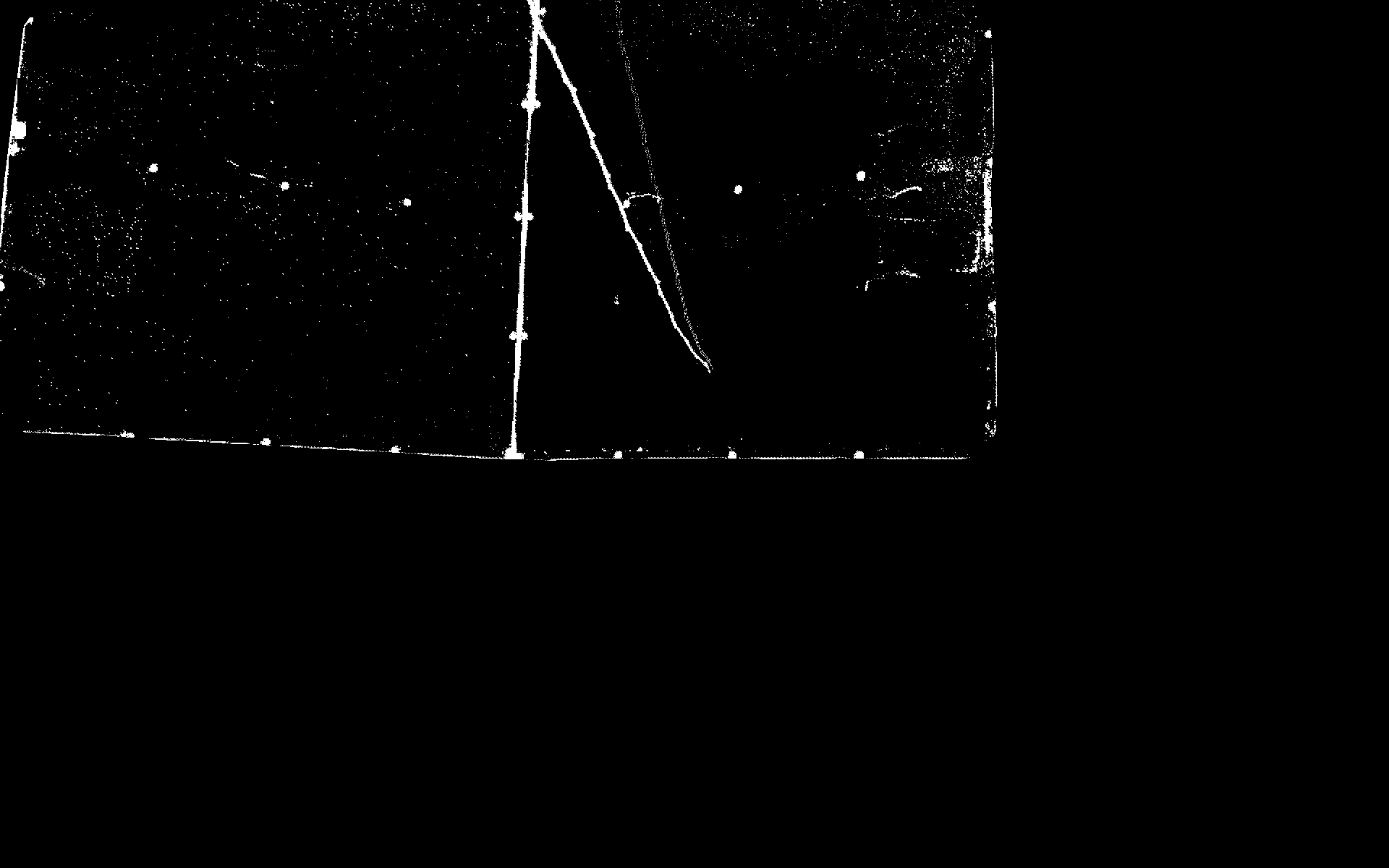}    
\includegraphics[width=0.25\linewidth]{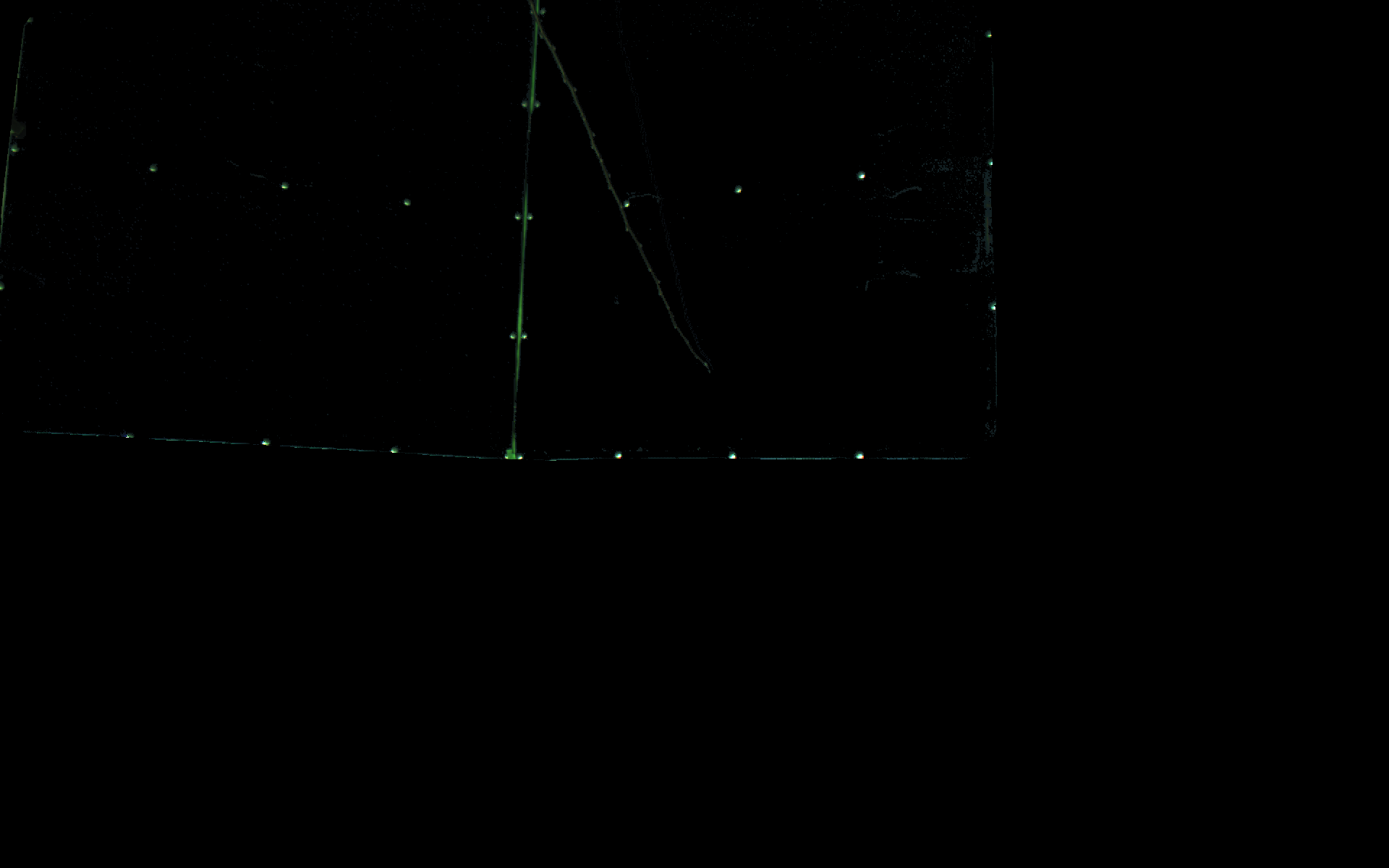}    

\caption{\textbf{[Best viewed in color]} Results from Dataset 4, acquired outdoors with significant shadows and upside down camera orientation.  Original images (first column), the corresponding segmentations obtained using the proposed approach (second column), and original image masked with the segmentation result (third column). }
\label{fig:dataset4}
\end{figure*}

\begin{figure*}
\centering
\includegraphics[width=0.25\linewidth]{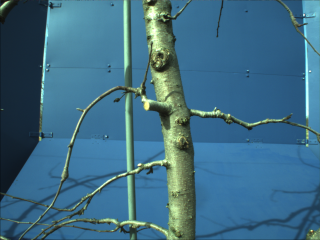}    
\includegraphics[width=0.25\linewidth]{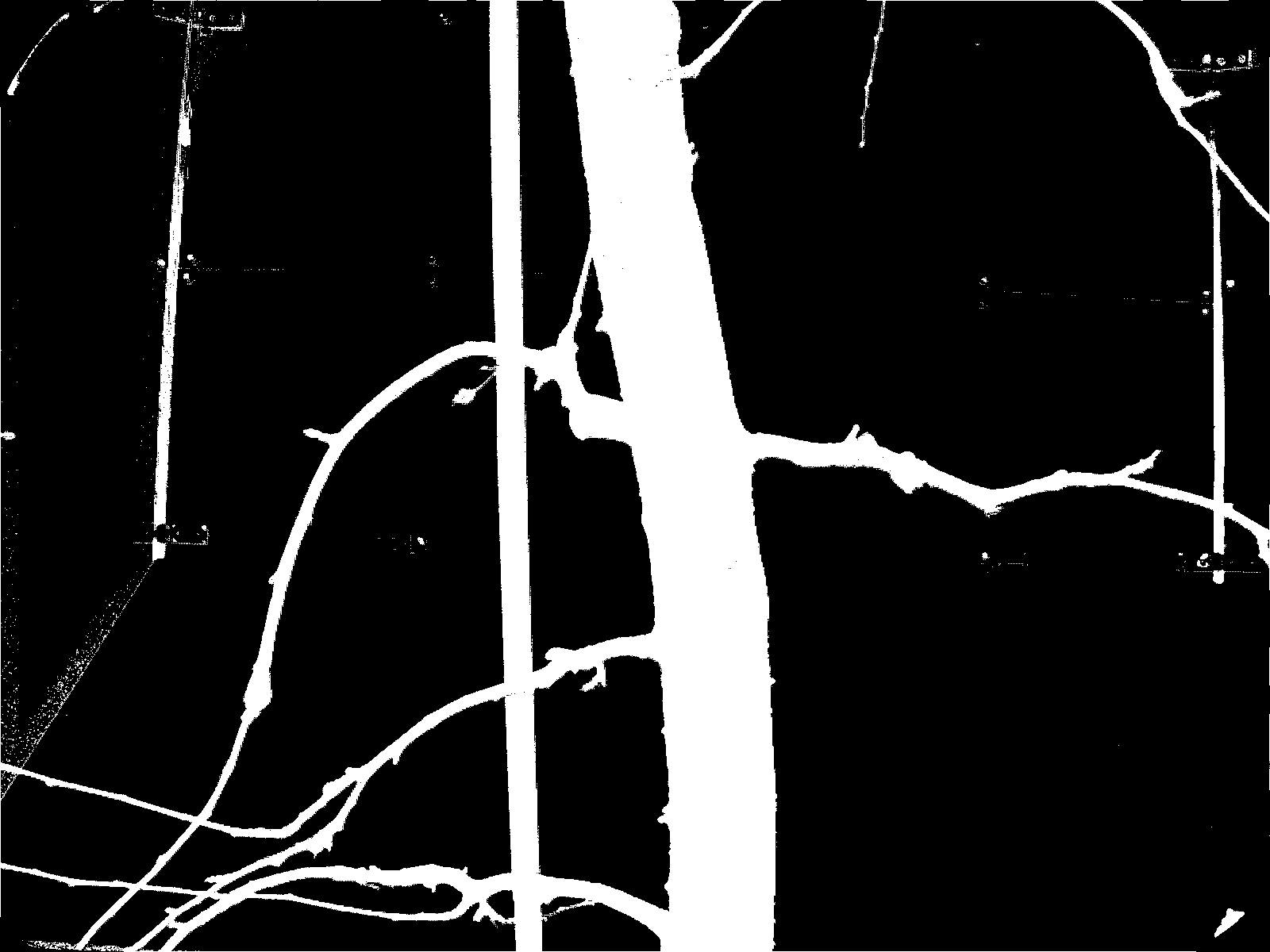}    
\includegraphics[width=0.25\linewidth]{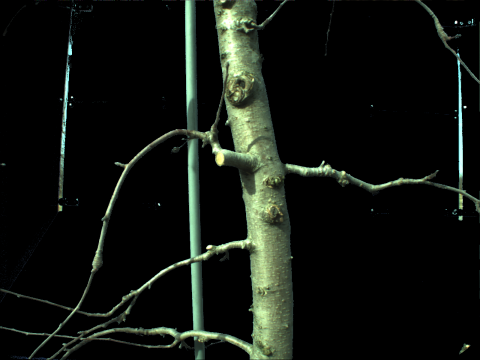}    

\includegraphics[width=0.25\linewidth]{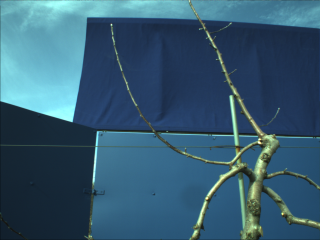}    
\includegraphics[width=0.25\linewidth]{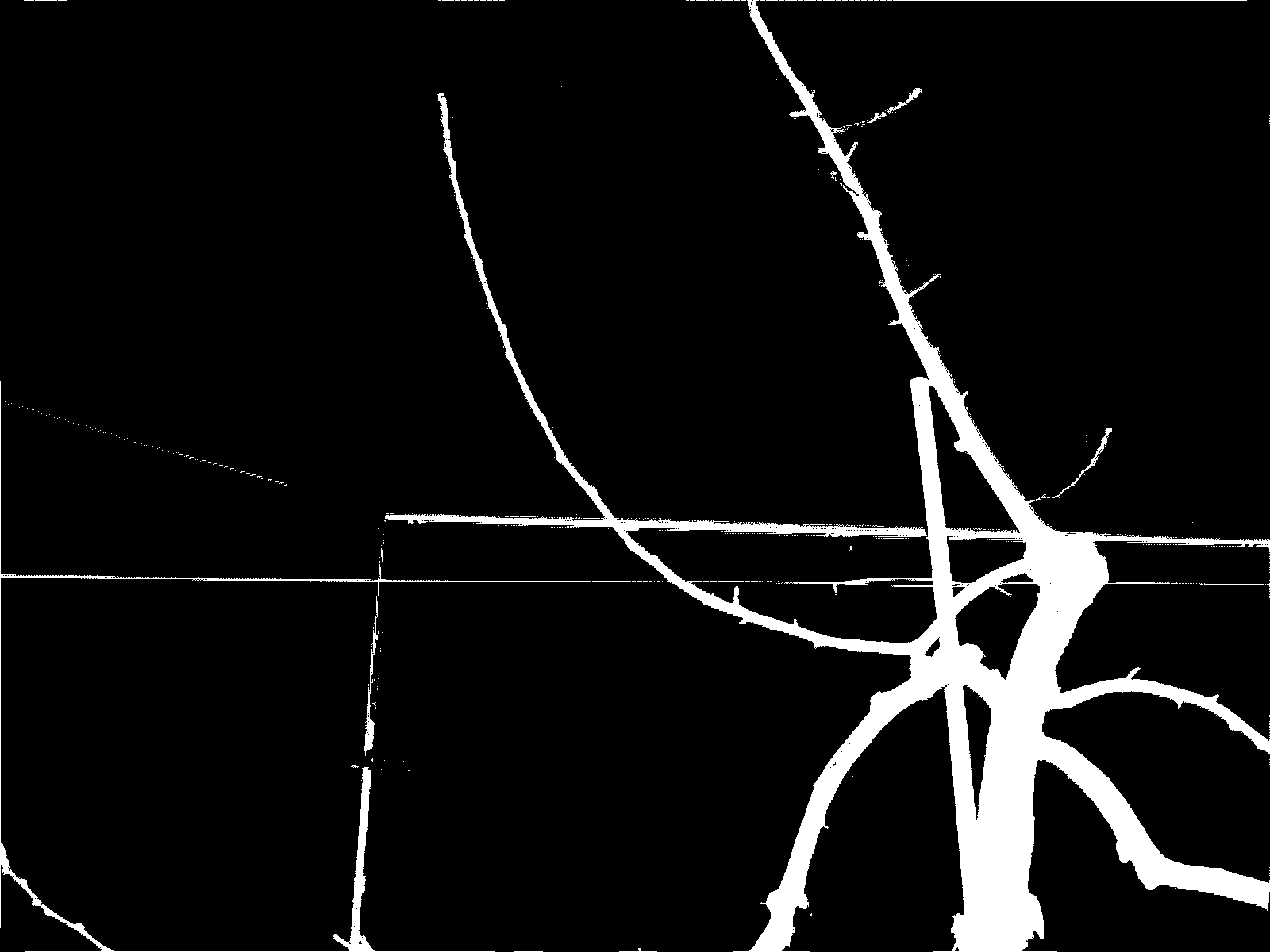}    
\includegraphics[width=0.25\linewidth]{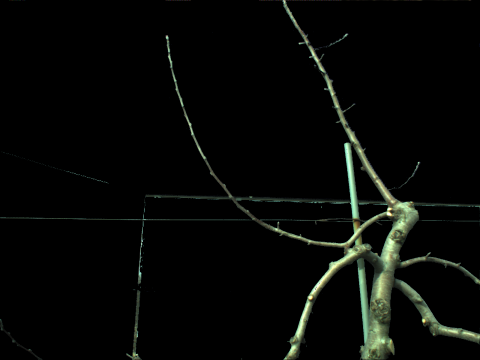}   
 
\caption{\textbf{[Best viewed in color]} Results from Dataset 6, acquired outdoors with a background unit with two different blue colors (fabric versus rigid painted areas).  Original images (first column), the corresponding segmentations obtained using the proposed approach (second column), and original image masked with the segmentation result (third column). }
\label{fig:dataset6}
\end{figure*}

The proposed method was also evaluated on images acquired from a range of conditions, including four different background unit designs in three different locations, from 2015-2017.  The datasets are described in Table \ref{tab:DatasetInfo} and images and results are displayed in Figures \ref{fig:dataset1}-\ref{fig:dataset6}.  Examples of Dataset 3 are shown in Figure \ref{fig:examples1}, while Dataset 5 examples are shown in Figure \ref{fig:examples2}.

\begin{figure*}
\centering
\includegraphics[width=1\linewidth]{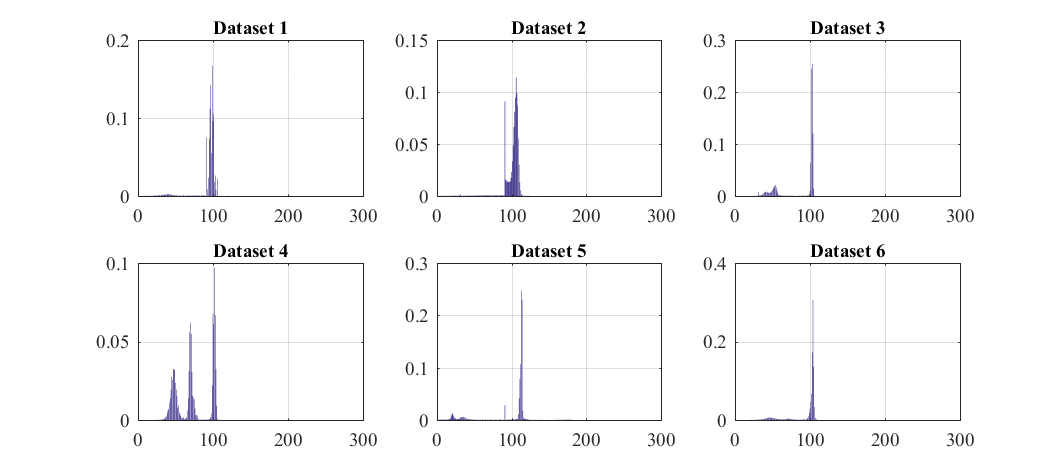} 
\caption{Examples of hue histograms from images from each of the six datasets. Although all of the histograms show a clear peak corresponding to the background unit, some datasets have multiple peaks (such as Datasets 2 and 4). }
\label{fig:hist}
\end{figure*}

The images in these datasets have differences in the distributions of hue components.  Figure \ref{fig:hist} shows hue histograms from the six datasets, and demonstrates that while all have peaks corresponding to the background unit, some datasets have multiple peaks, such as Dataset 2 and 4. {Multiple peaks may be present for a few reasons, such as different colors used in the background unit (Dataset 6), illumination differences across the background, or the presence of other large constant color objects such as the sky. Overall, despite the presence of these multiple histogram peaks, as the figures show, the proposed method segments the objects of interest in the various different scenarios accurately. As highlighted in Section \ref{ss:step2}, however, if there are scenarios in which simple histogram thresholding is not viable, multi-level methods could be employed.}

\subsection{Suitability in real-time automation contexts}

All of the results shown in this paper were generated on a workstation with one 12-core processor and 192 GB of RAM.  To the extent possible, implementation is parallelized with OpenMP.  Run times for the entire dataset and on the per-image basis are shown in Table \ref{tab:RunTimes}.  These times include loading each image and writing two result images for each image in the dataset, the binary result and the original image multiplied by the binary image for visualization purposes, such as shown in the third and fourth columns in Figures \ref{fig:examples1}-\ref{fig:examples2}.  From Table \ref{tab:RunTimes}, datasets containing larger images, such as Dataset 5, take approximately twice as long to run than smaller images. This relationship is not surprising considering that the area of the images in Dataset 5 is approximately $1.8 \times$ the area of the images in the other datasets.  In all the scenarios under consideration the low run times enable this method to be used in a real-time automation context.

\begin{table}
\centering{}\caption{\label{tab:RunTimes}Run times of the proposed method for the six datasets, for the whole dataset as well as on an average, per-image basis.}
\resizebox{\linewidth}{!}
{
\begin{tabular}{l|ccc}
\hline 
Dataset & Number of images & Total run time (s) & average time per image (ms) \\ \hline
\hline 
1 & 126 & 15.01 & 119.10 \\ \hline
2 & 228 & 27.63 & 121.20 \\ \hline
3 & 171 & 21.67 & 126.70 \\ \hline
4 & 228 & 29.14 & 127.79 \\ \hline
5 & 154 & 36.42 & 236.52 \\ \hline
6 & 94 & 12.29  & 130.74 \\ \hline
\end{tabular}
}
\end{table}

\section{Conclusion}
\label{sec:conclusions}
This paper proposed a method to perform automatic segmentation of objects of interest in dynamic outdoor conditions. We are interested in automation scenarios in which an object of interest must be segmented from a low-texture background such as in tree reconstruction. Our method estimates a Gaussian mixture model of the low-texture background, which may include the sky, by fusing information from its color distribution and from superpixels extracted from the background. As a result, the proposed method is particularly robust to substantial variations in illumination conditions.  We illustrated the performance of the proposed segmentation method in quantitative and qualitative experiments, and showed how its low run times enabled its use in real-time automation contexts. 

{\small
\bibliographystyle{ieee}
\bibliography{background_subtraction_refs}
}

\end{document}